\documentclass[10pt,journal,compsoc]{IEEEtran}
\ifCLASSOPTIONcompsoc
  \usepackage[nocompress]{cite}
\else
  \usepackage{cite}
\fi
\usepackage{graphicx}
\usepackage{amsmath}
\usepackage{longtable}
\usepackage{color}
\usepackage{longtable}
\usepackage{makecell}
\usepackage{lscape}
\usepackage{mathtools}
\usepackage{graphicx}
\usepackage{subfig}

\usepackage{color}

\ifCLASSINFOpdf
\else
\fi
\usepackage{algorithmic}
\usepackage{array}
\usepackage{fixltx2e}
\usepackage{stfloats}
\usepackage{url}

\begin{document}

\title{Comprehensive Comparative Study of Multi-Label Classification Methods}

\author{Jasmin~Bogatinovski,~Ljup\v{c}o~Todorovski,~Sa\v{s}o~D\v{z}eroski,~and~Dragi~Kocev
\IEEEcompsocitemizethanks{\IEEEcompsocthanksitem J.~Bogatinovski is with the Department of Distributed Operating Systems, Technical University Berlin, Berlin, Germany and with the Department of Knowledge Technologies, Jo\v{z}ef Stefan Institute, and Jo\v{z}ef Stefan International Postgraduate School, Ljubljana, Slovenia   \protect\\
L. Todorovski is with the Faculty of Public Administration, University of Ljubljana, and the Department of Knowledge Technologies, Jo\v{z}ef Stefan Institute, Ljubljana, Slovenia \protect\\
S. D\v{z}eroski and D. Kocev are with the Department of Knowledge Technologies, Jo\v{z}ef Stefan Institute, and Jo\v{z}ef Stefan International Postgraduate School, Ljubljana, Slovenia\protect\\
D. Kocev is also with the Bias Variance Labs, d.o.o., Ljubljana, Slovenia.
\protect\\
E-mails: Jasmin.Bogatinovski@tu-berlin.de (also ijs.si, gmail.com), Ljupco.Todorovski@fu.uni-lj.si, Saso.Dzeroski@ijs.si, Dragi.Kocev@ijs.si
}
\thanks{This work has been submitted to the IEEE for possible publication. Copyright may be transferred without notice, after which this version may no longer be accessible. \copyright 2021 IEEE.  Personal use of this material is permitted.  Permission from IEEE must be obtained for all other uses, in any current or future media, including reprinting/republishing this material for advertising or promotional purposes, creating new collective works, for resale or redistribution to servers or lists, or reuse of any copyrighted component of this work in other works.}}
\markboth{This work has been submitted to the IEEE for possible publication. Copyright may be transferred without notice, after which this version may no longer be accessible. © 20XX IEEE.  Personal use of this material is permitted.  Permission from IEEE must be obtained for all other uses, in any current or future media, including reprinting/republishing this material for advertising or promotional purposes, creating new collective works, for resale or redistribution to servers or lists, or reuse of any copyrighted component of this work in other works.}%
{Bogatinovski \MakeLowercase{\textit{et al.}}: Comprehensive Comparative Study of Multi-Label Classification Methods}

%
%
%
%

\newcommand\MYhyperrefoptions{bookmarks=true,bookmarksnumbered=true,
pdfpagemode={UseOutlines},plainpages=false,pdfpagelabels=true,
colorlinks=true,linkcolor={black},citecolor={black},urlcolor={black},
pdftitle={Bare Demo of IEEEtran.cls for Computer Society Journals},
pdfsubject={Typesetting},
pdfauthor={Bogatinovski et al.},
pdfkeywords={Comprehensive Study, Multi-Label Classification, }}

\newcommand{\jasmin}[1]{{\color{green} Jasmin: #1}}
\newcommand{\dragi}[1]{{\color{red} Dragi: #1}}
\newcommand{\ljupco}[1]{{\color{purple} Ljupco: #1}}

\markboth{}%
{Bogatinovski MakeLowercase{\textit{et al.}}: Comprehensive Comparative Study of Multi-Label Classification Methods}
%

\IEEEpubid{\copyright~2021 IEEE. Permission from IEEE must be obtained for all uses, in any current or future media, including reprinting/republishing this material for advertising or promotional purposes.}


\IEEEtitleabstractindextext{%
\begin{abstract}

Multi-label classification (MLC) has recently received increasing interest from the machine learning community. Several studies provide reviews of methods and datasets for MLC and a few provide empirical comparisons of MLC methods. However, they are limited in the number of methods and datasets considered. This work provides a comprehensive empirical study of a wide range of MLC methods on a plethora of datasets from various domains. More specifically, our study evaluates 26 methods on 42 benchmark datasets using 20 evaluation measures. The adopted evaluation methodology adheres to the highest literature standards for designing and executing large scale, time-budgeted experimental studies. 
First, the methods are selected based on their usage by the community, assuring representation of methods across the MLC taxonomy of methods and different base learners. Second, the datasets cover a wide range of complexity and domains of application. The selected evaluation measures assess the predictive performance and the efficiency of the methods.
The results of the analysis identify RFPCT, RFDTBR, ECCJ48, EBRJ48 and AdaBoost.MH as best performing methods across the spectrum of performance measures.  Whenever a new method is introduced, it should be compared to different subsets of MLC methods, determined on the basis of the different evaluation criteria.

\end{abstract}

}

\maketitle

\IEEEdisplaynontitleabstractindextext

%
\IEEEpeerreviewmaketitle

\IEEEraisesectionheading{\section{Introduction}\label{sec:introduction}}


\IEEEPARstart{P}{redictive modeling} is an area in machine learning concerned with developing methods that learn models for predicting the value of a target variable. The target variable is typically a single continuous or discrete variable, corresponding to the two common tasks of regression and classification, respectively. However, in practically relevant problems, more and more often, there are multiple properties of interest, i.e., target variables. Such practical problems include image annotation with multiple labels (e.g., an image can depict trees and at the same time the sky, grass etc.), predicting gene functions (each gene is typically associated with multiple functions) and drug effects (each drug can have effect on multiple conditions).
The problems with multiple binary variables as targets corresponding to the question if a given example is associated with a subset from a set of predefined labels belong to the widely known task of multi-label classification (MLC)\cite{Madjarov2012, Herrera2016, Tsoumakas2007}.


In binary classification, the presence/absence of a single label is predicted. In MLC, the presence/absence of multiple labels is predicted and multiple labels can be assigned simultaneously to a sample. Most often, the MLC task is confused with multi-class classification (MCC). In MCC, there are also multiple classes (labels) that a given example can belong to, but a given example can belong to only one of these multiple classes. In that spirit, the MCC task can be seen as a special case of the MLC task, where exactly one label is relevant for each example.
Furthermore, the MLC task is different from the task of multi-target classification (MTC) \cite{Kocev:Journal:2013}, which is concerned with predicting several targets, each of which can take only one value of several possible classes. MLC can be viewed as a collection of several binary classification tasks, and MTC of several MCC tasks. Finally, another task related to MLC is multi-label ranking. The goal of multi-label ranking is to produce a ranking/ordering of the labels regarding their relevance to a given example \cite{Madjarov2012}. 

MLC predicts the set of target attributes (called \textit{labels}) that are relevant for each presented sample. This task arises from practical applications. For example, Xu et al.~\cite{GO} introduce 3 instantiations of the task of predicting the subcellular locations of proteins according to their sequences. The dataset contains protein sequences for humans, viruses and plants. Both GO (Gene Ontology) terms and pseudo amino acid compositions are used to describe the protein sequences. The goal is to predict the relevant sub-cellular locations for each of the proteins. There are 6 subcellular locations for viruses, 12 subcellular locations for plants, and 14 subcellular locations for humans. Briggs et al.~\cite{birds} address the prediction of the type of birds whose songs are present in a given recording using audio signal processing. The set of labels consists of 19 species of birds. Multiple birds may sing simultaneously on a given recording. The problem of topic classification for a text document is a MLC problem, since many of the documents refer to more than one topic at once. For example, Katakis et al.~\cite{bibtex} present data containing entries from the BibSonomy social bookmark and publication sharing system, annotated with several tags. Next, Boutell et al.~\cite{scene} present an image dataset containing images annotated with several labels (beach, sunset, fall foliage, field, urban and mountain). Although the main focus of MLC applications is in text, biology and multimedia, the potential for using MLC in other domains is constantly increasing (medicine \cite{CHD, biomedicalimagesegmentation, neonatalbrains}, environmental modeling \cite{WaterQuality}  social sciences \cite{SocialSciences}, commerce \cite{commerce} etc.), 
In \cite{RecentTrends}, an extensive summary of the various emerging trends and subareas of MLC are given: extreme multi-label classification, multi-label learning with limited supervision, deep multi-label learning, online multi-label learning, statistical multi-label learning, and rule-based multi-label learning.

\begin{figure}[htb]
	\centering
	\includegraphics[width=0.5\textwidth]{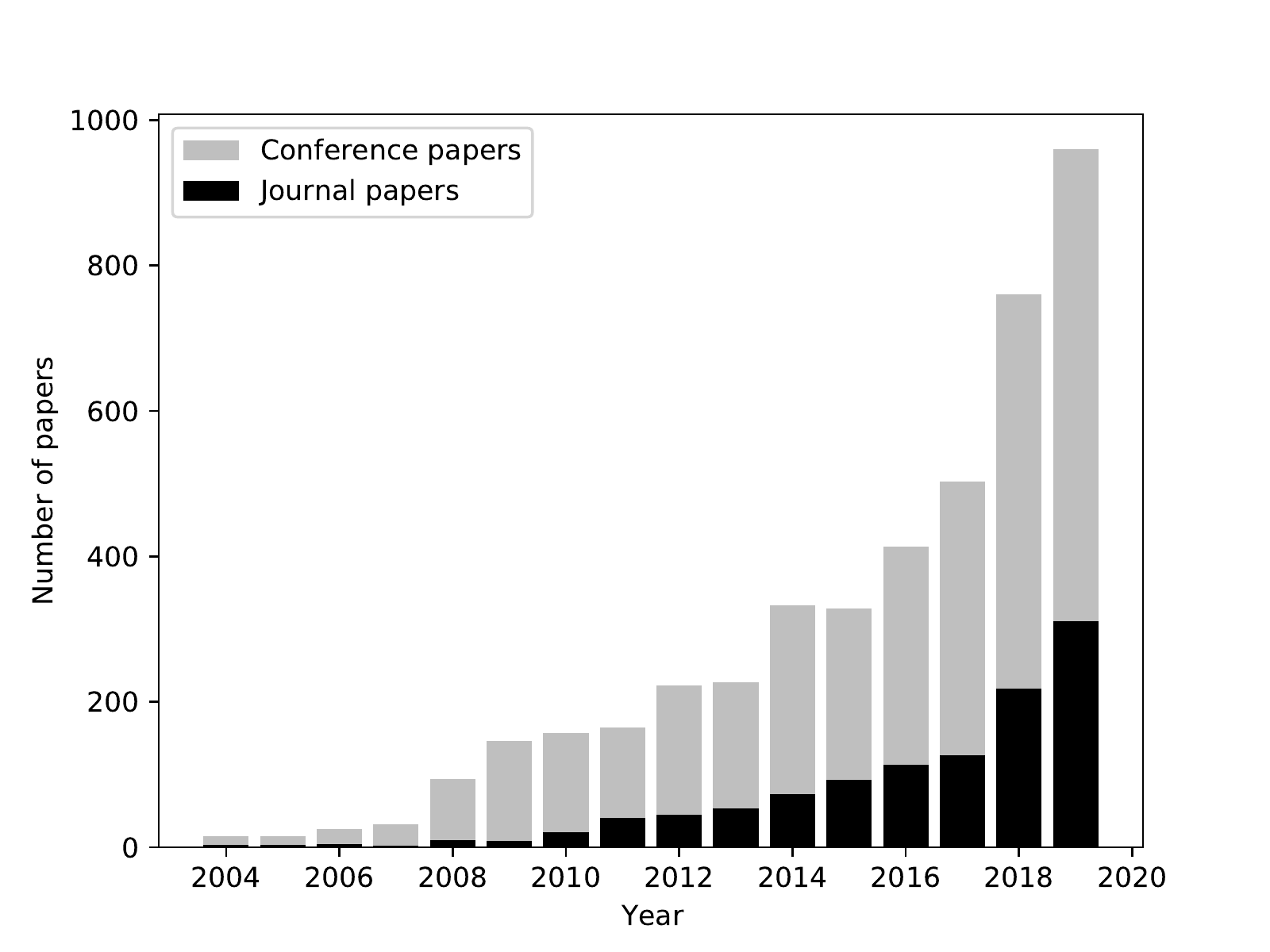}
	\caption[The number of papers related to the topic of MLC through years.]{A summary of the number of papers from the SCOPUS database (\url{https://www.scopus.com/}) related to the topic of MLC. The vertical axis represents the number of conference and journal papers related to the topic of MLC. An almost exponential curve of progress can be observed. The absence of large experimental studies with rigorous extensive experimental empirical comparison amplifies the importance of performing a comprehensive study on MLC methods to provide a survey of the landscape of methods.}
	\label{fig:SCOPUS}
\end{figure}

\figurename~\ref{fig:SCOPUS} shows the increasing interest in the task of MLC from the machine learning community. The increasing trend indicates the appearance of novel MLC methods. Given the large pool of problems, multi-label methods and datasets, it is not easy for a novice in the area and even for an experienced practitioner to select the most suitable method for their problem. Moreover, it is not clear what benchmarking baselines need to be used when proposing a novel method. Landscaping the existing methods and problems is thus as a necessity for further advancement of this research area. 

There are several previous attempts at addressing this issue. However, they have a limited scope in terms of the methods and/or the datasets used in the evaluation. Some of these studies require a special emphasis, since they have helped to shape the field by providing a theoretical and empirical discussion on the properties of the various MLC methods. We discuss these studies in the chronological order of their appearance in the literature.

The first comprehensive empirical study for the task of MLC was provided by Madjarov et al. \cite{Madjarov2012}. In their work, a comprehensive analysis of 12 methods on 11 benchmarking problems for 16 evaluation criteria is provided. It is a systematic review and provides guidance for the practitioners tackling MLC tasks, about which methods to choose. However, from the current perspective, given the wealth of newly proposed methods and problems/datasets, it is outdated in terms of the inclusion of problems and methods that have been introduced in the last decade. 

The second study provides details on the MLC tasks \cite{Gibaja2014}. It provides a concise organization of all points on the methods, evaluation criteria and problems. But it lacks a comprehensive empirical evaluation of the methods across different datasets. The third study \cite{Zhang2014} provides an in-depth theoretical treatise of 8 MLC methods, together with their pseudo-codes and a discussion of how the methods deal with the MLC task. 

Next, the drawbacks of the previous three studies are addressed (to some extent) in \cite{Herrera2016}. As a book on the topic of MLC, it provides an extensive overview of existing methods and their comparative analysis on subsets of problems. But, it lacks the experimental rigour and depth of \cite{Madjarov2012}.

Furthermore, the most recent study \cite{Moyano2018} provides a similar experimental setup as in \cite{Madjarov2012} with an extension towards the analysis of ensembles of MLC methods. It argues that ensemble learning methods are superior in terms of performance to other, single-model learning approaches. While this is true in many cases, the computational time one requires for building an ensemble is larger than the time for building a single model. Given the complexity of the MLC task, this may arise as a limitation in practical applications. Zhang et al.~\cite{Zhang2018} focus on providing an overview of a specific type of MLC methods, referred to as binary relevance, but do not assess their predictive performance. In a similar limited context, Rivolli et al.~ \cite{Rivolli2020} present an empirical study of 7 different base learners used in ensembles on 20 datasets.


A shared property of the previous studies is the focus on a smaller part of the landscape of methods and problems. However, given the plethora of problems and methods introduced in recent years, a comparative analysis on a larger scale is highly desired. \textbf{The main aim of this work is to fill this gap by performing an extensive study of MLC in terms of both methods and problems/datasets}. Simultaneously, it aims to identify the strengths and limitations of existing methods beyond predictive performance and efficiency.


By performing the extensive experimental study, a landscape map of the MLC will be obtained: It will \textbf{include the performance of 26 MLC methods evaluated on 42 benchmark datasets using 18 performance evaluation measures}. Next, it will reveal the best performing methods per method group and per evaluation measure. Hence, it will identify the most suitable baselines that need to be used when proposing a new method. Furthermore, it will outline the strengths and weaknesses of the MLC methods in respect to one another. Moreover, it will highlight the used MLC methods in terms of their potential for addressing several MLC specific properties (e.g., label dependencies and high-dimensional label spaces).

\textbf{This study is the most comprehensive experimental work for the task of MLC performed thus far}. In a nutshell, it identifies a subset of 5 methods that should be used in baseline comparisons: RFPCT (Random Forest of Predictive Clustering Trees) \cite{Kocev:Journal:2013}, RFDTBR (Binary Relevance with Random Forest of Decision Trees) \cite{Tsoumakas2007}, ECCJ48  (Ensemble of Classifier Chains built with J48) \cite{Read2010}, EBRJ48 (Ensemble of Binary Relevance built with J48) \cite{Read2010} and AdaBoost \cite{Schapire1999}. These methods show the best performance. The first two methods are computationally much more efficient as compared to the competition. Detailed results from this study as well as descriptions of the methods, data sets, and evaluation measures will be made available in a repository at \url{http://mlc.ijs.si/}.





The remainder of the paper is organized as follows. In Section~\ref{sec:MLCtask}, we formally define the task of MLC task, and review/ describe the available MLC datasets and problems. Section~\ref{sec:MLCmethods} organizes the MLC methods into a taxonomy of methods, describes the methods in detail and discusses the way these methods address specific properties of MLC, such as handling label-dependence and high-dimensional label spaces. Section~\ref{sec:MLCexperimentalDesign} outlines the design of the experimental study, by describing the experimental methodology and setup, the parameter instantiations, the evaluation measures, and the statistical analysis of the obtained results. Section~\ref{sec:MLCresults} discusses the obtained results from different viewpoints. Finally, Section~\ref{sec:Conclusions} concludes and presents the main outcomes of the study, together with some directions on the use of MLC methods for benchmarking.


%
%
%
%

 

\section{The task of Multi-label Classification}\label{sec:MLCtask}

This section begins by describing the benchmark datasets considered in the study. Next, it defines the task of MLC. Finally, it overviews the methods considered in the study.

\subsection{Multi-Label Classification Datasets}

The majority of the datasets considered in this study come from application areas such as biology, text and multimedia. The datasets from biology, in general, include proteins representation as descriptive variables and as targets either gene function prediction or sub-cellular localization. 
The datasets from the domain of text, most often represent the problem of topic classification for news documents. However, there are datasets from this domain that predict other targets, such as cardiovascular condition prediction based on medical reports or tag recommendation for reviews. The datasets from the multimedia domain can be split into two major categories: datasets concerned with the classification of images (most often scenes in given images) and datasets concerned with the prediction of audio content (either genre, emotions etc).

There are also datasets that come from other domains such as medicine and chemistry. The datasets that come from the medical domain represent the prediction of diseases based on symptoms or state of the patient based on vital measurements such as the blood pressure. The dataset from the domain in chemistry is concerned with the prediction of chemical concentration in observed subjects. The diversity of the available MLC datasets witness the great application potential of MLC.

\begin{figure}[htb]
	\centering
	\includegraphics[width=0.5\textwidth]{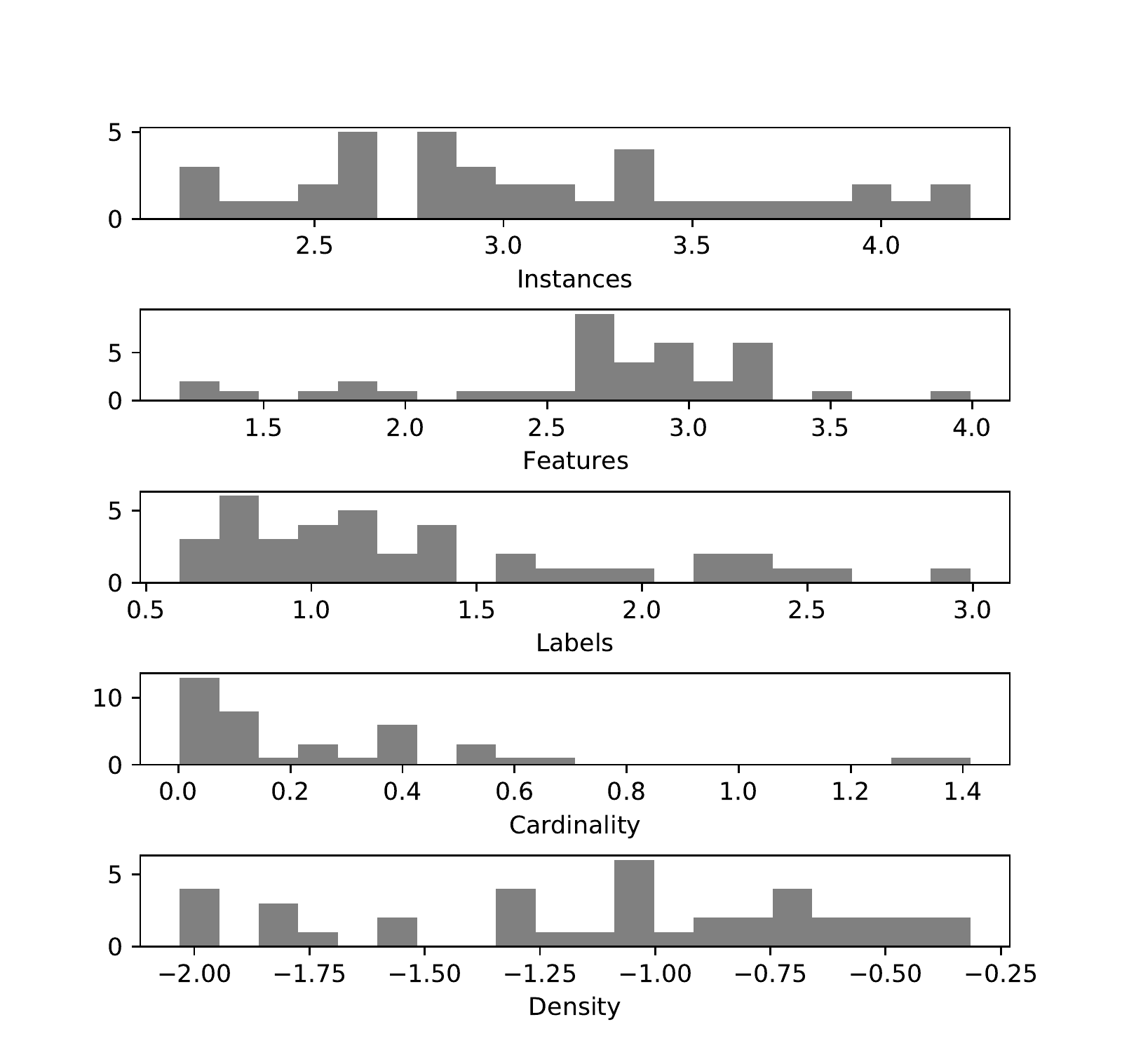}
	\caption{Distribution of the values of four meta-features on 42 used datasets. The feature values are shown on a log scale.}
	\label{fig:datasetPropertiesBasicLog}
\end{figure}

The MLC datasets are described with five basic meta-features: number of instances/examples, number of features, number of labels, label cardinality and label density. The distribution of the datasets across these meta-features is depicted in \figurename~\ref{fig:datasetPropertiesBasicLog}. The \textit{number of training examples} ranges from 174 to 17190. The wide range of the number of samples provides an opportunity to test the strengths and weaknesses of the MLC methods when presented with different level of richness of input data in terms of the number of examples.

Additionally, the richness in the description of the data (\textit{number of features}) ranges from 19 to 9844. Predominantly the number of features range from 300 until 2000. Regarding the type of features, there are few datasets with a mixture of nominal and numeric attributes. In most of the datasets, the features are either solely numeric or solely nominal. The \textit{number of labels} is a unique property of the MLC classification task. The number of labels ranges from 4 to 374, with one dataset being an exception, containing 983 labels. Predominantly the number of labels is in the range from 4 until 53.

Additional discussion about the datasets follows in terms of the meta properties of \textit{label cardinality} and \textit{label density}. Label cardinality is a measure of labels distribution per example. It is defined as the mean number of labels associated with an example \cite{Tsoumakas2007}. In many of the datasets, this meta-feature is smaller than 1.5. It indicates that these datasets on average have one label associated with their examples. In most of the cases, there are no more than three labels assigned to an example in a dataset. As exceptions are the datasets \texttt{delicious} and \texttt{cal500}, which have around 19 and 26 labels assigned for each example on average, respectively. 

Label density is a measure of the frequency of the labels. It is calculated as the division of the label cardinality by the number of labels. It indicates the frequency of the labels among all the instances.

\subsection{Task description}
The task of MLC can be viewed as an instantiation of the structure output prediction paradigm \cite{Kocev2011,Kocev:Journal:2013}. The goal is for each example to define two sets of labels -- the set of relevant and the set of irrelevant labels. Following \cite{Madjarov2012}, the task of MLC is defined as: \\

{\bf Given:}
\begin{itemize}
 \item {an example space $\mathcal{X}$ consisting of tuples of values of primitive data types (categorical or numeric), i.e., $\forall {\bf x_{i}} \in \mathcal{X}, {\bf x_{i}} = (x_{i_{1}},x_{i_{2}}, ...,x_{i_{D}})$, where $D$ denotes the number of descriptive attributes,}
  \item {a label space $\mathcal{L} = \{\lambda_{1},\lambda_{2},. . .,\lambda_{Q}\}$ which is a set of $Q$ possible labels,}
  \item {a set of examples $E$, where each element is a pair of a tuple from the example space and a subset of the label space, i.e., $E = \{({\bf x_{i}}, \mathcal{Y}_{i}) | {\bf x_{i}} \in \mathcal{X}, \mathcal{Y}_{i} \subseteq \mathcal{L}, 1 \leq i \leq N \}$ and $N$ is the number of examples of $E$ ($N = |E|$), and}
  \item {a quality criterion $q$, which rewards models with high predictive performance and low complexity.}
\end{itemize}

{\bf Find:} a function $h$: $\mathcal{X} \rightarrow 2^{\mathcal{L}}$ such that $h$ maximizes $q$. 

\subsection{Multi-Label Classification Methods}

There is a plethora of MLC methods presented in the literature. In this paper, we follow the taxonomy of the methods as proposed in \cite{Tsoumakas2007}. The MLC methods are separated into two categories problem transformation and algorithm adaptation. 
The group of problem transformation methods approaches the problem of MLC with transforming the multi-label dataset into one or multiple datasets. These datasets are then approached with simpler, single-target machine learning methods and build one or multiple single-target models. At prediction time, it is required that all built models are invoked to generate the prediction for the test example. 

Algorithm adaptation methods include some adaptation of the training and prediction phases of the single target methods towards handling multiple labels simultaneously. For example, trees change the heuristic used when creating the splits, Support Vector Machines (SVMs) employ additional threshold technique etc. The adaptations provide a mechanism to handle the dependency between the labels directly. Their grouping is based on the underlying paradigm being adapted. The literature recognizes five defined groups of algorithm adaptation methods according to the performed adaptation: trees, neural networks, support vector machines, instance-based and probabilistic \cite{Herrera2016}. There are additional methods that utilize various approaches from other domains e.g. genetic programming, but they lack a common ground to unite them and are characterized as an unspecified group of methods. For more details, one can refer to \cite{Herrera2016}.

\section{Methods for multi-label classification}\label{sec:MLCmethods}

In this section, we discuss the MLC methods used in the experimental evaluation. We first describe the problem transformation methods. 
Next, we describe the algorithm adaptation methods and their ensemble variants. Finally, a discussion on how each of the methods is addressing two properties of the MLC task: label dependencies and high dimensional label spaces (including computational complexity analysis of the methods). 

\subsection{Problem transformation methods}

\begin{figure}[htb]
	\centering
	\includegraphics[width=0.45\textwidth]{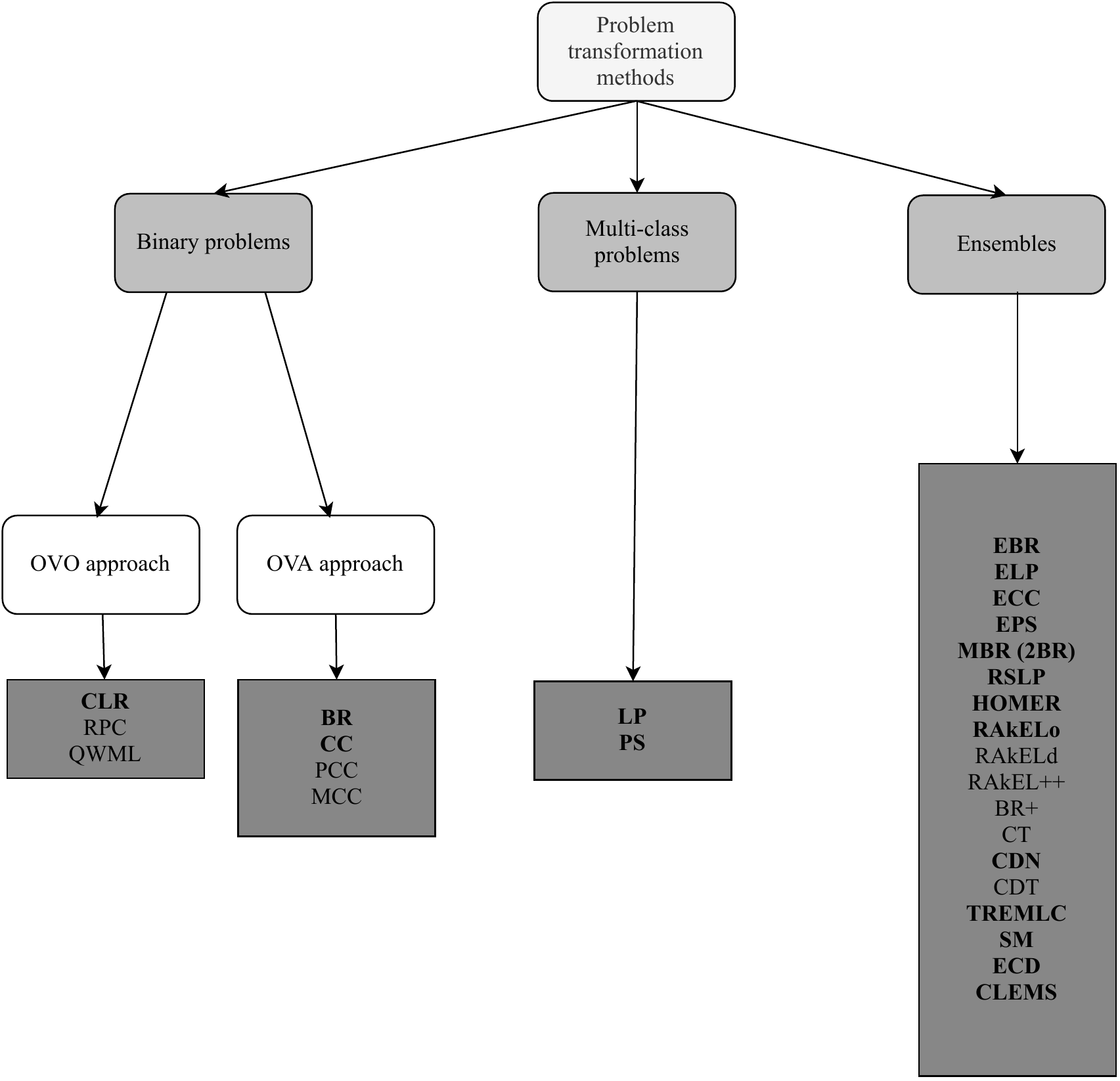}
	\caption{Problem transformation methods. The methods used in this study are shown in bolded.}
	\label{fig:PTT}
\end{figure}

There are two main ideas in the problem transformation methods -- decomposition of the problem of a set of binary problems or a (set of) multi-class problem(s). \figurename~\ref{fig:PTT} depicts the problem transformation methods used in this study and their organization into a taxonomy of methods. 

The first idea observes a multi-label dataset as a composition of multiple single-target datasets sharing the same feature space. Such an approach has the benefit of providing a straightforward application of single-target binary methods. At prediction time all trained single target models are invoked to produce the result for the new test sample. This approach, however, loses information about the dependency between the labels. 
Regarding the process of creating the multiple single binary target datasets, this group of methods is further grouped into \texttt{One-Vs-One}-like and \texttt{One-Vs-All}-like methods. In the former approach (also known as binary relevance or pairwise approach \cite{Gibaja2014}), each pair of labels is considered to produce a quadratic number of single-target binary datasets. 
In the latter approach, the problem is transformed directly to $\mathcal{|L|}$ single target multi-class problems by using all unique label sets as a separate class (also known as label powerset approach). 

Both approaches allow for the use of simpler classifiers: the binary relevance uses binary classifiers, while the label powerset uses multi-class classifiers. The advantages of the latter over the former are that a single model is learned (compared to quadratic number of models) and label dependencies are preserved (compared to the complete obliteration of this information in the binary relevance approach). However, a strong limitation of the label powerset methods is the inability to generalize beyond the label-sets present in the training dataset. 
The shortcommings of these two approaches are addressed to some extend by using them in the context of ensemble learning. We next discuss these methods in more detail.


\subsubsection{Binary relevance methods}

The Binary Relevance method \textbf{(BR)} \cite{Tsoumakas2007} transforms the MLC problem into $\mathcal{|L|}$ binary classification problems that share the same feature (descriptive) space as the original descriptive space of the multi-label problem. Each of the binary problems has assigned one of the labels as a target. It trains one base binary classifier for each of the transformed problems. It has only one hyperparameter - the base classifier. This method generalizes beyond the label-sets present in the training samples. It is not suitable for a large number of labels and ignores the label correlations. Due to the necessity of building models for each label, the training of the method can be time-consuming, especially if the computational complexity of the base learning method is large.

Calibrated Label Ranking \textbf{(CLR)} is a pairwise technique for multi-label ranking. It provides a built-in mechanism to extract bipartitions and thus can be used as a MLC method. The core of the pairwise methods is creating $\frac{\mathcal{|L|}(\mathcal{|L|}-1)}{2}$ single target binary datasets from the multi-label dataset with label-set of size $\mathcal{|L|}$, maintaining the original descriptive space. The binary target is generated in such a way that if one of the labels in a given pair, chosen as positive, is different from the other, the example in the newly created dataset obtains a value of 1 and 0 otherwise. If the labels are the same, the example is excluded. In such way $\frac{\mathcal{|L|}(\mathcal{|L|}-1)}{2}$ binary datasets are created. A base classifier is built on these datasets. CLR introduces one artificial label \cite{Brinker2006, Frnkranz2008}. This artificial label acts as a complementary label for each of the original labels, thus introducing $\mathcal{|L|}$ more models to be built. When ranking for each of the labels is obtained, the artificial variable acts as a split point between the relevant and irrelevant labels, producing bipartition. It has one hyperparameter to be chosen - the base learner. A strong advantage of this method is that it generates both ranking and bipartition. The main drawback is that it is not so suitable for datasets with a large number of labels, due to the large exploration space and time complexity.

Classifier Chains \textbf{(CC)} \cite{Read2011} learning procedure involves two steps. It consists of training $\mathcal{|L|}$ single target binary classifiers as in BR connected in a chain. Each classifier deals with single target problem of augmented feature space consisting of all the descriptive features and the predictions obtained from the previous classifier in the chain (the first classifier on the chain is learned only using the descriptive features). The only hyperparameter to set is the base classifier. A strength of this method is the introduction of label correlation to some extent (the order of the labels in the chain is important) and can generalize beyond seen label-sets. The limitations of the method include its prohibitive use for datasets with a large number of labels because it is applying BR, and the dependence on the ordering of the labels along the chain.

\subsubsection{Label Powerset methods}
Label Powerset \textbf{(LP or LC)} \cite{Tsoumakas2007} transforms the MLC method into a multi-class classification problem in such a way that it treats each unique label-set as a separate class. Any classifier suitable for solving multi-class classifier can be applied to solve the newly created single target multi-class problem. It has only one hyperparameter, the base multi-class classifier. An advantage of the method is that preserves the label relationships. The limitations of this method are that it can not predict novel label combinations and is prone to underfitting when the number of unique label-sets is large.

The method of Pruned Sets \textbf{(PSt)} \cite{Read2008} aims at reducing the number of unique classes (label-sets) appearing when a multi-label problem is approached with LP. To achieve this goal, the method has two phases. The first step is the so-called pruning step. Pruning step removes the infrequently occurring label-sets from the training data. The decision on what means infrequent label-set is a hyperparameter of the method. The second phase consists of introducing the removed examples into the training set. It is done by subsampling the label-sets of the infrequent samples for label subsets which satisfy the pruning criterion. The method introduces this as a tuning hyperparameter and it defines the maximal number of frequent label-sets to subsample from the infrequent label-sets. On such a newly created dataset, LP is trained. Additional improvement of the method is the introduction of a threshold function that enables new label combination to be created at prediction time \cite{Read2010}. In total, there are three hyperparameters for the method: the base multi-class classifier, the pruning value (if the count of the label-sets in the datasets exceeds this number the example is preserved) and the maximal number of frequent label-sets to be reintroduced. An advantage of this method is its efficiency - it is much faster than Label Powerset. A limitation of the metod is that the assumptions can break and the method (without threshold parameter) is not able to introduce novel multi-label sets that have not been seen in the training data.

\subsubsection{Ensembles of problem transformation methods}

This category of problem transformation methods contains all the methods that utilize ensemble techniques such as stacking, bagging, random sub-spacing or employ different transformations of the datasets, such as embeddings.

Conditional Dependency Network \textbf{(CDN)} \cite{Guo2011} aims at encapsulating the dependencies between the labels using dependency networks. Dependency networks are cyclic directed graphical models, where the parents of each variable are its Markov blanket \cite{Heckerman2001}. Markov blanket in the graphical model literature represents a set of nodes around a specific node that shield it. The Markov blanket of a node is the only knowledge needed to predict the behaviour of that node and its children \cite{Pearl1988}. The label dependency information is encoded into the graphical model parameters - the conditional probabilistic distribution associated with each label. The probabilistic distributions are modelled via simple binary classifier models, that as input take the whole feature space augmented by all other labels, with the exclusion of the label being modeled. In the inference phase, it uses the standard model for inference in graphical models - the Gibbs sampling method. This method assumes that one of the labels can change, assuming that all others are fixed. First, random ordering of the labels is chosen and each label is initialized to some value. In each sampling iteration, all the nodes modeling the labels are visited and the new value of the label being modelled is re-sampled according to the probability model that represents the current label being predicted. It has 3 hyperparameters to tune, the model trained at each node of the network, the number of iterations to perform until achieving stationarity of the chain the network and the \textit{burnin} number of operation. This model preserves the label dependencies, however, if there are many numbers of labels the inference performed by the Gibbs sampling method,  may need larger time for convergence and stationarity may not be achieved. 

Meta Binary Relevance \textbf{(MBR)} \cite{Tsoumakas2009}, also known as the 2BR method, consists of two consecutive stages of applying BR. First, $\mathcal{|L|}$ binary base models are built. At the second (meta) stage, the feature space is augmented with the predictions from the first stage ($\mathcal{|L|}$ features are added). New $\mathcal{|L|}$ binary models are trained as in BR. There exist few approaches to generated the predictions in the first stage. The predictions can be generated using the full training set, via $k$ fold cross-validation or by ignoring the irrelevant variables into the meta-level. The cross-validation approach is slow since it requires training of each model at the first level $k$ times, but it passes non-biased information to the meta-level. The irrelevant information can be filtered using the $\Phi$ correlation coefficient to determine if two labels are correlated or not. If they are not correlated, the label is not introduced into the meta-level. Moyano et. al. \cite{Moyano2018} show that the version of this method where the full training set is used at the first stage is better regarding the other two. To further reduce the bias towards the label being predicted, Cherman et al. \cite{Cherman2012} suggest reducing the number of meta-labels to $|L|-1$ (the label being predicted is excluded). This method is known as \textbf{BR+}. MBR has one hyperparameter to tune, the single target base method. A limitation is that MBR and its variants inherit the drawbacks of BR, being not suitable for a large number of labels, due to the large time needed to learn the model.

Ensembles of Classifier Chains \textbf{(ECC)} \cite{Read2011} creates an ensemble of CC build on sampled instance from the original dataset. The sampling is done with replacement. In \cite{Read2011} is argued that sampling with replacement provides better results compared with sampling without replacement. Choosing the percentage of the data for building the models (bag size) is allowed. So the hyperparameters of the method are the number of CC models in the meta architecture and the bag size. In this method also a random ordering of the chain is considered to provide compensation for introducing non-existence dependency between labels. Using the different random subspace of the training set and utilizing different ordering in the chain introduce diversity in the meta architecture. This method takes into account the label correlation, but has a drawback of the large time for training.

Ensembles of Binary Relevance \textbf{(EBR)} \cite{Read2011} build ensemble of BR as base learner. The sampling is done with replacement. The hyperparameter of the method is the number of BR models in the meta architecture. Although it can provide novel label-sets at prediction, it still has the assumption of labels independence. It can be treated as a binary relevance approach with bagging as a meta learner.

Chi-dep \cite{Chekina2009} is a multi-label method that is based on the identification of label dependencies using statistical tests between the labels. $\chi^2$ statistical test for independence for each possible combination of two labels is used. It first tries to identify groups of dependent and independent labels. After their identification, the BR approach for the independent groups, and LP for the dependent labels are trained. At prediction time, the sample is processed by each of the models and the prediction is generated accorfdingly. This method provides a trade-off between the assumption of independence of the labels of the BR method, and the problem of a large number of unique label-sets the LP method is facing. The hyperparameters of the method are the base learners for BR and LP method and the selection of a confidence level for the test. The positive aspects of the method is that it provides a trade-off between the high bias and variance of the BR method, and the low bias and high variance of the LP method.

Ensemble of Chi-dep \textbf{(ECD)} \cite{Chekina2010} builds several Chi-dep models. First, it generates a large number of possible label-sets partitions at random. Each of the partitions is represented by normalized $\chi^2$ score of all the label pairs inside the partition, based on the inside pairwise $\chi^2$ scores. Then, the top \textit{m} distinct sets with the highest scores are included in the meta architecture. The hyperparameters of the method are the number of meta architecture members and the number partitions to evaluate. The positive aspects of the method are that it can further reduce the variance of a single Chi-dep method, however, it suffers from large time complexity. Thus a fast base learning method is recommended.

Ensembles of Label Powersets \textbf{(ELP)}  \cite{Moyano2018} create an ensemble of LP method on sampled prototypes from the original set. The sampling is done with replacement. The hyperparameters of the method include the number of LP models build in the meta architecture and the type of base models. It provides an opportunity to enable LP to predict unseen label combinations (through the ensemble voting), however, it inherits its large computational complexity, and it is practically inefficient for datasets with a large number of unique label-sets.

Ensemble of Pruned Sets \textbf{(EPS)} \cite{Read2008} creates a meta architecture of MLC methods of the PSt method on sampled prototypes from the original set. The sampling is done without replacement with a specific percentage of the dataset being sampled. Additional parameters of the method are the number of members of the meta architecture as well as the number of examples to be sampled from the training size (bag size). This method can predict novel label-sets, thus diminishing one of the disadvantages of a standalone PSt method without thresholding.  Its disadvantage is that it is not able to perform well when there are many diverse label-sets without frequent reoccurring of some of the label-sets. This is due to reducing the training set to a handful training example due to pruning strategy.


Random k Labelsets \textbf{(RAkEL)} \cite{Tsoumakas2011} is an architecture of MLC methods. It uses multiple LP models trained on random partitions of the label space. Usually, the size of the label set is small. Each of the LP methods should learn $2^{k}$ classes instead of $2^{\mathcal{|L|}}$, where $k<<\mathcal{|L|}$. Moreover, the resulting multi-class problems have a much better-balanced distribution of the classes. In \cite{Tsoumakas2011}, two versions of the method are introduced. The first version does not allow for overlap between the groups when creating the label-sets and is called RAkEL disjoint. The second version allows for overlapping between the labels in the created label-sets. This gives the advantage for the same label to be included by the different LP models. The predictions are obtained by voting. Further improvements of the method are proposed in \cite{Rokach2014}. They propose using the classification confidence intervals instead of voting. However, in \cite{Moyano2018} is showed that voting versions of RAkEL achieve better results. There are three hyperparameters to be tuned: the size of label-sets $k$ and the number of models $m$, as well as the base method. The underlying base method can be either LP or PSt. The positive aspect of RAkEL is that uses a smaller number of classifiers than BR and can provide better generalization and is not underfitting as LP. However, it does not scale well in time, as the number of labels and number of instances increases.

Hierarchy of Multi-label Classifiers \textbf{(HOMER)} \cite{Tsoumakas2008} is a meta architecture based on the transformation of the problem into a tree-shaped hierarchy of simpler, better-balanced, MLC problems, utilizing the divide and conquer strategy. The tree is constructed in such a manner that, at the leaves, there are the singleton labels, while the internal nodes represent joint label-sets. A node will contain a training sample iff the sample is annotated with at least one of the labels of the label-set contained in a node. The method consists of two phases: first, the tree is built such that labels from the parent node are distributed to the children nodes using balanced clustering algorithm. Second, the multi-label model is trained on a reduced label-subset, and the process is repeated until all nodes are with one label. Such an approach provides the opportunity to cluster dependent labels into a single node. The hyperparameters of the method are the number of children for a parent node (number of clusters) and the base learner.  It is predominantly useful in tasks with a large number of labels where it is shown to has the best predictive performance \cite{Madjarov2012}. However, the constructed hierarchy is not utilized in problems with a smaller number of labels, hence this method does not show its full potential on datasets with such property \cite{Moyano2018}.

Random Subspace \textbf{(RS)} multi-label method is an extension of the Random Subspace methodology for single target prediction \cite{Ho1998} into the area of MLC. It works with a random sampling of the features. Additionally one can subsample the instances from the training set. For each of the subsamples generated alongside the two dimensions of features and instances, either problem transformation or algorithm adaptation method can be used. There are four hyperparameters to tune: the percentage of the attribute space to be used, the percentage of sample space to be used, the number of models in the ensemble to be built and the multi-label classifier at the base level. This ensemble method is usually faster than bagging and other ensemble methods likewise, depending on the used base multi-label learners.

The AdaBoost \textbf{(AdaBoost, AdaBoost.MH)} \cite{Schapire2000} method is introducing a set of weights maintained both on the examples (as in classical AdaBoost method \cite{Schapire1997}) and the labels. The formula for calculating the weights incorporates the example-label pairs that are miss-classified by the base classifier. At each iteration, the method builds a simple classifier (e.g., s decision stump -- a decision tree of depth 1). The classifier uses weights to focus more on the examples that are hard to predict. The base classifier should provide confidences, that are used to obtain a prediction. The final prediction is obtained by combining the confidences of each of the base models, weighted by the corresponding model weights. The parameter of the method is the number of boosted decision trees. This method is the same as applying AdaBoost to $\mathcal{|L|}$ binary datasets as in BR \cite{Schapire2000, Moyano2018}.

Cost-Sensitive Multi-label Embedding \textbf{(CLEMS)} \cite{Huang2017} belongs to a special type of family of multi-label methods, known as Label Embedding methods. In general, these methods try to embed the label-space into a particular number of dimensions using some embedding technique. It is assumed that the embedded space represents a latent structure of the labels. For learning, either problem transformation or algorithm adaptation method is applied to the augmented feature space. At prediction time, embedding methods employ regression technique to predict the value of the embedded features. One type of label embedding method is known as Cost-Sensitive Embedding. It considers the performance criteria being optimized, as a parameter. In particular, the method considered here employs weighted multidimensional scaling as embedding technique \cite{Kruskal1964}. It embeds cost-matrix of unique label combinations. The cost matrix contains the cost of mistaking a given label combination for another. The hyperparameters of CLEMS are the performance (cost) function, underlying MLC method, the regression method used to predict the values of the embedding features and the number of embedding dimensions. The most effective value for the number of embedding dimensions is the number of labels. The positive aspect of the method is that it can provide good results for a specific cost function being optimized. On the negative side, this method is dependent on the underlying MLC method and requires building a specific model for each cost function for optimal performance per measure.

Triple Random Ensemble \textbf{(TREMLC)} \cite{Tsoumakas2010b} is a meta architecture for MLC. It is a combination of 3 strategies: a sampling of the instance space, sampling of the feature space and sampling of the target space. It is in essence combination of Random Forest built with RAkEL as base classification method. The parameters of the method are bag size, the number of features to subsample, the size of label-sets and the number of models to be built. Its drawback is that it inherits the large computational time of the RAkEL method, however, it scales better regarding the number of instances and features since the Random Forest method reduces the instance and label space.

Subset Mapper \textbf{(SM)} \cite{Schapire1997} uses Hamming distance to make mapping between the output of a multi-label classifier and a known label combination seen in the training set. From the predicted probability distribution, SM will produce a label subset and will calculate the hamming distance to the labels of the training instances. The new test sample as prediction will take the labelset that resulted in the smallest distance. The parameter of this method is the base learning MLC method. This method does not generalize beyond the labelsets seen in the training set.

\subsection{Algorithm Adaptation Methods}

\begin{figure*}[htb]
	\centering
	\includegraphics[width=1.0\textwidth]{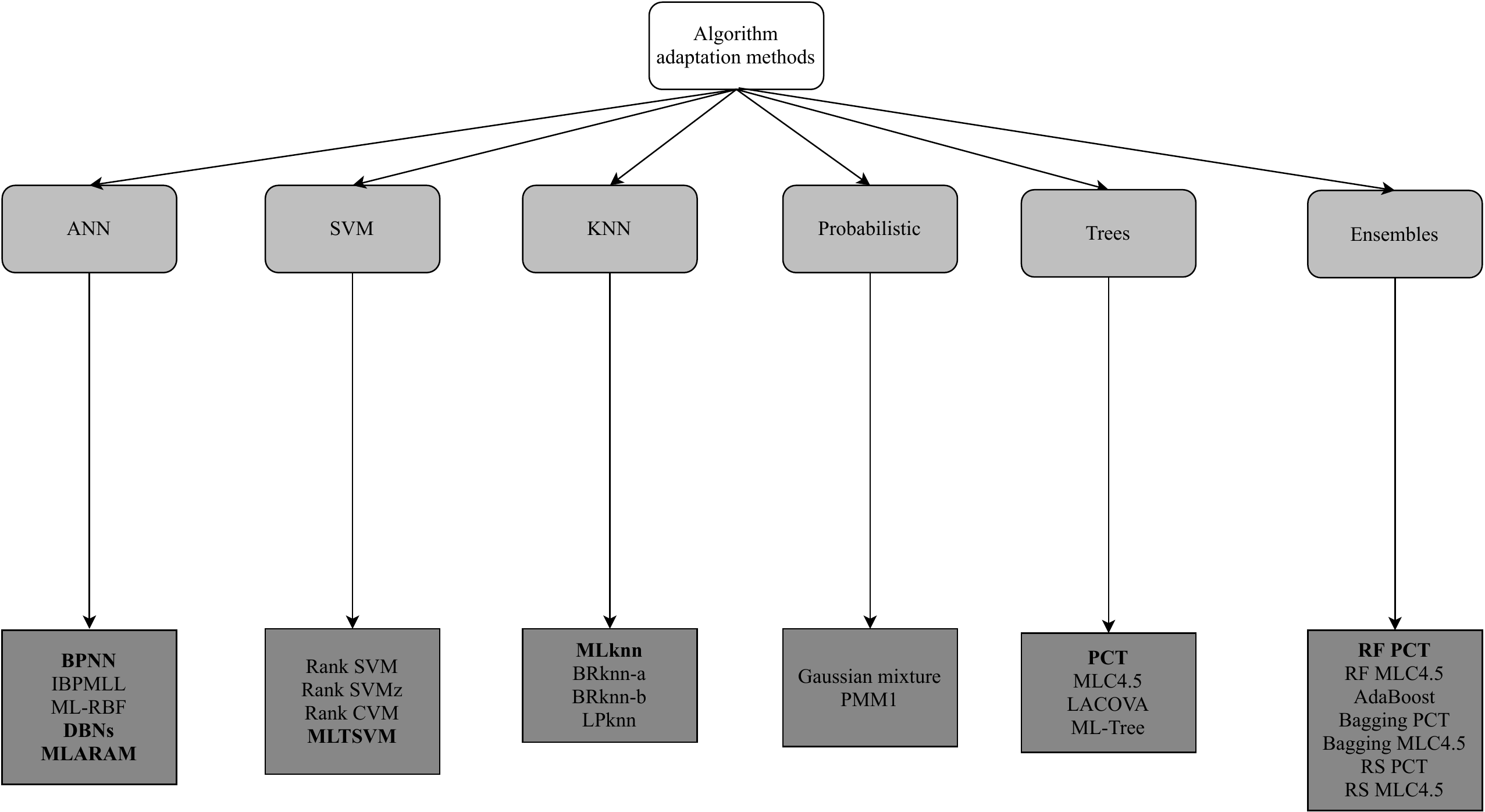}
	\caption{Algorithm adaptation methods. The methods used in this study are shown in bolded.}
	\label{fig:AAA}
\end{figure*}

The adjustments of the underlying algorithm of the single target methods is the core idea on which the algorithm adaptation methods are built. For example, the adjustments for the trees to handle the multi-label problem are two-fold. First, the nodes in the trees are allowed to predict multiple labels at once. Second, the splits are generated by adjusting the impurity measure to take into account the membership and non-membership of a label in the set of relevant and irelevant labels for the samples. This two adjustment allows for each of the labels to have its contribution when creating the splits and obtaining the predictions. Neural networks are inherently designed to tackle multiple targets simultaneously. This is usually done by allowing each of the output neurons to generate score estimates from 0 to 1 in the output neurons. Instance-based such as kNN (k Nearest Neighbours) based methods can also be used for MLC by design: the search of nearest neighbors is in the descriptive features space, the only difference is the calculation of the prediction.
For example, Ml-KNN uses Bayesian posterior probability for estimation of the scores. Support vector machine-based methods use SVM principles when building the model, often with modified cost function to optimize. Probabilistic models, in general, try to use the Bayes formula or mixture models in a multi-label scenario. \figurename~\ref{fig:AAA} depicts the algorithm adaptation methods used in this study. 

\subsubsection{Singleton algorithm adaptation methods}

Predictive Clustering Trees \textbf{(PCTs)} \cite{Blockeel1998} are decision trees viewing the data as a hierarchy of clusters. This method uses standard TDITD algorithm for induction of the tree \cite{Quinlan1986}. At the top node, all data samples belong to the same cluster. This cluster is recursively partitioned into smaller clusters, such that the variance (impurity measure) is reduced. The variance function and the prototype function are selected for the task at hand. In the case of MLC, the variance function is computed as the sum of the Gini indices of the labels. The prototype function returns a vector of probabilities that a sample is labelled with a particular label. As stopping criteria for growing the tree the F-test is used. That is the only hyperparameter of the model needed to be tuned. The positive aspects of this method include the fast time for training and prediction and it is one of the rear MLC methods that can provide interpretable results. As negative aspects are that single tree may be poor in performance, however, an ensemble of PCT can be powerful learning model. As the trees in single target tasks it suffers from large variance. 

Back-propagation Neural Networks \textbf{(BPNN)} \cite{Zhang2006, Read2014} is a neural network approach to the problem of MLC. It is the standard multi-layer perception method. It uses the back-propagation algorithm to calculate the parameters of the network. The hyperparameters of the method are the learning rate, the number of epochs and the number of hidden units. The positive aspects in general for neural networks is that it can provide good performance if a large number of training samples are available and inherently target multi target problems. They face drawbacks on the time needed for hyperparameter optimization. A popular approach in this family of methods is the stacking multiple layers of hidden units, thus increasing the neural network architecture in depth. This is part of a much broader range of methods referred to as deep-learning. Given its popularity, a separate paragraph is dedicated to one deep-learning approach used to model higher-level features, Restricted Boltzmann Machines \textbf{(RBMs)}.

RBMs is a type of deep-learning method that aims to discover the underlying regularities of the observed data \cite{Salakutinov2006}. A Boltzmann machine can be represented as a fully connected network. The restricted Boltzmann machine additionally has the restriction of connections between neurons in the same layer. Usually, the parameters of the network are learned by minimizing contrastive divergence \cite{Hinton2002}. Stacking of multiple RBMs creates so-called Deep Belief Networks \textbf{(DBNs)}. The standard back-propagation algorithm can be used to fine-tune the parameters of the network in a supervised fashion. Using DBNs one can generate new features like a different representation of the data. Those features can be used as input to any multi-label classifier. It is expected that those features are close representatives of the labels being predicted. The hyperparameters of this method include the same as for the BPNN method and additional two: the number of hidden layers and the output multi-label classifier. The novel representation of the input data provided by DBNs can lead to improved performance, on the cost of increased time and space complexity for training the method \cite{Read2014}.


Multi-label ARAM \textbf{(MLARAM)} network \cite{Sapozhnikova2009} is an extension of Adaptive Resonance Associative Map neural-fuzzy networks. ARAM networks for supervised learning consists of two self-organizing maps sharing the same output neurons. The first self-organizing map tries to encode the input space into prototypes, while simultaneously tries to characterize the prototypes with a mapping encoding the labels. A parameter called vigilance is used to control the specificity of the prototypes. Larger values indicate more specific prototypes \cite{Tan1995}. MLARAM is an extension of ARAM in such a way that it allows flexibility in determining when a particular node is activated, taking into consideration label dependencies. The output predictions may vary due to the order in which training examples are presented. The flexibility of inclusion depends on a threshold parameter. The parameters to be tuned are the vigilance and threshold. The positive aspect is that is fast to train and is useful in text classification of a large volume of data, however since it is based on Adaptive Resonance Theory neural-fuzzy networks it has generalization limitation if too many prototypes are built.

Twin Multi-Label Support Vector Machine \textbf{(MLTSVM)} \cite{Chen2016} tries to fit multiple nonparallel hyperplanes to the data to capture the multi-label information embedded in the data. It follows the Twin SVM concept \cite{Chandra2007}, where (in the binary classification case) one tries to find two nonparallel hyperplanes such that each one is closer to its class, but it is further than the others. At the training phase, this method constructs multiple nonparallel hyperplanes to exploit the multi-label information via solving several quadratic programming problems using fast procedures. The prediction is obtained by calculating the distance of the test sample to the different hyperplanes. The hyperparameters of the method are the threshold above which a label is assigned, empirical risk penalty (determines the trade-off between the loss terms in the loss function) and a regularization parameter. \cite{Chen2016} show that MLTSVM outperforms other SVM-based methods for MLC (based on Hamming loss and ranking evaluation measures). Its advantage is that it is fast to train because of the fast underlying procedures for solving the quadratic problem.

Multi-label k Nearest Neighbor \textbf{(MLkNN)} \cite{Zhang2005} method is an adaptation of the Nearest Neighbor \cite{Ruiz1986} paradigm for multi-label problems. It finds the $k$ nearest neighbours of a given sample as in single target kNN algorithm. It constructs prior and conditional probabilities from the training data and thus can use Bayes formula to obtain the posterior probability for a given label on a given test sample. The parameter of the method is the number of neighbours. It is fast to build, but as a lazy method, obtaining the prediction is a more expensive operation. Thus it may not be suitable in a situation where fast predictions are required.

\subsubsection{Ensemble of algorithm adaptation methods}

Random Forest of Predictive Clustering Trees \textbf{(RFPCT)} \cite{Kocev2011,Madjarov2012} uses the random forest method \cite{Breiman2001} with a PCT as a base learning model. It samples both the instance space (sampling with replacement) and the feature space (at random at each tree node). The parameters of the method are the number of features to be used when building the trees and the number of ensemble members. The positive aspects of the method are that it is fast and can tackle the correlation between the labels inherently. On the negative side it that, RFPCT is not suitable for datasets with large sparse feature vectors. Due to the process of a random selection of attributes, it often can happen that this sparse features will be chosen for building the trees. In such a scenario, the trees will have low predictive performance and that will hurt the overall predictive performance of the ensemble.



$n_{tr}$ denotes the number of training samples; $l$ is the number of labels; $f$ is the number of features; $F_{M}$ is the training time complexity of a base single target classification method (also for multi-class); $F^1_{M}$ is the per instance test time complexity of a base single target classification method (also for multi-class); $F_{R}$ is the training time complexity of a single target regression model; $F^1_{R}$ is the per instance test time complexity of a base single target regression model; $C$ is the training time complexity of multi-label classifier; $C^1$ is the per instance time complexity of a multi-label classifier; $e$ is the number of epochs; $m$ is the number of models in ensembles; $K$ is the complexity of clustering algorithm; $l_s$ is the fraction of label-set; $I$ is the number of iterations; $f^1$ is the subset of features; $O(G(.))$ is the complexity of Gibbs sampling method used for inference; $I_c$ is the number of iterators to obtain; $h$ is the number of hidden units; $k$ is the number of neighbours.

\subsection[Addressing Specific Properties of the Multi-label Classification Task]{Addressing Specific Properties of MLC}

Since there exist multiple targets to predict, the task of MLC is more difficult than binary classification \cite{Herrera2016}. While in binary classification the complexity of the models depends on the number of relevant features and number of samples, the MLC task has an additional complexity along the target dimension. These issues present specific challenges for and influence the development of methods for MLC: label dependencies and high dimensional multi-label space.

\subsubsection{Label dependencies}

Label dependencies has a central position in the definition of the MLC task. It presents the ways in which the labels are related among themselves -- for example, consider labelling an image of a seaside; Given the presence of the label 'sea', it is more probable that it will also be labelled with 'beach' than 'city street'. Exploiting these label dependencies can strongly influence the performance of a given MLC method. In the extreme case of non-existence of such dependencies then the best way approaching MLC is by looking at the task as separate $L$ tasks, i.e., binary relevance. However, in real world applications, typically their is a strong influence of the label dependencies on the performance of the MLC methods.

The most straightforward problem-transformation method, BR (and its corresponding ensemble - EBR) does not exploit the dependence information and it is its most common referenced drawback \cite{Read2011}. To bridge that gap, there are various ways to introduce the dependency information, hence the appearance of different methods such as CC, CDN, MBR, SM, CLEMS and ECC. Binary relevance's counterpart, the LP method, takes into consideration the label dependencies. However, it also considers the nonexistent dependency. Pruned sets method utilizes LP and thus has the same positive aspect and drawback. 

The ensemble of MLC models such as HOMER, where the groups of similar labels are being joined together (in one way or another), exploit the label dependency explicitly. HOMER is regrouping the labels into smaller groups, such that dependent labels belong to closer nodes in the tree. Chi-dep's has a statistical driven built-in mechanism for resolving the dependencies between labels. The ensembles of LP and PSt exploit the label dependencies given their base MLC classification method.  RAkEL and TREMLC provide an opportunity to exploit the dependencies between the labels. Since they are randomly subsampling the label space, it may occur that independent labels are grouped together, thus non-existing dependencies are modeled. Nevertheless, if large number of base models are being built it is expected that these non-existing dependencies will be averaged out. The random sampling of the labels increases the bias of the methods. However, the variance of the methods is decreased with the averaging.

On the other side of the spectrum, most of the algorithm adaptation methods (and their corresponding ensembles) in the names of PCT, RFPCT, BPNN, DBNs, MLARAM and MLTSVMs, have the built-in mechanism to deal with this challenge. The presence of label dependencies in methods that have as hyper-parameter a MLC method is tackled depending on the choice of the particular method. Since AdaBoost can be viewed as applying AdaBoost as base-learner to a BR \cite{Schapire2000}, this method has no mechanism of dealing with dependencies between the labels. MLkNN can not exploit label dependencies since it is similar to applying BR with kNN as base learner.

All in all, different methods have different approaches to how they tackle the dependencies between the labels. Some try to augment the descriptive space, others try to model the dependencies modifying the dataset, others are modifying the learning method. In general, it is expected that adding additional information to the method can help to improve performance. This means that it is somewhat expected for methods considering the label-dependencies to perform better. 

\subsubsection{High-dimensional label space}

The challenge of high-dimensional label space mimics the well-known problem of \textit{curse of dimensionality} \cite{Bellman1954, Guyon2003} appearing in the feature space. The curse of dimensionality references the issues appearing in datasets with many features available for the given data \cite{Kira1992}. Following the same analogy, the large number of labels imposes a problem for the multi-label methods. It influences both, the time needed to obtain a prediction and the performance of the methods \cite{Herrera2016}. It goes without saying that the \textit{curse of dimensionality} in the input space exist in MLC tasks also. We discuss these issues in more details in the remainder of this section. 

The curse of dimensionality in the label space is best illustrated through the computational complexity analysis for both training and testing time provided in \tablename~\ref{cmpComp}.  
We give the complexity analysis as provided by the authors proposing the specific method. If this is not available, then we performed the analysis and the summary of it is given in the table (as reference these methods are marked as [ours]). The training time complexity is the time needed to learn the predictive models, while the testing time complexity is the time needed to make a prediction for a given example.

The notation used in \tablename~\ref{cmpComp} is as follows:
$n_{tr}$ denotes the number of training samples; $l$ is the number of labels; $f$ is the number of features; $F_{M}$ is the training time complexity of a base single target classification method (also for multi-class); $F^1_{M}$ is the per instance test time complexity of a base single target classification method (also for multi-class); $F_{R}$ is the training time complexity of a single target regression model; $F^1_{R}$ is the per instance test time complexity of a base single target regression model; $C$ is the training time complexity of multi-label classifier; $C^1$ is the per instance time complexity of a multi-label classifier; $e$ is the number of epochs; $m$ is the number of models in ensembles; $K$ is the complexity of clustering algorithm; $l_s$ is the fraction of label-set; $I$ is the number of iterations; $f^1$ is the subset of features; $O(G(.))$ is the complexity of Gibbs sampling method used for inference; $I_c$ is the number of iterators to obtain; $h$ is the number of hidden units; $k$ is the number of neighbours.

\begin{table*}[!t]
	\caption[Computational complexity of the multi-label methods.]{Computational complexity of the multi-label methods.}
	\label{cmpComp}
	\centering
	\begin{tabular}{lll}
		\hline
		Method   & Training time complexity                       & Test time complexity                   \\
		\hline
		CLR    \cite{Zhang2014}&  $O((l^{2}+l)F_{M}(n_{tr},f))$                 &  $O((l^{2}+l)F^1_{M}(f))$                    \\
		BR     \cite{Zhang2014}&  $O(lF_{M}(n_{tr}, f))$                        &  $O(lF^1_{M}(f))$                        \\
		CC     \cite{Zhang2014}&  $O(lF_{M}(n_{tr}, f+l))$                      &  $O(lF^1_{M}(f+l))$                      \\
		LP     \cite{Zhang2014}&  $O(F_{M}(n_{tr}, f, 2^{l}))$                  &  $O(F^1_{M}(f, 2^{l}))$                 \\
		PSt     \cite{Read2010} &  $O(F_{M}(n_{tr}, f, 2^{l}))$                  &  $O(F^1_{M}(f, 2^{l}))$                  \\
		CLEMS  [ours] &  \makecell{ $O(lF_{R}(n_{tr}, f)) + O(l^3)$ \\ $+ O(C(f+l, n_{tr}, l))$} & \makecell{$O(lF^1_{R})$ \\ $+ C^1(f+l, n_{tr}, l)$} \\
		CDN    [ours]                   &  $O(lF_{m}(n_{tr}, f+l))$                        &  $O(G(I,I_c,F^1_{M}(f+l))$     \\
		PCT    \cite{Kocev2011}  &  $O(fn_{tr}\log(n_{tr})(\log(n_{tr})+l))$                        & $O(log(n_{tr}))$                            \\
		BPNN   \cite{Zhang2006}  &  $O(((f+1)h+(h+1)l)n_{tr}e)$                   &  $ O((f+1)h+(h+1)l)$                    \\
		MLARAM [ours] & $O(n_{tr}^2(|f|+|L|))$ &   $O(n_{tr}(|f|+|L|))$                                      \\
		MLkNN  \cite{Zhang2005}  &  $O(n^2_{tr}f + n_{tr}lq)$                    &  $O(n_{tr}f + lk)$                     \\
		MLTSVM [ours] &  $O(l(O(n^3) + IO(n_{tr}))$                    &  $O(l)$                                 \\
		MBR    [ours]                   &  $O(lF_{M}(n_{tr}, f) + lF_{M}(n_{tr}, f+l))$  &  $O(lF^1_{M}(f) + lF^1_{M}(f+l))$       \\
		EBR    [ours]                   &  $O(mlF_{M}(n_{tr}, f))$                       &  $O(mlF^1_{M}(f))$                      \\
		ELP    [ours]                   &  $O(mF_{M}(n_{tr}, f, 2^{l}))$                 &  $O(mF^1_{M}(f, 2^{l}))$                \\
		ECC    [ours]   &  $O(mlF_{M}(n_{tr}, f+l))$                     &  $O(mlF^1_{M}(f+l))$                    \\
		EPS    \cite{Read2010}    &  $O(mF_{M}(n_{tr}, f, 2^{l}))$                 &  $O(mF^1_{M}(f, 2^{l}))$                  \\
		HOMER  \cite{Tsoumakas2008}&  $O(K(l)+l)$                                   &  $O(l)$                           \\
		RAkEL  \cite{Tsoumakas2011}&  $O(mC(F_M(n_{tr}, f, 2^{l_{s}})))$    &  $O(mC^1(F^1_{M}(f, 2^{l_{s}})))$                 \\
		ECD    \cite{Chekina2010}&  $O(mC(F_M(n_{tr}, f, l)))$                &  $O(mC^1(F^1_M(f, l)))$             \\
		RFPCT  \cite{Kocev2011}&  $O(mfn_{tr}log(n_{tr}+log(l)))$                 &  $O(mlog(n_{tr}))$                            \\
		RSLP   [ours] &  $O(mC(f^1, n_{tr}, l))$                       &  $O(mC^1(f^1, l))$                       \\  
		AdaBoost \cite{Schapire2000} &  $O(mlF_{m}(n_{tr}, f, l))$                    &  $O(mlF_{m}(n_{tr},f))$              \\
		TREMLC \cite{Tsoumakas2010b} &  $O(mC(f^1, n_{tr}, 2^{l_{s}}))$  &  $O(mC^1(f^1,2^{l_{s}}))$              \\
		DBN  [ours]&   \makecell{ $O(((f+1)h+(h+1)l)n_{tr}e) + $ \\ $ O(F_{M}(n_{tr}, f, l))$} &   \makecell{ $ O((f+1)h+(h+1)l) +$ \\ $ O(F_{M}(n_{tr}, f, l))$  }            \\
		SM  [ours] &   $O(lF_{M}(n_{tr}, f))$   & $O(lF^1_{M}(f))$           \\
		\hline
 	\end{tabular}
\end{table*}

BR and CC scale linearly with the number of labels. 
CLR scales quadratically with the number of labels since it needs to build pairwise base learner models.
The time complexity of LP scales exponentially with the number of labels in the worst case. However, in practice, the number of classifiers is limited to the $min(2^{\mathcal{|L|}}, n_{tr})$, where $n_{tr}$ is the number of training instances. Additionally, this method should take into consideration the applied strategy for solving the multi-class problem by the base learner. If the base learner is SVM and if the applied strategy is one vs one \textit{(OVO)} the complexity scales quadratically with the number of unique label-sets. If the applied strategy is one vs all \textit{(OVA)} the complexity scales linear with the number of unique label-sets. Since PSt utilize LP method, at worst if no label-set is removed from the training set its computational complexity is equivalent to the LP method. However, in practice, it is much faster since depending on the pruning parameter infrequent label-sets are removed. 

The complexity of the architectures for MLC built from BR, CC, PSt and LP preserve the same complexity concerning the labels as their base MLC models. Their complexity differs in the number of base multi-label models being built. However, this is not true for HOMER. HOMER requires splitting the label space into smaller clusters when building the hierarchy, this means that its speed is dependent on the clustering algorithm. In \cite{Tsoumakas2008}, it is shown that balanced k-means method scales linearly with the number of labels. MBR, CDN, ECD, MLTSVM and AdaBoost scale linearly with the number of labels. For RFPCT and PCT, the computational complexity depends on a logarithmic function of the number of labels.

An important aspect of the computational complexity analysis for a specific method is the base learner it uses (and especially the susceptibility of the base learner to the 'curse of dimensionality'). For example, using SVM as a base learner requires calculation of the kernel, which amounts to computational complexity of $O(n_{tr}f^2)$ for $f<n_{tr}$. If the dataset has a large number of samples than the time needed for training the method will be larger: $O(n_{tr}^3)$ if $f\sim n_{tr}$.



The high-dimensional label space influences both problem transformation and algorithm adaptation approaches to MLC, not just in terms of computational complexity but also in terms of making the problem more imbalanced. Namely, the high-dimensional label spaces encountered in real-life datasets are usually also sparse: low label cardinality (average number of labels per example) and low label density (frequency of labels). The sparsity then poses a challenge for both problem transformation and algorithm adaptation approaches as follows. In the former case, in binary relevance and label power set, the simpler classification tasks are imbalanced. Hence, once could resort to specific approaches addressing this issue, thus even more increasing the computational cost. In the latter case, the sparse output spaces could make the learning of predictive models more difficult (for example, see PCTs \cite{Kocev:Journal:2013}, extreme MLC \cite{Jain16:PFAST}). A way to approach this is to embed the sparse space in a more compact space through matrix factorization, deep embedding methods etc (for an example, see \cite{ismis:hmlc}).

\section{Experimental Design}\label{sec:MLCexperimentalDesign}

In this section, we discuss the experimental design. First, we present the experimental methodology adopted for conducting the comprehensive study. Second, we present the details on the specific experimental setup used throughout the experiments. Third, we give the specific parameter instantiations used to execute the experiments. Next, we present the evaluation measures used to access the performance of the methods. Finally, we discuss the statistical evaluation used to analyse the results from the study.

\subsection{Experimental methodology}
\begin{figure}[htb]
	\centering
	\includegraphics[width=0.45\textwidth]{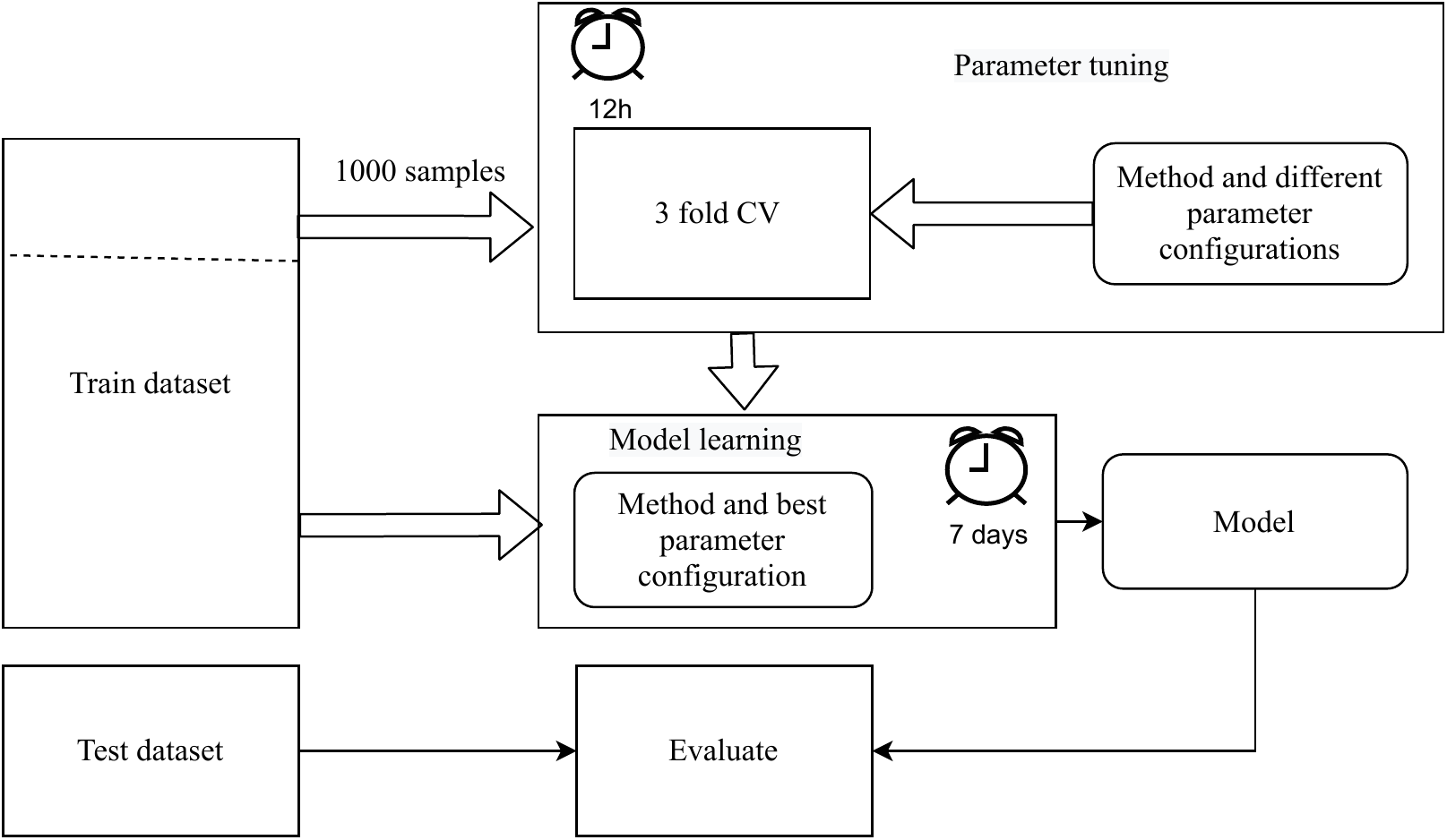}
	\caption{The design of the experimental setup and protocol.}
	\label{fig:experimentalMethodology}
\end{figure}

At the basis of any large scale comprehensive study lies a suitable experimental methodology for hyperparameter optimization of the MLC methods and their base classifiers as well as mesures and procedures for accessing the (predictive) performance of the methods on the datasets. In this study, we adopted and adapted the experimental methodology presented in \cite{Caruana2006}.
\figurename~\ref{fig:experimentalMethodology} depicts the experimental methodology used here and it consists of four stages.

In the first stage, the multi-label datasets considered in this study come in predefined train-test splits. We first sample 1000 examples from the training set using iterative stratification \cite{Sechidis2011} (for the datasets with less than 1000 examples, we take all of the examples). 

In the second stage, on the selected portion of the data we perform 3-fold cross-validation to select the optimal hyper-parameters under a time-budget constraint of 12 hours (similarly as in AutoML \cite{automl}). It means that we allow for evaluation as much as possible (uniformly randomly selected) parameter combinations within the time-budget, and select the best one out of these. 
For each of the methods, we evaluate a multitude of hyperparameter combinations defined with ranges taken from the literature \cite{Tsoumakas2011, Read2011, Madjarov2012, DeSa2018, Moyano2018} (detailed range values for the parameters are given in Sec.~\ref{ref:Section_params}). After the expiration of the time budget or evaluating all of the combinations, whichever comes first, the hyperparameter combination that leads to the smallest Hamming loss is selected as best. If the time-budget did not allow for the evaluation of at least one combination then the literature recommended values are used. 

In the third stage, we learn a predictive model using the complete training set and the selected optimal hyperparameter combination. Also in this stage, we set a time-budget for learning a predictive model to 7 days. In the cases where this occurs the performance of that specific method is marked as \texttt{(DNF)}.


In the fourth and final stage, we evaluate the predictive performance of the method using the test set. The test set is used only to assess the predictive performance of the learned models and it has not been used at any other stage of the experimental evaluation. For the methods that did not yielded a predictive model from the previous stage, their performance was set to the worst possible value for each evaluation measure.


\subsection{Implementation of the experimental methodology}

For undertaking such an extensive experimental study, we needed to implement a multi-platform experimental methodology. 
The implementation of the experimental methodology is done using the Python programming language. It follows the design principles and guidelines of the \texttt{skmultilearn} \cite{Piotor2017} and \texttt{sckit-learn} \cite{Pedregosa2013} ecosystem. We designed a unified experimental methodologyTo include methods and their well-tested implementations from \texttt{CLUS}, \texttt{MULAN} \cite{Tsoumakas2011b} and \texttt{MEKA} \cite{Read2016}, a unified experimental setup was designed. An \texttt{Ubuntu-Xenial} image was built using \texttt{Singularity} \cite{Singulariy} to provide the same experimental conditions for the experiments.

The methods are abstracted into a generic form to provide a unique way of accessing. Using these libraries require specific pre-processing and formatting of the data: For example, \texttt{MULAN} requires \texttt{XML} files storing the names of the labels. After learning the model on a given dataset with a specific method, the predictions (as raw scores) are stored. Next, the raw prediction scores are used as an input to the evaluation measures. The \texttt{sckit-learn} implementation of the measures is used to access the performance of the methods. Since, the one error measure is not implemented in \texttt{sckit-learn}, we implemented it.
Furthermore, we used a wrapper to access the \texttt{MEKA} library the \texttt{skmultilearn}. The wrapper provides a uniform way to obtain the scores, predictions and additional information describing the models. Methods accessed through this library are MBR, BR, LP (LC), PSt, CC, RAkEL, EBR, ELP, ECC, CDN, EPS, BPNN, RSLP, DBPNN, SM, TREMLC. Additionally, the methods CLR and HOMER were accessed via \texttt{MEKA}'s wrapper for \texttt{MULAN}. 
Methods from \texttt{MEKA} that do not provide score estimates are LP, CC and SM. To use \texttt{MULAN}, a suitable wrapper around it was used to access the CDE method.  Next, \texttt{CLUS} was used to access RFPCT and PCT. RFPCT and PCT do provide prediction scores that can be seen as the probability of a given example being labeled with a given label. Finally, the \texttt{skmultilearn} library was used to access MLRAM, MLkNN, CLEMS, AdaBoost, RFDTBR and MLTSVM. MLTSVM does not provide score estimates. It has an internal mechanism for providing predictions. AdaBoost and RFDTBR required implementation of additional functions to retrieve probability estimates.

\subsection{Parameter tuning}\label{ref:Section_params}
The goal of the experimental study is to provide the same conditions for all methods and to provide an opportunity for each method to give its best results. Hence, we need to select the optimal parameters for each of the methods. This is especially relevant for MLC methods that use as base classifiers methods that require tuning (e.g., SVMs). Based on our experimental methodology outlined in \figurename~\ref{fig:experimentalMethodology}, we select the optimal combination of parameters for each method using parameter ranges as defined in the literature. Detailed description of the specific parameter ranges and values evaluated in this study are provided in the Supplementary material.

\subsection{Evaluation Measures}
\begin{figure}[!t]
	\centering
	\includegraphics[width=0.48\textwidth]{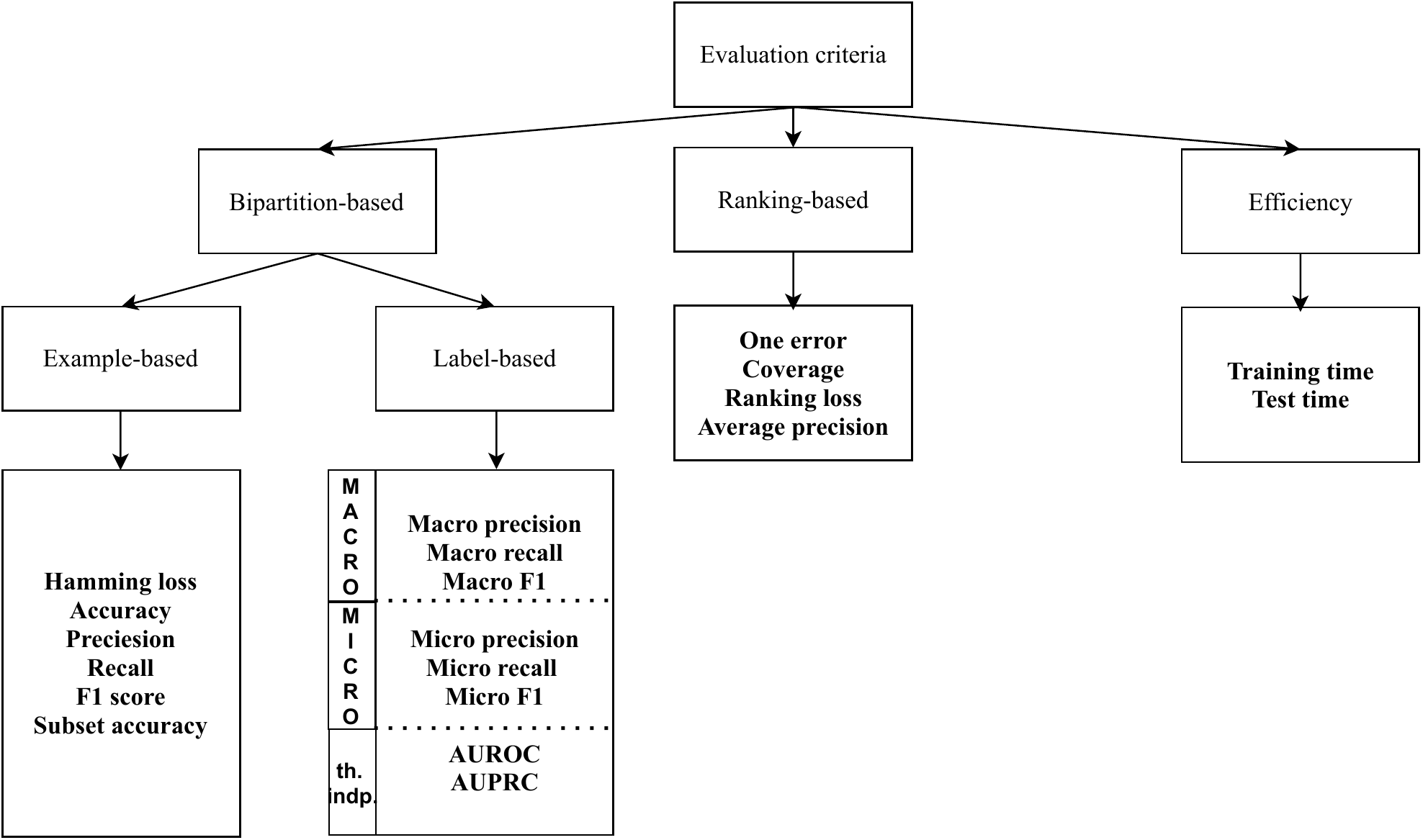}
	\caption{Taxonomy of the evaluation criteria. There are two types of measures, one used to evaluate the performance of the methods and the other one used to evaluate the efficiency of the methods. The performance evaluation can be done either via evaluating bipartitions or via evaluating the relevance of a label for the sample.}
	\label{fig:Measures}
\end{figure}

We use 18 predictive performance and 2 efficiency criteria to evaluate the performance of the methods across the datasets. \figurename~\ref{fig:Measures} depicts a taxonomy of the evaluation measures (or \textit{criteria}, \textit{scores}) \cite{Madjarov2012}. 


For the evaluation of the predictive performance of the methods, \texttt{sckit-learn} implementation of the measures for MLC is used. The evaluation measures requiring score estimates as input are provided with both the calculated scores and the ground truth labels. Since LP, CC, SM and MLTSVM do not generate scores, score-based evaluation criteria are not calculated for them and the predictions as generated by the implementations of the methods are used. 
Most of the methods used in this study return raw scores as predictions. These scores then need to be thresholded in order to obtain the label predictions.
Hence, we use the global \texttt{PCut} thresholding method \cite{Read2011}: It selects a threshold using an iterative procedure such that the label cardinality of the training set is equal to the cardinality of the test set. 
A detailed description of the evaluation measures is given in the Supplementary material as well as in the referenced literature (especially \cite{Madjarov2012}, \cite{Herrera2016}, \cite{Gibaja2014}, \cite{Zhang2014}, \cite{Moyano2018}, \cite{Zhang2018} and \cite{Rivolli2020}).

The assessment of the performance of an MLC method relies on the type of predictions it produces: relevance scores per label or a bipartition. If a method produces relevance scores, special postprocessing techniques can be employed to produce bipartitions \cite{Reem2014}.
In the case of bipartitions, the evaluation measures can be separated into example-based and label-based measures. The latter can then be micro- or macro-averaged, based on the fact whether the joint statistics over all labels are used to calculate the measure or the per label measures are averaged into a single value. 
In the case of providing relevance scores per label, threshold independent or ranking-based measures can be calculated.
The different methods might be more biased towards optimizing a given evaluation measure than other methods. For example, the methods predicting label sets have favourable evaluation using example-based measures, while the methods predicting each label with a different model and combine the predictions have favourable evaluation using macro-averaged measures. Hence, for an unbiased view on the performance of the MLC methods, one needs to consider multiple evaluation measures. 

To evaluate the efficiency of the MLC methods, the training and test time is measured. Training time measures the time needed to learn the predictive model, and the testing time measures the time needed to make predictions for the available test set. While we are aware that the execution times are dependant on the specific implementation of a given method, measuring these times have practical relevance. They provide a glimpse into the time a method needs to produce a model or a prediction, and can serve as a guideline for a practitioner when a decision needs to be made on the use of a specific method. These efficiency estimates should be looked in conjunction with the computational complexity of the methods (as provided in \tablename~\ref{cmpComp}).


\subsection{Statistical Evaluation}

Accessing the overall differences in performance across the datasets to determine if the differences in performance of the methods are statistically significant, we used the corrected Friedman test \cite{Iman1980} and the post-hoc Nemenyi test \cite{Nemenyi1963}. 
Friedman test is a non-parametric multiple hypothesis test \cite{Friedman1940}. It ranks the methods according to their performance for each data separately. Then it calculates the average ranks of the methods and calculates the Friedman statistics. Due to the conservatives of the test the corrected Friedman statistics is preferred.

If statistical significance between the methods exists, the post-hoc Nemanyi test is used to identify the methods with statistically significant differences in performance. The performance of the two methods is statistically significant if their ranks differ more than the critical distance (calculated for a given number of methods, datasets and a significance level). The significant level $\alpha$ is set to 0.05. 


\emph{Limitations of the study.} While having a great practical relevance for detailed depicting the landscape of a learning task, these form of studies have inherit limitations. One of the drawbacks of making large experimental studies relates to that they are computationally expensive. 
This time is linearly dependent with respect to the number of performance criteria one is optimizing. Thus optimization over all of the performance criteria is practically infeasible with lack of appropriate infrastructure. Moreover, the task of making sense out of abundance of results that will emerge is hard.

To overcome this challenge, we usually adopt design choices following recognized literature standards \cite{Caruana2006}. The iterative stratified cross-validation strategy preserves the frequency of the labels. In the study, there are 19 datasets with more then 1000 samples and 7 datasets with more than 5000 labels (2 with more than 9000). The iterative stratified sampling strategy sub-samples the datasets, trying to preserve the frequency of the labelsets \cite{Sechidis2011}. Thus, the potential effect of overfitting to the data is not expected to have noticeable influence over the choice of the best method in the given experimental scenario for all of the datasets. Even if it has, the influence will be small, and will not hurt the drawn conclusions.

We selected Hamming loss \cite{Madjarov2012} since it is analogous  to  error  rate in  single  target classification. It  is  selected  as  evaluation  criteria  for  best  hyper parameter selection since it provides penalization for miss-classification of individual labels. Optimizing for other measures e.g F1, precision and recall (micro, macro and example-based) have inherit bias towards specific paradigms that are correlated with the assumptions done by the families of methods as their bias. Considering threshold independent measures discard methods that cannot produce rankings. Accuracy example-based evaluates just the correctly predicted labels, while the subset accuracy is blind to correct prediction of right and incorrect prediction of wrongly predicted labels. Thus, Hamming loss seems as the most fair-choice for optimization. 

A second constraint when performing large scale study emerges from the included datasets. This is related with the maturity of the task - hence the number of datasets existing in the literature. The conclusions in such study find their validity to hold in the meta-space constrained by the values of the meta-features of the included datasets. For MLC, to the best of our knowledge, this is the greatest amount of datasets and methods being evaluated. Thus we believe that it depicts the current state of the field pointing out guidelines for both practitioners and experts to design and choose the most suitable methods for their MLC problem at hand and further expand the field of MLC as important task in machine learning.

\section{Results and discussion}\label{sec:MLCresults}

This section paints the landscape of MLC methods by providing a discussion on the results from the comprehensive empirical study. The discussion is organized into four parts: (1) comparison of problem transformation methods, (2) comparison of algorithm adaptation methods, (3) analysis of selected best performing methods and (4) computational efficiency analysis of the methods. We focus the discussion on the predictive performance using four evaluation measures (Hamming Loss, F1 example based, Micro precision and AUPRC) thus ensuring the inclusion of all the different groups of measures. The complete results and their detailed analysis are provided in the Supplementary material and at \url{http://mlc.ijs.si}.

\subsection{Problem transformation method comparison}

\begin{figure*}[!ht]
\centering
\subfloat[Hamming loss]{\includegraphics[width=0.5\textwidth]{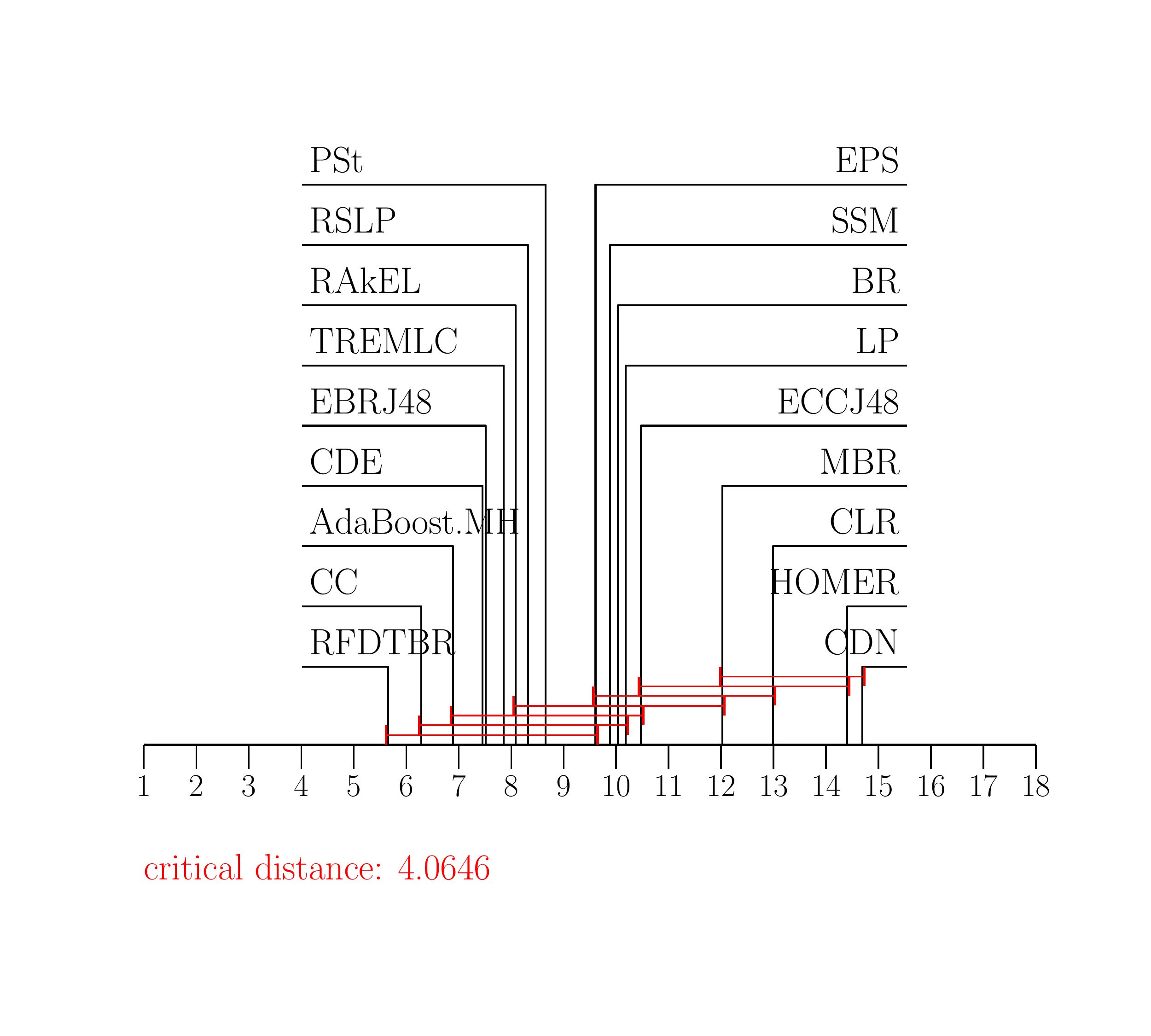}%
\label{fig:PT:HammingLoss}}
\hfil
\subfloat[F1 example based]{\includegraphics[width=0.5\textwidth]{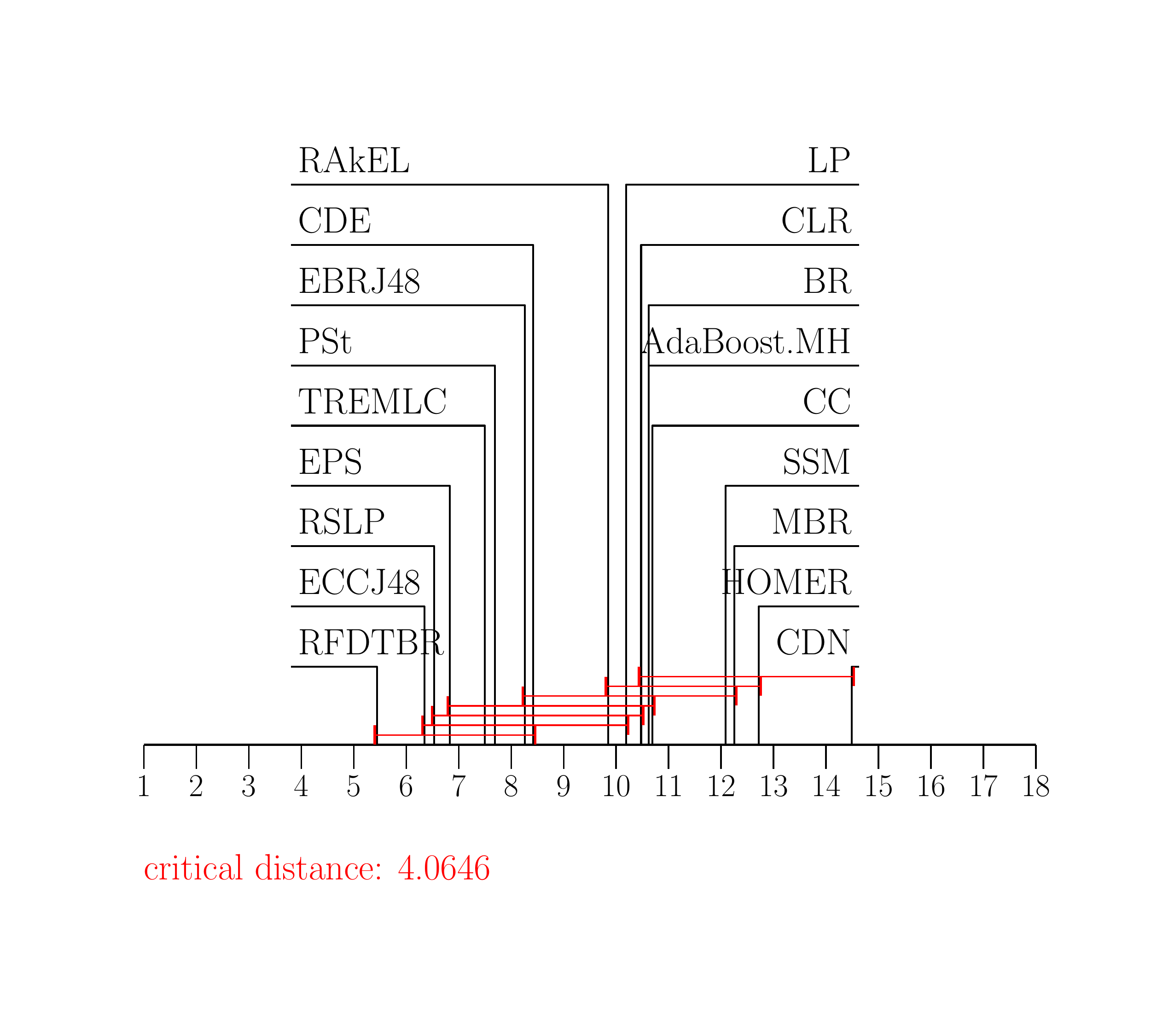}%
\label{fig:PT:F1ExampleBased}}

\subfloat[Micro precision]{\includegraphics[width=0.5\textwidth]{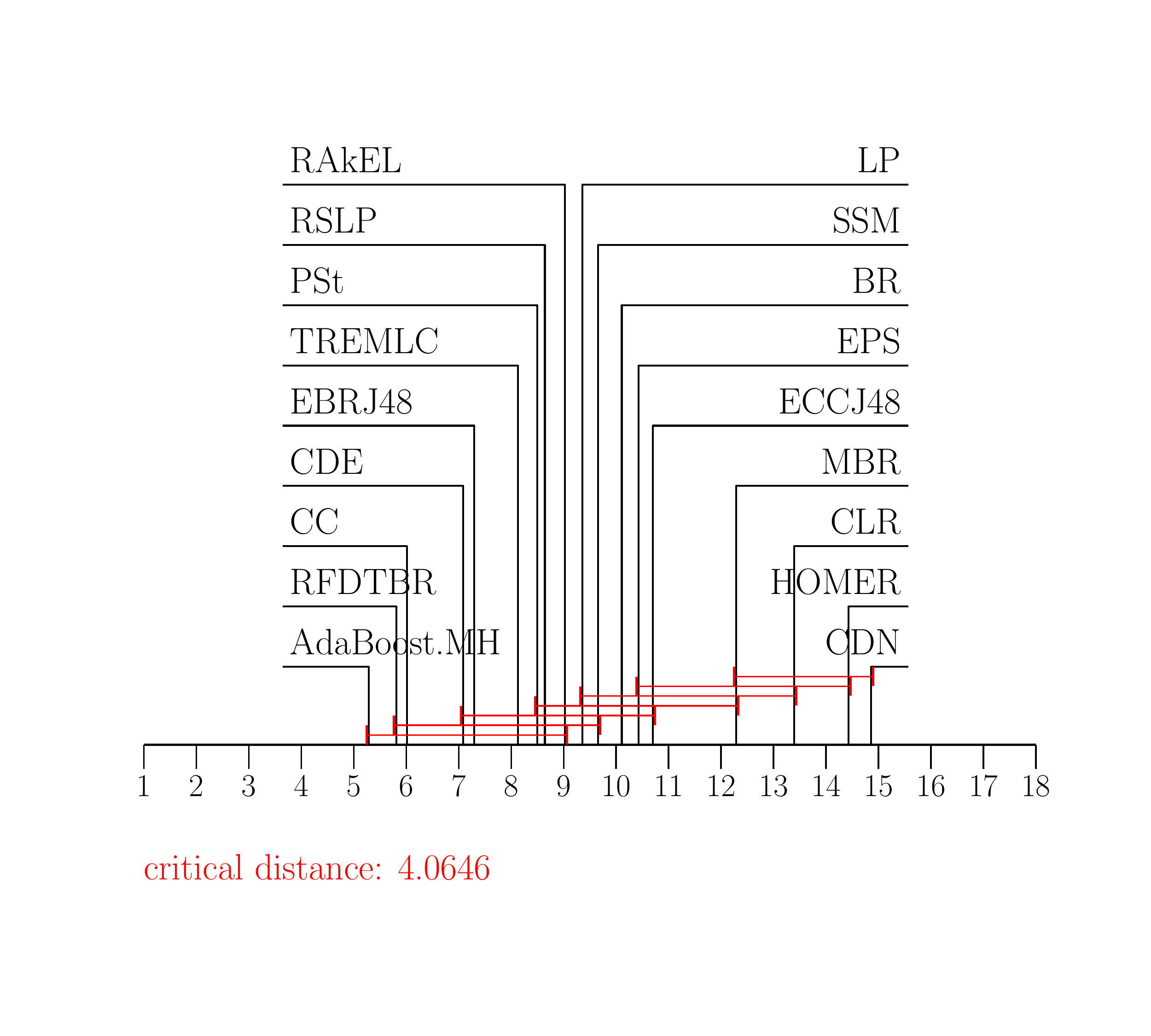}%
\label{fig:PT:MicroPrecision}}
\subfloat[AUPRC]{\includegraphics[width=0.51\textwidth]{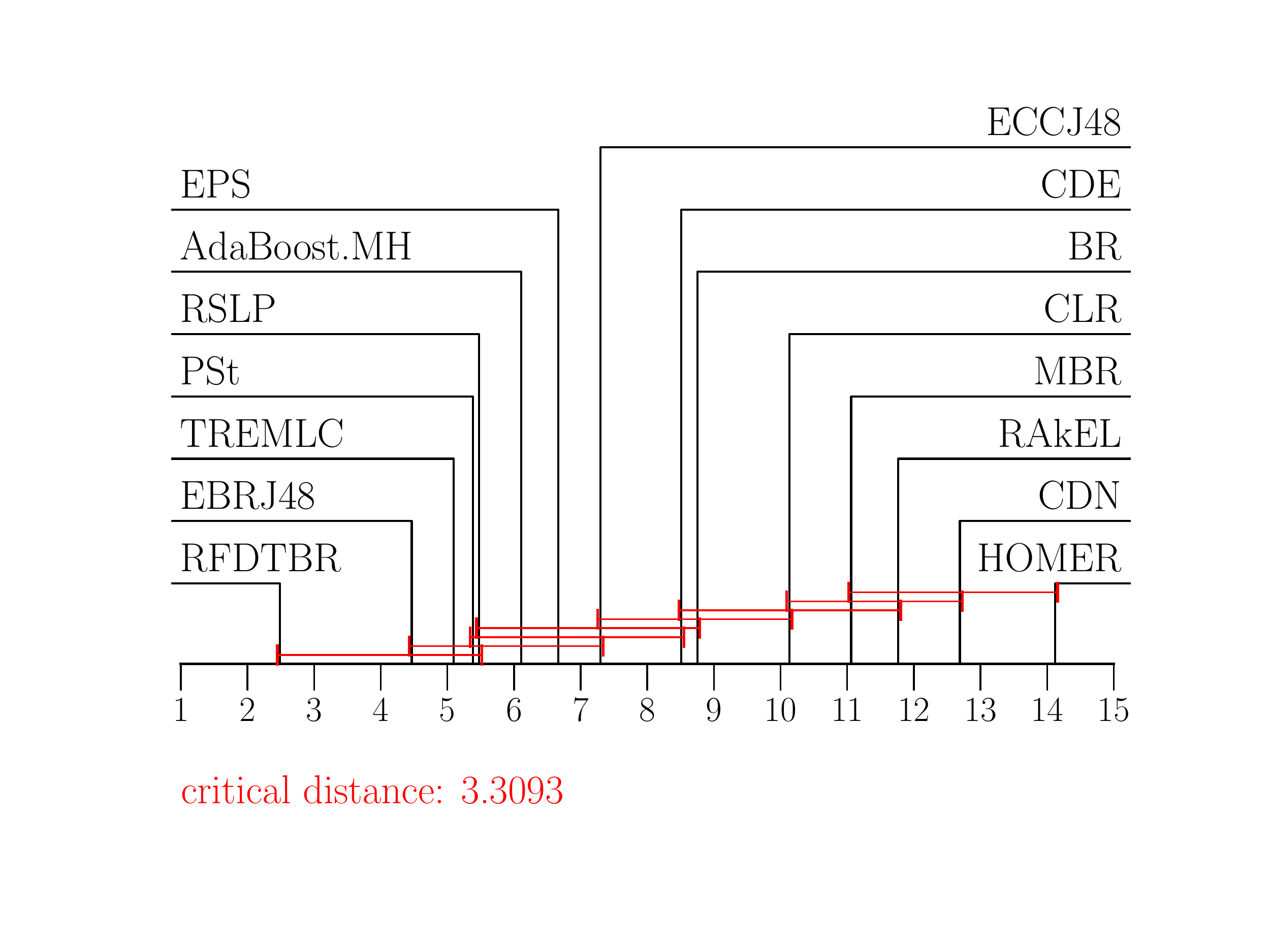}%
\label{fig:PT:AUPRC}}
\caption{Average rank diagrams comparing the predictive performance of problem transformation methods.}
\label{fig:PT}
\end{figure*}

\figurename~\ref{fig:PT} depicts the average rank diagrams comparing the problem transformation methods. At a first glance, we can observe that the problem transformation methods that are utilizing BR outperform all other methods across all evaluation measures. 
More specifically, the best performing method is RFDTBR -- it is the best ranked method on 12 evaluation measures (and second best on 3 more) and is the most efficient in terms of both training and testing time. EBRJ48, AdaBoost.MH, ECC J48, TREMLC and PSt are often among the top ranked methods, while CDN, SM and HOMER are the worst performing methods (CDN is worst ranked on 13 evaluation measures). 
In the remainder of this section, we analyze the performance of all methods in more detail, along the different types of evaluation measures and different subgroups of methods, finally selecting the most promising problem transformation methods. 

We first summarize the performance of the methods according to the different groups of evaluation measures. Considering the example-based evaluation measures, the best performing methods are RFDTBR, ECC J48 and RSLP, while the worst performing methods are MBR, HOMER and CDN. Next, focusing on the label-based evaluation measures, we can make the following observations: i) on the threshold-independent measures (AUCROC and AUPRC) the best performing methods are RFDTBR and EBR J48, while RAkEL, HOMER, and CDN are on the losing end; ii) on the micro averaged measures, the best performing methods are RFDTBR, TREMLC, ECC J48, AdaBoost.MH; and iii) on the macro averaged measures, BR, ECC J48, CLR and AdaBoost.MH perform best. It is interesting to observe the drop of RFDTBR in the ranking based on macro-averaged measures. This suggests that RFDTBR as base learner fails to provide good individual predictions per label, as opposed to BR with SVMs. Furthermore, analyzing the performance on ranking-based measures, RFDTBR, EBRJ48, AdaBoost.MH, and PSt have the best performance, while RAkEL, HOMER, and CDN the worst. Finally, considering efficiency, the best training times are obtained with RFDTBR, HOMER and PSt, and the worst with ECC J48, EBR J48 and AdaBoost.MH, The best testing times are obtained with RFDTBR, CDE, SSM, and the worst with LP, RAkEL, and CLR.

When it comes to using ensemble methods to approach MLC with problem transformation (using the BR, CC or LP approach), it is very important to select the proper base predictive models of those ensembles \cite{Madjarov2012}\cite{Moyano2018}\cite{Rivolli2020}. There are two widely used options concerning the base predictive models: J48 and SVMs. 

We performed an additional experimental study concerning this choice. Namely, we evaluated EBR, ECC and ELP built with J48 trees and SVMs as base predictive models. The results showed that using J48 as base predictive model is generally beneficial in terms of predictive performance: on a large majority of predictive performance measures, the ensembles using J48 are better ranked than their SVM counterparts. Moreover, EBR with J48 is the best ranked method on 14 out of 18 predictive performance measures. The ensembles with SVMs perform better on the macro aggregated evaluation measures and subset accuracy. In terms of efficiency, ensembles with J48 are undoubtedly the preferred choice: they are faster to learn and make predictions than the SVM counterparts. In addition, J48 does not need parameter tuning, while 2 SVMs parameters need to be tuned. A detailed discussion of the evaluation are available in the Supplementary material.

Several observations can be made by comparing the performance of the different ensembles of problem transformation methods (EBR, ECC, ELP). First, it can be observed that EBR tends to perform best according to ranking-based measures, micro-averaged label-based and threshold-independent measures. The predictions of EBR are characterized with good precision scores, hence they provide more exact predictions (i.e., the labels predicted as relevant are truly relevant). Conversely, according to the example-based and macro-averaged label-based measures the ECC method ranks best. ECC has high values for recall on the example-based measures, meaning that its predictions are more complete (i.e., the truly relevant labels are indeed predicted as relevant). EBR performs good on precision and ECC performs good on recall. EBR and ECC models with J48 tend to provide good results (EBR on micro and ranking-based measures and ECC on micro and example-based measures), do not require tuning of parameters and are fast to built. ELP shows the worst performance in general.


LP-based architectures of problem transformation methods have good performance as measured by recall, and consequently accordingly to the F1 measure. More specifically, according to example-based measures, these methods produce complete predictions, where truly relevant labels are indeed predicted as relevant. In contrast, they fall short on precision-based measures, meaning that not all the predicted labels are truly relevant for the examples. Similar observations can be made for the micro-averaged label-based measures, but not for the macro-averaged label-based measures, where LP-based methods suffer reduced performance. Since LP-based methods predict partitions, they can preserve the label-sets. This reflects the good rankings achieved in example-based measures and micro-averaged label-based measures that are calculated by taking the labels jointly, before averaging them. However, when macro-averaged label-based measures are considered, the preservation of the label-sets seems not to be beneficial and the methods are unable to produce sufficient diversity per label. 
These conclusions are further confirmed by analyzing the performance of the methods on the ranking and threshold-independent measures (see \figurename~\ref{fig:PT:AUPRC}). Again, LP-based methods are ranked lower as compared to the BR-based methods.

Comparing the results of LP-based and BR-based singletons utilzing SVMs as base learners, it can be observed that PSt is the best ranked, except for the macro measures. PSt prunes the infrequent label-sets and trains a LP method on the modified dataset. The better ranking of PSt shows that the infrequent label-sets hurt the performance of the LP-based approaches. PSt is superior to its counterpart BR according to the example-based and micro-averaged label-based measures. Comparison of the BR-based and LP-based singletons versus the corresponding architectures shows better performance for the architectures, which most often is significantly large, for the best performing methods.


Based on the above discussion and all of the empirical evidence presented, we select RFDTBR, AdaBoost.MH, ECCJ48, TREMLC, PSt and EBRJ48 as the best performing group of problem transformation methods.

\subsection{Algorithm adaptation method comparison}

\begin{figure*}[!hb]
\centering

\subfloat[Hamming loss]{\includegraphics[width=0.5\textwidth]{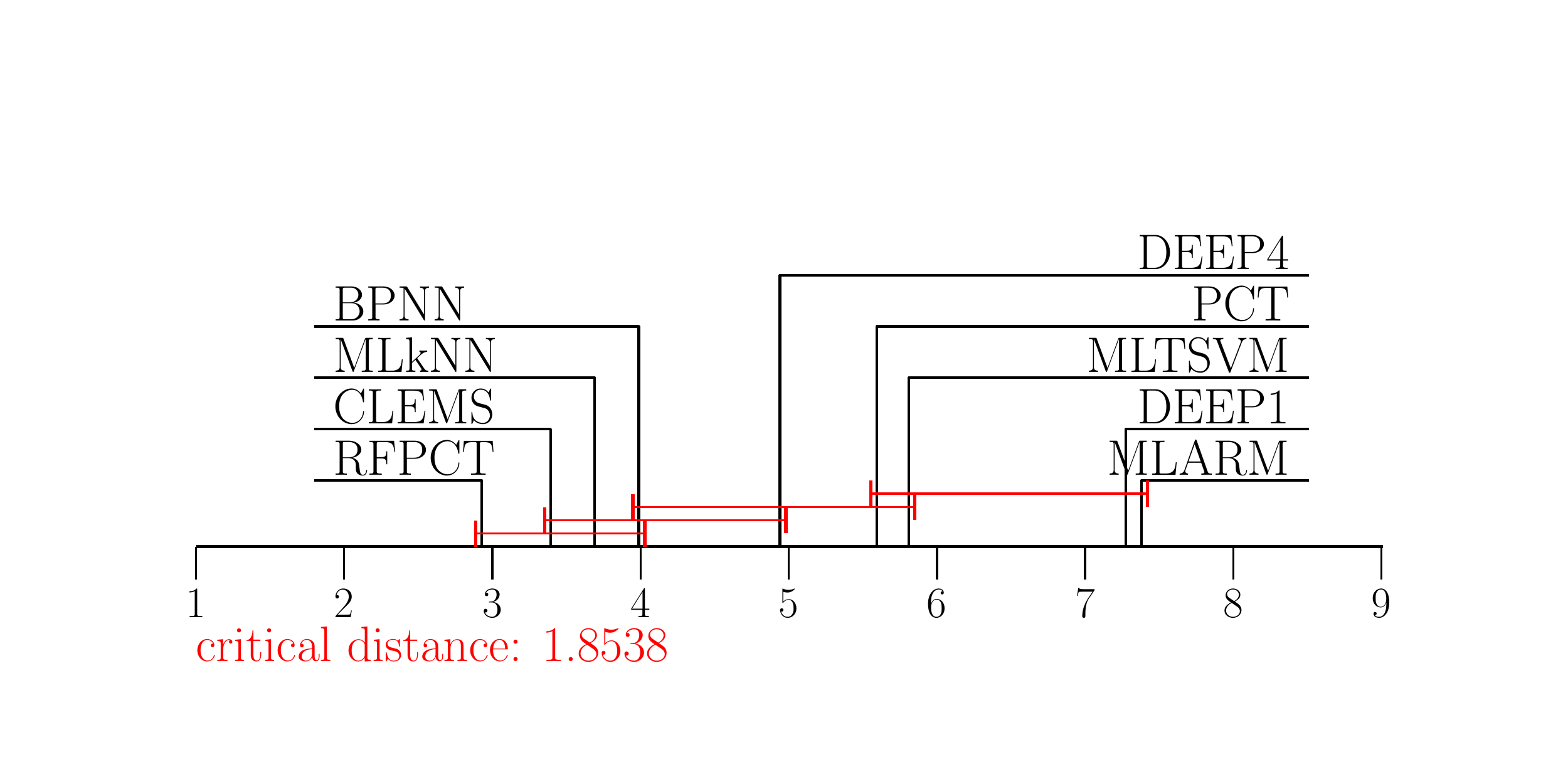}%
\label{fig:AA:HammingLoss}}
\hfil
\subfloat[F1 example based]{\includegraphics[width=0.5\textwidth]{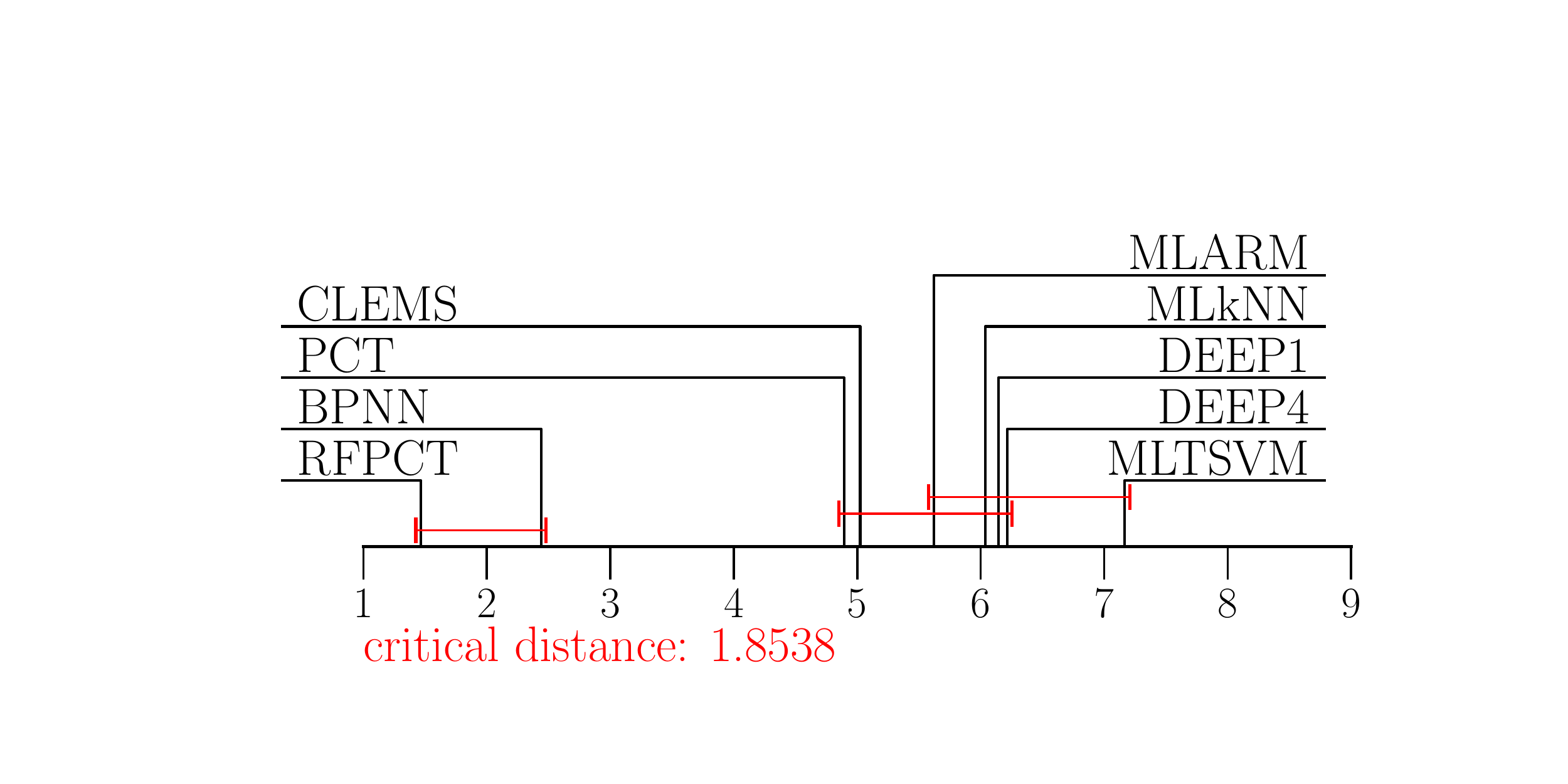}%
\label{fig:AA:F1ExampleBased}}

\subfloat[Micro precision]{\includegraphics[width=0.5\textwidth]{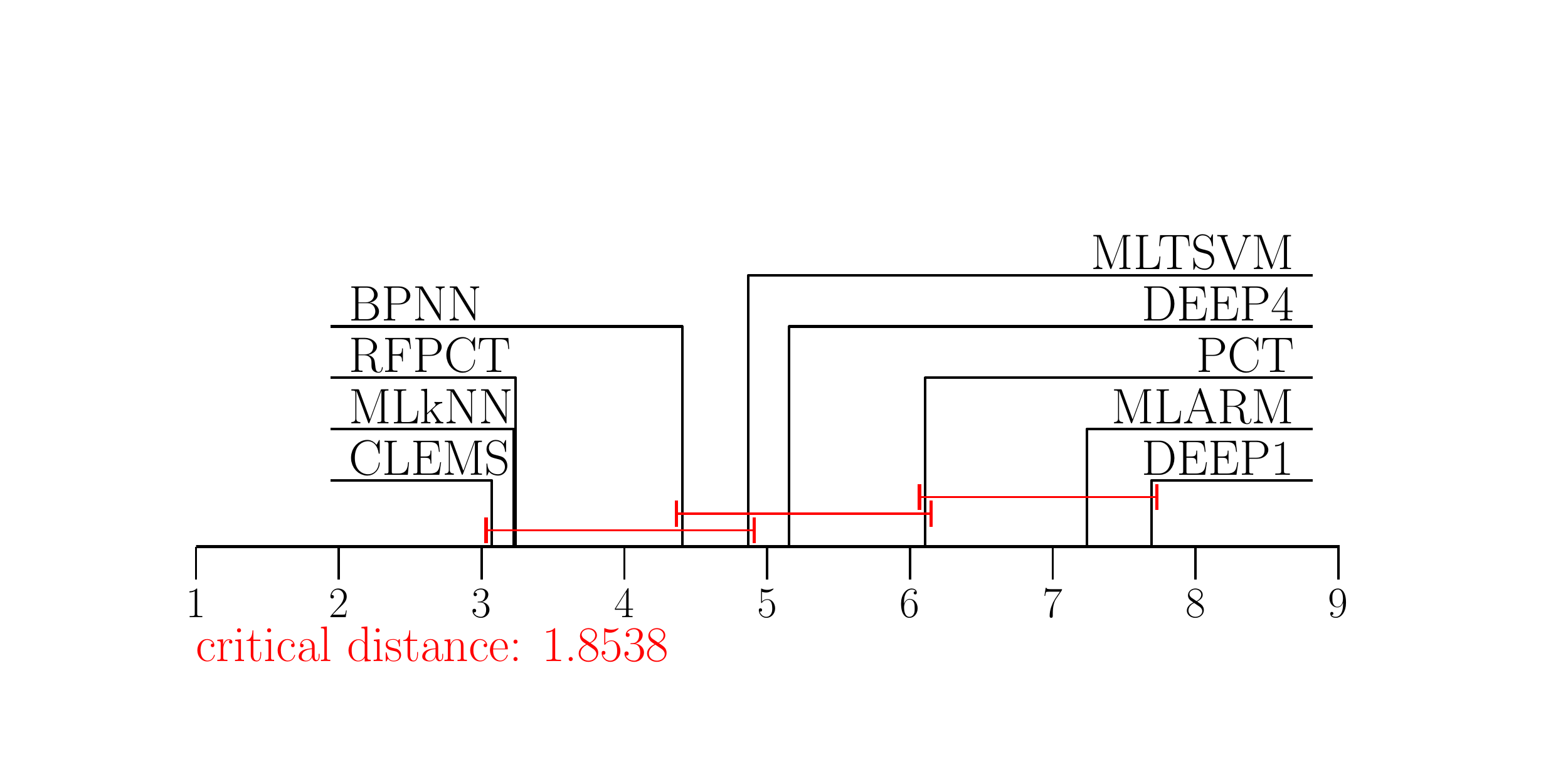}%
\label{fig:AA:MicroPrecision}}
\subfloat[AUPRC]{\includegraphics[width=0.5\textwidth]{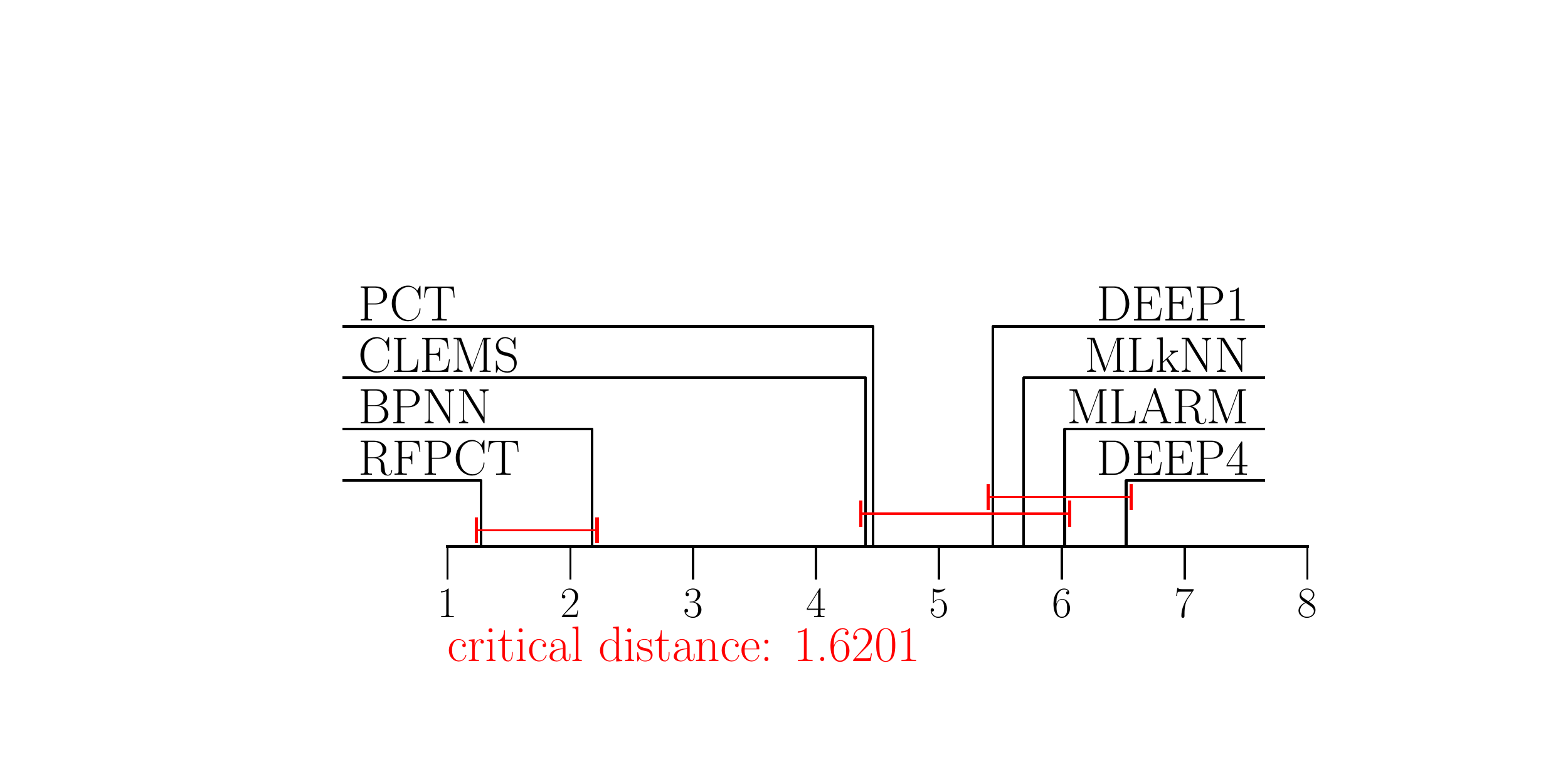}%
\label{fig:AA:AUPRC}}
\caption{Average rank diagrams comparing the performance of algorithm adaptation methods.}
\label{fig:AA}
\end{figure*}

\figurename~\ref{fig:AA} depicts the average rank diagrams for the algorithm adaptation methods. At a first glance, we can make the following observations: i) The best performing method is RFPCT -- it is best ranked according to 17 out of 18 performance evaluation measures; ii) It is closely followed by BPNN -- ranked second according to 16 out of 18 performance evaluation measures, with performance differences to RFPCT that are not statistically significant; and iii) the worst ranked methods are MLTSVM, DEEP 1 and DEEP 4 (DEEP 1 and 4 are the two architectures using DBNs to create a lower dimensional representation of the input space and then using BPNNs or ECC as a second stage MLC classifier).

By inspecting in detail the results across all types of evaluation measures, we find that the above observations hold: RFPCT and BPNN are typically the best ranked methods. Here, we mention the two evaluation measures where this is not the case: Hamming Loss and micro-averaged precision. 
For both evaluation measures, CLEMS and MLkNN achieve good predictive performance (according to micro-averaged precision, CLEMS and MLkNN are the top-ranked methods, see \figurename~\ref{fig:AA:MicroPrecision}). The good performance on precision indicates that these methods are more conservative in assigning relevant labels. Usually, this means a weaker performance on recall-based measures.


Next, we have performed an extensive evaluation of 4 different architectures of deep belief networks (DBNs), as representatives of deep learning methods for MLC. The 4 DBN-based models were trained on the whole training set to increase their chance of preventing over-fitting due to the small number of instances. These 4 architectures are obtained as a Cartesian product of two sets of parameters of the optimizer (learning rate and momentum) and the MLC classifier with fixed parameter at the second stage (BPNN and ECC). 

Detailed analysis of the DBN results, given in the Supplementary material reveal the following. Using ECC as a MLC classifier at the second stage is beneficial according to the example-based and label-based measures, while using BPNN for that purpose is beneficial according to the threshold-independent label-based measures and the ranking-based measures. 
The better ranked architectures for the two different MLC classifiers (DEEP1 and DEEP4) were selected for comparison with the other algorithm adaptation methods.
Still, these two architectures have much worse performance as compared to other methods. This might be due to the fact that the benchmarking datasets are of different sizes and there is a good portion of them with a small number of examples, which makes the DBNs overfit \cite{Read2014}: This prompts for better exploration of the parameter space of DBNs in this context.

Based on the discussion and all of the empirical evidence presented, we select RFPCT and BPNN as the best performing group of algorithm adaptation methods.



\subsection{Selected MLC methods performance comparison}

\begin{figure*}[!hb]
\centering

\subfloat[Hamming loss]{\includegraphics[width=0.45\textwidth,height=0.15\textheight]{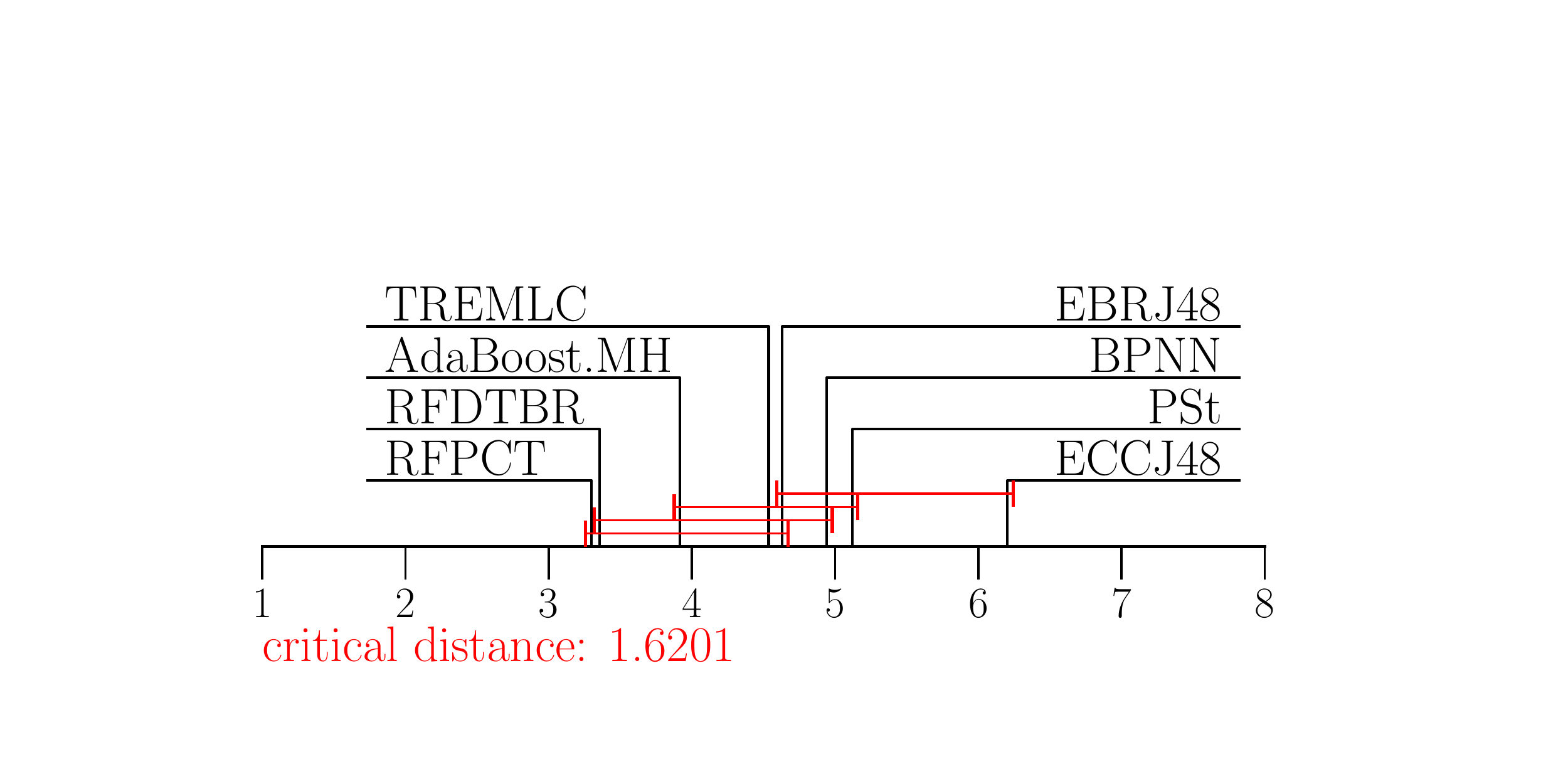}%
\label{fig:BM:HammingLoss}}
\hfil
\subfloat[F1 example based]{\includegraphics[width=0.45\textwidth,height=0.15\textheight]{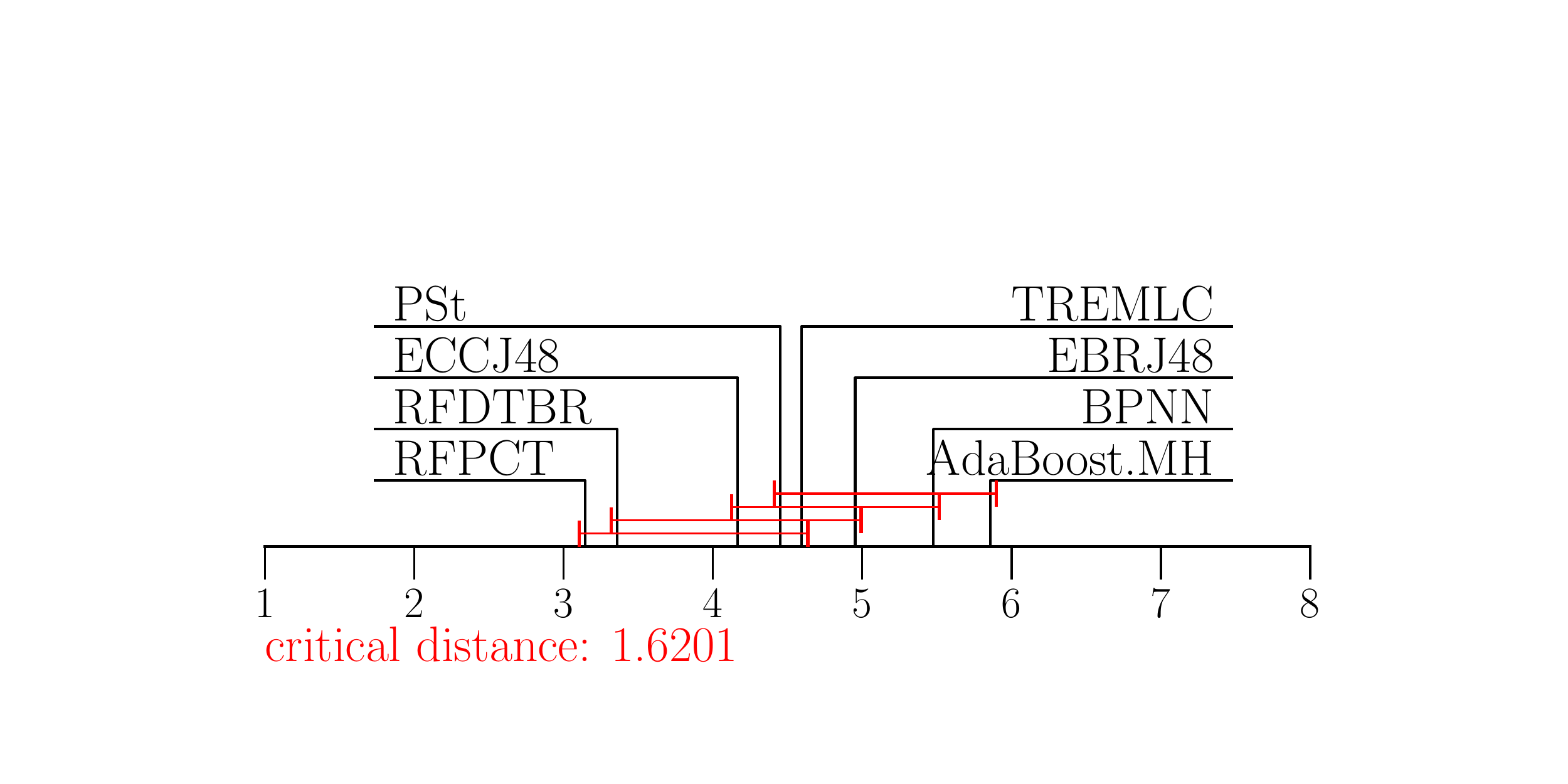}%
\label{fig:BM:F1ExampleBased}}

\subfloat[Micro precision]{\includegraphics[width=0.45\textwidth,height=0.15\textheight]{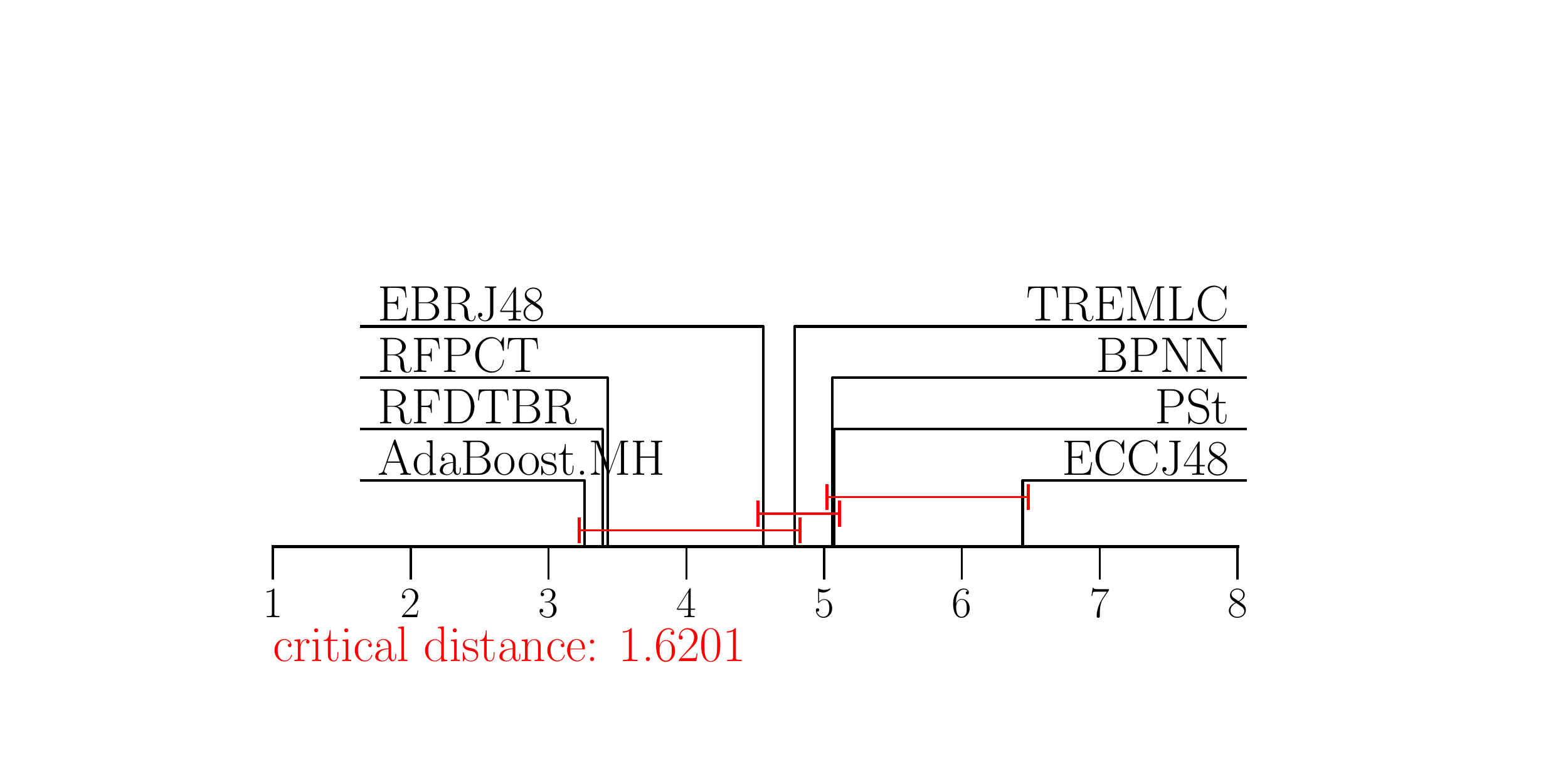}%
\label{fig:BM:F1Micro}}
\hfil
\subfloat[AUPRC]{\includegraphics[width=0.45\textwidth,height=0.15\textheight]{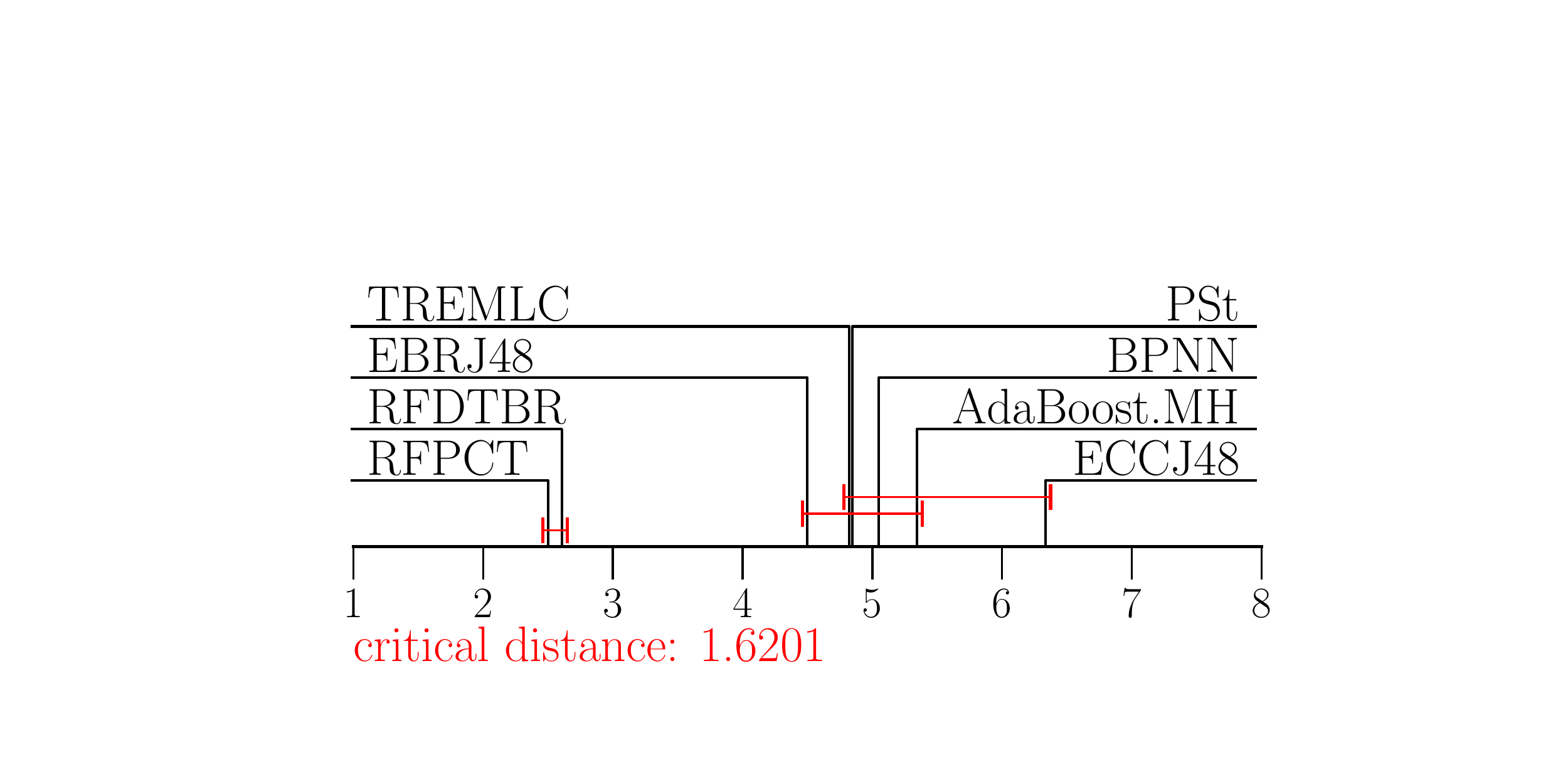}%
\label{fig:BM:F1Macro}}

\caption{Average rank diagrams for the best performing methods from both groups (problem transformation and algorithm adaptation) selected after the per-group analysis of the results. The average ranking diagrams for all evaluation measures are available in the Appendix.}
\label{fig:BM}
\end{figure*}

We further compare the results of the selected best performing methods from both groups, i.e., the problem transformation methods and the algorithm adaptation methods. \figurename~\ref{fig:BM} depicts the results of this comparison. As explained above, we selected 6 problem transformation methods and 2 algorithm adaptation methods. At a glance, the results shown here, as well as the detailed results from the Appendix, clearly identify that tree-based models as the state-of-the-art, especially tree-based ensembles based on random forests. Below, we first discuss the performance of all considered methods along the lines of the different types of evaluation measures, and then drill down to the performance of each selected method.

We start with discussing the results for the example-based evaluation measures (\figurename~\ref{fig:main:example_based}). Here, RFPCT is best ranked according to 4 out of 6 measures and second best on the other two. RFDTBR is best ranked on 1 measure and second best on 4. These two methods are clearly the best performers, except on recall (where ECC J48 is top ranked). This means that the predictions made by RFPCT and RFDTBR assign relevant labels more conservatively. 
Also, on the threshold-independent measures (\figurename~\ref{fig:main:th_independent}), which provide the most holistic view on MLC method performance, RFPCT and RFDTBR are dominant (according to AUPRC, they statistically significantly outperform the competition). 

In terms of the label-based evaluation measures (\figurename~\ref{fig:main:label}), the situation is not as clear. ECCJ48 is the best performing method according to 3 evaluation measures and worst performing according to 2 evaluation measures. Namely, ECCJ48 is strong according to the recall measures (for the macro-averaged recall it even statistically significantly outperforms all competitors). This comes at the price of it being the worst ranked method on precision. On precision-based measures, AdaBoost.MH is the best performing (with RFPCT and RFDTBR following closely), and among the worst performing methods on recall-based measures. Interesting to note here is the difference in performance of RFPCT due to the averaging of the recall: with macro averaging, it is worst ranked, while with micro averaging, it is ranked second best. This indicates that RFPCT focuses on predicting the more frequent labels correctly at the cost of misclassifying the less frequent ones. This is in line with the understanding that the BR-type of methods are more appropriate for macro-averaging of the performance: they try to predict each of the labels separately as well as possible. Conversely, the methods that predict the complete or partial label set (such as LP-based methods and algorithm adaptation methods, incl. RFPCT) are well suited for micro-averaged measures.


Furthermore, in terms of ranking-based measures (\figurename~\ref{fig:main:ranking}), the best performing method is RFDTBR -- it is top-ranked on all 4 evaluation measures, while RFPCT is second best on 3 evaluation measures. These results indicate that by further improving the thresholding method for assessing whether a label is relevant or not, one can expect further improvement of the performance of these methods on the other evaluation measures. Here, the worst performing method is ECC J48 -- it has the worst ranks according to 3 evaluation measures.

We next dig deeper into the performance of each of the selected methods. We start with the random forest approach to learning ensembles for MLC. When used either in local/problem transformation (RFDTBR) or global/algorithm adaptation (RFPCT) context, random forests show the best performance across the example-based, micro-averaged label-based and ranking-based measures, as well as threshold-independent measures. When considering the macro-averaged label-based measures, both of the methods, despite having good rankings on macro precision measures, are not able to predict as relevant all instances where the labels are truly relevant (per label) and thus have worse rankings on macro recall. This leads us to the observation that these methods are rather conservative when deciding whether a label is relevant or not. 



ECCJ48 is best ranked on the recall-based measures, but underperforms on the precision-based measures, where it is often being ranked worst. This indicates that ECCJ48 is rather liberal when assigning labels as relevant, i.e., it indeed truthfully predicts most of the relevant labels, but at the cost of also predicting irrelevant labels as relevant.

In contrast to ECCJ48, AdaBoost.MH shows good results on precision, as compared to the results on recall. The weights over the samples help AdaBoost to be conservative in its predictions. Additionally, it is ranked favorably according to the ranking-based measures (indicating that there is room for further improvement of the scores for the other measures by adjusting the thresholding method). For coverage and ranking loss, it is among the best-ranked methods.

EBRJ48 shows competitive performance on the ranking measures. It closely follows the best-ranked methods and it is often not statistically significantly different from them. On the threshold-independent measures, it is also ranked as the 3-rd best method. However, it suffers on the other measures, even though it is not statistically significantly worse than the best method in 4 out of 12 remaining measures (after excluding the ranking-based and the threshold-independent based ones).

The PSt method has good ranks for the example-based measures -- it is not statistically significantly different from the best ranked method in 4 out of 6 measures. However, on the remaining measures, it has hood ranks for 2 measures - macro F1 and coverage. The rationale behind this behaviour of PSt is that it predicts label-sets, hence it can provide good performance on the per example-based measures. TREMLC, as an LP-based architecture, has better ranks for macro and micro-based measures, as compared to the singleton LP-based method, PSt. However, it has statistically significantly worse ranks than the best ranked methods, similar to PSt. The average rank diagrams show that LP-based methods, in general, are not competitive on ranking-based measures.

BPNN has the weakest ranking across the measures -- it is often statistically significantly different from the best ranked method and it is only ranked as not statistically significantly different to the best method in 2 out of 18 predictive performance measures. These results point out that BPNN is very sensitive to its architectural design and parameter settings. Finding the best configuration for a BPNN, for a given problem is a computationally expensive challenge on its own \cite{NAS}. Moreover, this observations follows the general observation for single target tasks, for tabular data, where ensemble methods have the competitive edge \cite{shap}. 

\subsection{MLC methods efficiency comparison}

\begin{figure}[!t]
\centering
\subfloat[Train time]{\includegraphics[width=0.45\textwidth]{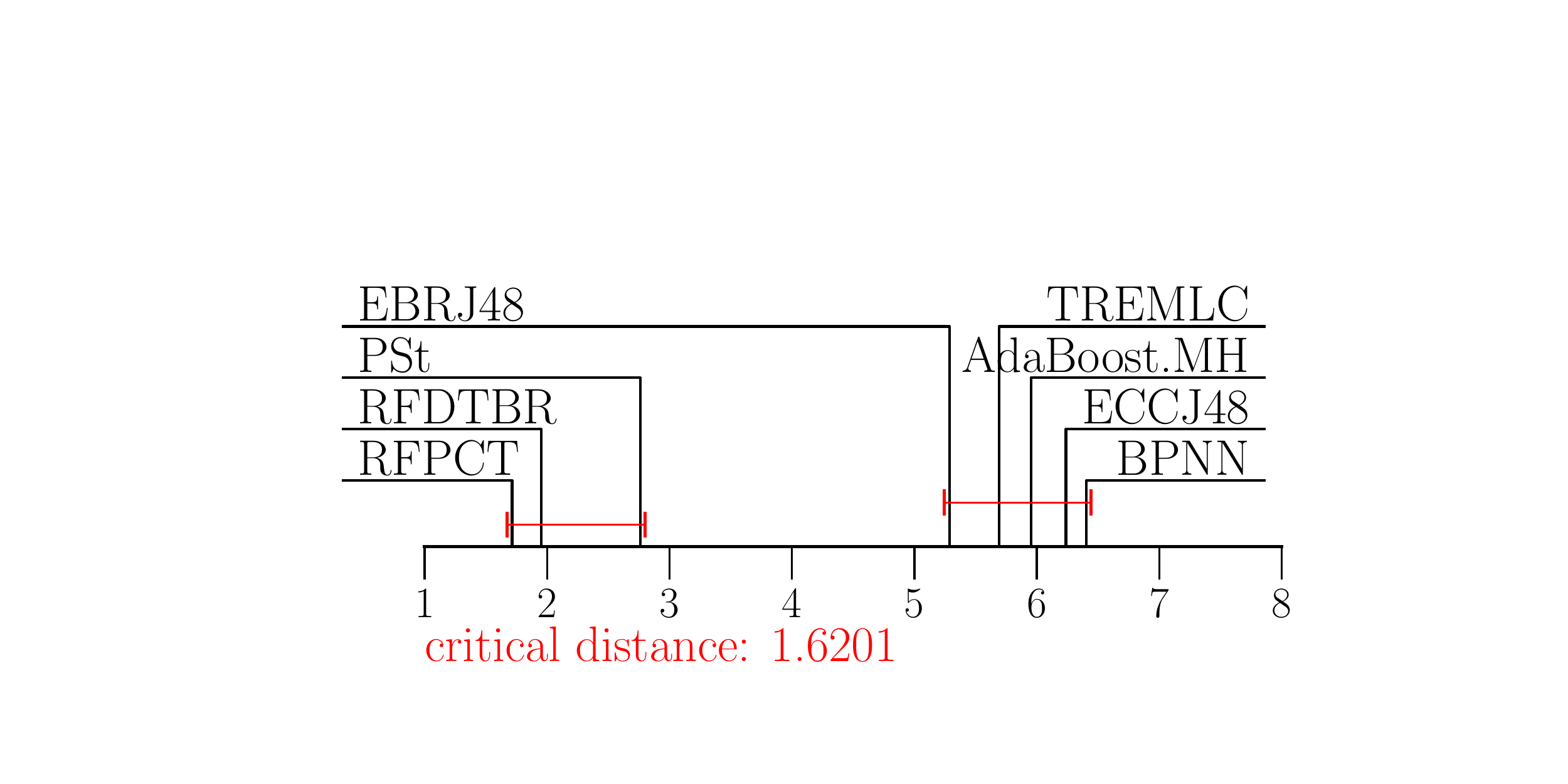}%
\label{fig:BM:trainTime}}
\hfil
\subfloat[Test time]{\includegraphics[width=0.45\textwidth]{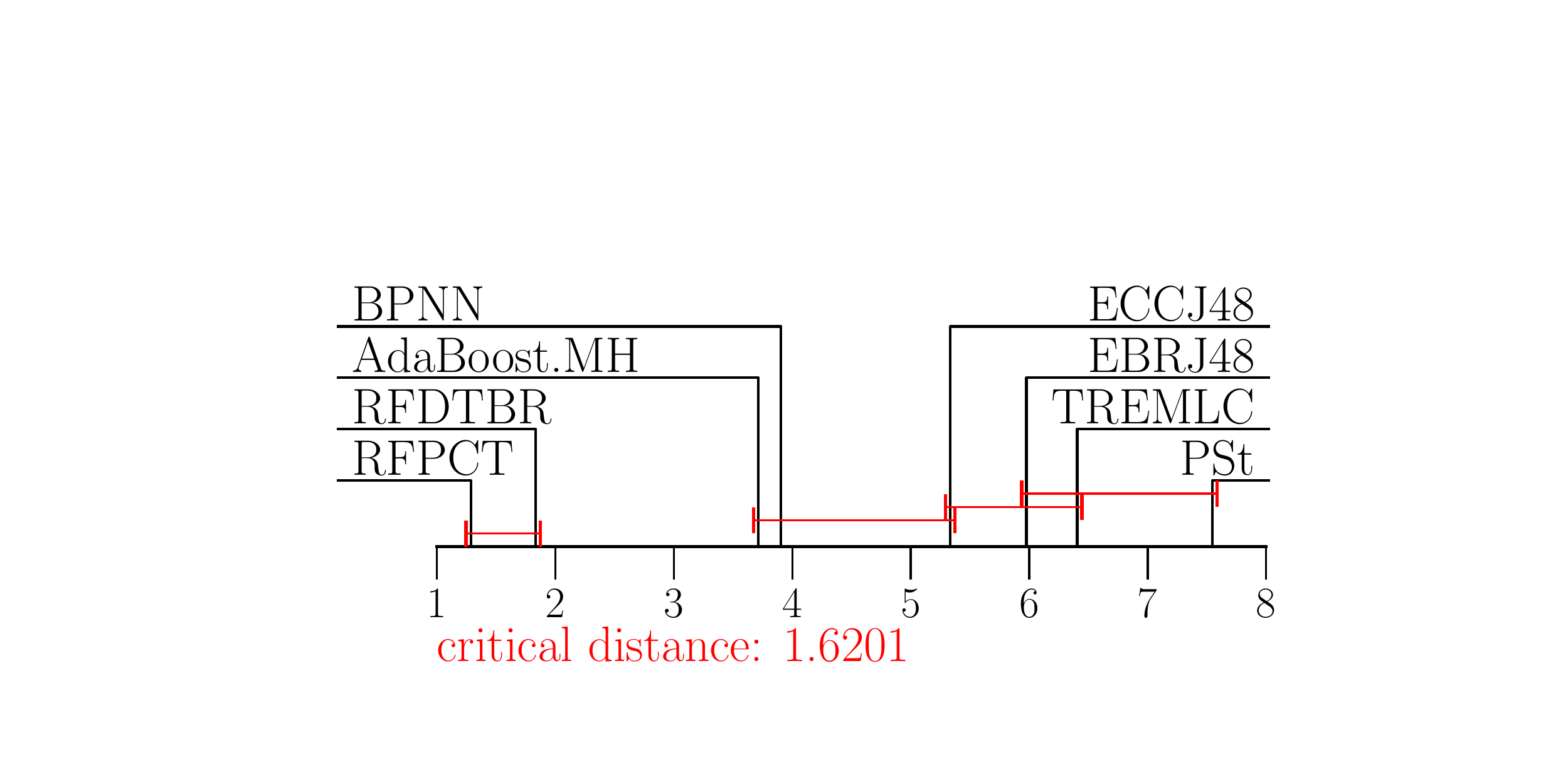}%
\label{fig:BM:testTime}}
\caption{Average rank diagrams comparing the efficiency (running times) of the best performing methods.}
\label{fig:time}
\end{figure}

We focus the discussion now on comparing the efficiency of the selected best performing methods in terms of training and testing times.
\figurename~\ref{fig:time} shows that RFPCT is the most efficient method -- it learns a predictive model fastest and makes predictions the fastest. It is then followed by RFDTBR. The differences in efficiency of these two methods as compared to the rest of the methods are statistically significant for both the training time (except for the PSt method) and the testing time.

We next consider the speed up of RFPCT (as the top efficient method) relative to the remaining methods across all datasets\footnote{The relative speed up is calculated as the average over the datasets of the ratio of the times needed to learn a model by the other methods and by using RFPCTs.}. In a nutshell, RFDTBR is slower than RFPCT $\sim 2.5$ times, AdaBoost.MH $\sim 28.1$ times, PSt $\sim 29.6$ times, TREMLC $\sim 48.1$, EBRJ48 $\sim 61$ times, BPNN $\sim 63$ times and ECCJ48 $\sim 76.7$ times. 
We believe that the difference in efficiency between the two top-ranked methods (RFPCT and RFDTBR) is due to the fact that RFPCT typically learns shallower/smaller trees \cite{Kocev:Journal:2013}. 
Notwithstanding this difference, the comparison of efficiency clearly identifies RFPCT and RFDTBR as the most efficient MLC methods.

\section{Conclusion}\label{sec:Conclusions}

This works presents the most comprehensive comparative study of MLC methods to date. It presents an in-depth theoretical and empirical analysis of a variety of MLC methods. Considering the ever increasing interest in MLC by the research community and its increased practical relevance, this study maps the landscape of MLC methods and provides guidelines for practitioners on the usage of the MLC methods, on selecting the best baselines when proposing a new method, or selecting the first methods to try out on new MLC datasets.


The theoretical analysis of the MLC methods focuses on aspects covering different viewpoints of the methods, such as (1) detailing the inner working procedures of the methods; (2) stressing their strengths and weaknesses; (3) discussing their potential to address specific properties of the MLC task, i.e., exploit the potential label dependencies and handle the high-dimensionality of the label space; and (4) analysis of the computational cost for training a predictive model and making a prediction using it. We divide the methods into two groups: problem transformation and algorithm adaptation. While the former group of methods decomposes the MLC problem into simpler problem(s) that are addressed with standard machine learning methods, the latter group of methods addresses the MLC problem in a holistic manner - it learns a model predicting all labels simultaneously.


Our empirical study of the methods is by far the \emph{largest empirical study for MLC methods to date}: It considers 26 MLC methods learning predictive models for 42 datasets, and evaluating them by 18 predictive performance measures and 2 efficiency criteria. The datasets stem from various domains, including text (news, reports), medicine, multimedia (images and audio), bioinformatics, biology and chemistry. The 18 predictive performance evaluation criteria provide a whole range of viewpoints on the performance of the MLC methods, including their capability of predicting bi-partitions (per example and label), label ranking, and the independence of the predictions regarding the threshold. 

Regarding the experimental design, we adhere to the literature recognized standards for conducting large experimental studies in the machine learning community. It includes time-constrained hyperparameter-optimization of the methods' parameters on a sub-sampled portion of the datasets. 
The parameterization is performed using literature recognized values for the parameters.
For analysis of the results, we use the Friedman and Nemenyi statistical tests and present their outcomes using average ranking diagrams.

We analyze and discuss the results of the experiments in detail, first separately for problem transformation and algorithm adaptation methods, and then on a selection of the best performing methods from both groups. Based on the analysis of the performance within each group, we selected 8 best performing methods (RFDTBR, AdaBoost.MH, ECCJ48, TREMLC, PSt and EBRJ48 among problem transformation, and RFPCT and BPNN among algorithm adaptation methods). We then compare these in order to identify an even more compact set of methods as best performing. 

The evaluation outlines RFPCT, RFDTBR, EBRJ48, AdaBoost.MH and ECCJ48 as best performing, considering the 18 evaluation measures. These methods have their strengths and weaknesses and should be selected based on the context of use. For example, ECCJ48 is very strong on recall-based measures and weak on precision-based measures -- this is reversed for AdaBoost.MH. Notwithstanding, RFPCT and RFDTBR are clearly the top-performing methods (having the top-ranked positions across the majority of the evaluation measures) as well as the most efficient MLC methods (within the selected best performing group of methods).

Being very comprehensive in terms of MLC methods, datasets and evaluation measures, this study opens several avenues for further research and exploration. To begin with, while it is important to provide different viewpoints by using different evaluation measures, this makes it difficult to select an optimal method. To alleviate this problem, we aim to use our empirical results to study the evaluation measures and identify relevant relationships between them. Furthermore, the results of the large set of experiments will allow us to relate the data set properties and the performance of the methods in a meta learning study and investigate the influence of dataset properties on predictive performance. For this purpose, we will first describe the MLC datasets with features that describe their specific MLC task properties and then use these features to learn meta models. Finally, considering that we store the actual prediction scores of the models, we can further experiment with thresholding functions that separate the relevant from irrelevant labels. We will analyse different MLC thresholding techniques, with the aim to improve existing or design novel thresholding methods.

\section*{Acknowledgment} 
JB and DK acknowledge the financial support of the Slovenian Research Agency (research project No.~J2-9230). LT acknowledges the financial support of the Slovenian Research Agency (research program No.~P5-0093 and research project No.~V5-1930). SD acknowledges the financial support of the Slovenian Research Agency (research program No.~P2-0103) and the European Commission through the project TAILOR - Foundations of Trustworthy AI - Integrating Reasoning, Learning and Optimization (grant No. 952215).


%





\ifCLASSOPTIONcaptionsoff
  \newpage
\fi

\bibliographystyle{IEEEtran}
\bibliography{IEEEabrv,referenceValid.bib}

\begin{thebibliography}{10}
\providecommand{\url}[1]{#1}
\csname url@samestyle\endcsname
\providecommand{\newblock}{\relax}
\providecommand{\bibinfo}[2]{#2}
\providecommand{\BIBentrySTDinterwordspacing}{\spaceskip=0pt\relax}
\providecommand{\BIBentryALTinterwordstretchfactor}{4}
\providecommand{\BIBentryALTinterwordspacing}{\spaceskip=\fontdimen2\font plus
\BIBentryALTinterwordstretchfactor\fontdimen3\font minus
  \fontdimen4\font\relax}
\providecommand{\BIBforeignlanguage}[2]{{%
\expandafter\ifx\csname l@#1\endcsname\relax
\typeout{** WARNING: IEEEtran.bst: No hyphenation pattern has been}%
\typeout{** loaded for the language `#1'. Using the pattern for}%
\typeout{** the default language instead.}%
\else
\language=\csname l@#1\endcsname
\fi
#2}}
\providecommand{\BIBdecl}{\relax}
\BIBdecl

\bibitem{Madjarov2012}
G.~Madjarov, D.~Kocev, D.~Gjorgjevikj, and S.~D\v{z}eroski, ``An extensive
  experimental comparison of methods for multi-label learning,'' \emph{Pattern
  Recognition}, vol.~45, pp. 3084 -- 3104, 2012.

\bibitem{Herrera2016}
F.~Herrera, A.~J. Rivera, M.~J. del Jesus, and F.~Charte, \emph{Multilabel
  Classification}.\hskip 1em plus 0.5em minus 0.4em\relax Springer Cham,
  Switzerland: Springer, 2016.

\bibitem{Tsoumakas2007}
G.~Tsoumakas and I.~Katakis, ``Multi-label classification: {A}n overview,''
  \emph{International Journal of Data Warehousing and Mining}, vol. 2007, pp.
  1--13, 2007.

\bibitem{Kocev:Journal:2013}
D.~Kocev, C.~Vens, J.~Struyf, and S.~D\v{z}eroski, ``Tree ensembles for
  predicting structured outputs,'' \emph{Pattern Recognition}, vol.~46, pp.
  817--833, 2013.

\bibitem{GO}
J.~Xu, J.~Liu, J.~Yin, and C.~Sun, ``A multi-label feature extraction algorithm
  via maximizing feature variance and feature-label dependence
  simultaneously,'' \emph{Knowledge-Based Systems}, vol.~98, pp. 172--184,
  2016.

\bibitem{birds}
F.~Briggs, B.~Lakshminarayanan, L.~Neal, X.~Z. Fern, R.~Raich, S.~Hadley,
  A.~Hadley, and M.~Betts, ``Acoustic classification of multiple simultaneous
  bird species: A multi-instance multi-label approach,'' \emph{The Journal of
  the Acoustical Society of America}, vol. 131, no.~6, pp. 4640--4650, 2012.

\bibitem{bibtex}
I.~Katakis, G.~Tsoumakas, and I.~Vlahavas, ``Multilabel text classification for
  automated tag suggestion,'' in \emph{{Proceedings of the {ECML/PKDD} 2008
  Discovery Challenge}}, 2008.

\bibitem{scene}
M.~Boutell, J.~Luo, X.~Shen, and C.~Brown, ``Learning multi-label scene
  classification,'' \emph{Pattern Recognition}, vol.~37, no.~9, pp. 1757--1771,
  2004.

\bibitem{CHD}
G.~P. Liu, G.~Z. Li, Y.~L. Wang, and Y.~Q. Wang, ``Modelling of inquiry
  diagnosis for coronary heart disease in traditional chinese medicine by using
  multi-label learning,'' \emph{BMC Complementary and Alternative Medicine},
  vol.~10, p.~37, 2010.

\bibitem{biomedicalimagesegmentation}
L.~Grady and G.~Funka-Lea, ``Multi-label image segmentation for medical
  applications based on graph-theoretic electrical potentials,'' in
  \emph{Computer Vision and Mathematical Methods in Medical and Biomedical
  Image Analysis}.\hskip 1em plus 0.5em minus 0.4em\relax Berlin, Heidelberg:
  Springer Berlin Heidelberg, 2004, pp. 230--245.

\bibitem{neonatalbrains}
N.~Ratnarajah and A.~Qiu, ``Multi-label segmentation of white matter
  structures: Application to neonatal brains,'' \emph{NeuroImage}, vol. 102,
  pp. 913 -- 922, 2014.

\bibitem{WaterQuality}
H.~Blockeel, S.~D{\v{z}}eroski, and J.~Grbovi{\'{c}}, ``Simultaneous prediction
  of multiple chemical parameters of river water quality with {TILDE},'' in
  \emph{{Principles of Data Mining and Knowledge Discovery}}.\hskip 1em plus
  0.5em minus 0.4em\relax Berlin, Heidelberg: Springer, 1999, pp. 32--40.

\bibitem{SocialSciences}
A.~Schulz, E.~{Loza Mencía}, and B.~Schmidt, ``A rapid-prototyping framework
  for extracting small-scale incident-related information in microblogs:
  Application of multi-label classification on tweets,'' \emph{Information
  Systems}, vol.~57, pp. 88 -- 110, 2016.

\bibitem{commerce}
H.~Wang, Z.~Li, J.~Huang, P.~Hui, W.~Liu, T.~Hu, and G.~Chen, ``Collaboration
  based multi-label propagation for fraud detection,'' in \emph{Proceedings of
  the Twenty-Ninth International Joint Conference on Artificial Intelligence,
  {IJCAI-20}}.\hskip 1em plus 0.5em minus 0.4em\relax International Joint
  Conferences on Artificial Intelligence Organization, 2020, pp. 2477--2483.

\bibitem{RecentTrends}
W.~Liu, X.~Shen, H.~Wang, and I.~W. Tsang, ``The emerging trends of multi-label
  learning,'' 2020.

\bibitem{Gibaja2014}
E.~Gibaja and S.~Ventura, ``A tutorial on multilabel learning,'' \emph{ACM
  Computing Surveys}, vol.~47, no.~3, pp. 52:1--52:38, 2015.

\bibitem{Zhang2014}
M.~L. Zhang and Z.~H. Zhou, ``A review on multi-label learning algorithms,''
  \emph{IEEE Transactions on Knowledge and Data Engineering}, vol.~26, pp.
  1819--1837, 2014.

\bibitem{Moyano2018}
J.~M. Moyano, E.~L.~G. Galindo, K.~J. Cios, and S.~Ventura, ``Review of
  ensembles of multi-label classifiers: Models, experimental study and
  prospects,'' \emph{Information Fusion}, vol.~44, pp. 33--45, 2018.

\bibitem{Zhang2018}
M.-L. Zhang, L.~Yu-Kun, L.~Xu-Ying, and X.~Geng, ``Binary relevance for
  multi-label learning: {A}n overview,'' \emph{Frontiers of Computer Science},
  vol.~12, pp. 191--202, 2018.

\bibitem{Rivolli2020}
A.~Rivolli, J.~Read, C.~Soares, B.~Pfahringer, and A.~C. P. L.~F. de~Carvalho,
  ``An empirical analysis of binary transformation strategies and base
  algorithms for multi-label learning,'' \emph{Machine Learning}, pp. 1--55,
  2020.

\bibitem{Read2010}
J.~Read, ``Scalable multi-label classification,'' Ph.D. dissertation,
  University of Waikato, Hamilton, New Zeland, 2010.

\bibitem{Schapire1999}
R.~Schapire and Y.~Singer, ``Improved boosting algorithms using
  confidence-rated predictions,'' \emph{Machine Learning}, vol.~37, no.~3, pp.
  297--336, 1999.

\bibitem{Kocev2011}
D.~Kocev, ``Ensembles for predicting structured outputs,'' Ph.D. dissertation,
  {Jožef Stefan International Postgraduate School}, Ljubljana, Slovenia, 2011.

\bibitem{Brinker2006}
K.~Brinker, ``On active learning in multi-label classification,'' in \emph{From
  Data and Information Analysis to Knowledge Engineering}.\hskip 1em plus 0.5em
  minus 0.4em\relax Berlin, Heidelberg: Springer, 2006, pp. 206--213.

\bibitem{Frnkranz2008}
J.~F{\"u}rnkranz, E.~H{\"u}llermeier, E.~Loza Menc{\'i}a, and K.~Brinker,
  ``Multilabel classification via calibrated label ranking,'' \emph{Machine
  Learning}, vol.~73, no.~2, pp. 133--153, 2008.

\bibitem{Read2011}
J.~Read, B.~Pfahringer, G.~Holmes, and E.~Frank, ``Classifier chains for
  multi-label classification,'' \emph{Machine Learning}, vol.~85, p. 333, 2011.

\bibitem{Read2008}
J.~Read, B.~Pfahringer, and G.~Holmes, ``Multi-label classification using
  ensembles of pruned sets,'' in \emph{{Proceedings of the 8th {IEEE}
  International Conference on Data Mining}}.\hskip 1em plus 0.5em minus
  0.4em\relax Washington, DC, USA: IEEE Computer Society, 2008, pp. 995--1000.

\bibitem{Guo2011}
Y.~Guo and S.~Gu, ``Multi-label classification using conditional dependency
  networks,'' in \emph{{Proceedings of the 22nd International Joint Conference
  on Artificial Intelligence}}.\hskip 1em plus 0.5em minus 0.4em\relax
  Barcelona, Spain: AAAI Press, 2011, pp. 1300--1305.

\bibitem{Heckerman2001}
D.~Heckerman, D.~M. Chickering, C.~Meek, R.~Rounthwaite, and C.~Kadie,
  ``Dependency networks for inference, collaborative filtering, and data
  visualization,'' \emph{Journal of Machine Learning Research}, vol.~1, pp.
  49--75, 2001.

\bibitem{Pearl1988}
J.~Pearl, ``Markov and bayesian networks: Two graphical representations of
  probabilistic knowledge,'' in \emph{{Probabilistic Reasoning in Intelligent
  Systems}}, J.~Pearl, Ed.\hskip 1em plus 0.5em minus 0.4em\relax San Francisco
  (CA): Morgan Kaufman Publishers, 1988, pp. 77--141.

\bibitem{Tsoumakas2009}
G.~Tsoumakas, D.~Anastasios, S.~Eleftherios, M.~Vasileios, K.~Ioannis, and
  I.~P. Vlahavas, ``Correlation-based pruning of stacked binary relevance
  models for multi-label learning,'' in \emph{{1st International Workshop on
  Learning from Multi-Label Data}}, 2009, pp. 101--116.

\bibitem{Cherman2012}
E.~Alvares-Cherman, J.~Metz, and M.~C. Monard, ``{Incorporating label
  dependency into the binary relevance framework for multi-label
  classification},'' \emph{Expert Systems with Applications}, vol.~39, no.~2,
  pp. 1647 -- 1655, 2012.

\bibitem{Chekina2009}
L.~Tenenboim, L.~Rokach, and B.~Shapira, ``Multi-label classification by
  analyzing labels dependencies,'' in \emph{{Proceedings of the 1st
  International Workshop on Learning from Multi-Label Data}}, 2009, pp.
  117--131.

\bibitem{Chekina2010}
------, ``Identification of label dependencies for multi-label
  classification,'' in \emph{{2nd International Workshop on Learning from
  Multi-Label Data}}, 2010, pp. 53--60.

\bibitem{Tsoumakas2011}
G.~Tsoumakas, I.~Katakis, and I.~Vlahavas, ``Random k-labelsets for multi-label
  classification,'' \emph{IEEE Transactions on Knowledge and Data Engineering},
  vol.~23, pp. 1079--1089, 2011.

\bibitem{Rokach2014}
L.~Rokach, A.~Schclar, and E.~Itach, ``Ensemble methods for multi-label
  classification,'' \emph{Expert Systems with Applications}, vol.~41, pp. 7507
  -- 7523, 2014.

\bibitem{Tsoumakas2008}
G.~Tsoumakas, I.~Katakis, and I.~P. Vlahavas, ``Effective and efficient
  multilabel classification in domains with large number of labels,'' in
  \emph{{Proceedings of the Workshop on Mining Multidimensional Data at
  {ECML/PKDD} 2008}}, 2008, pp. 53--59.

\bibitem{Ho1998}
T.~K. Ho, ``The random subspace method for constructing decision forests,''
  \emph{IEEE Transactions on Pattern Analysis and Machine Intelligence},
  vol.~20, no.~8, pp. 832--844, 1998.

\bibitem{Schapire2000}
R.~Schapire and Y.~Singer, ``Boostexter: A boosting-based system for text
  categorization,'' \emph{Machine Learning}, vol.~39, pp. 135--168, 2000.

\bibitem{Schapire1997}
Y.~Freund and R.~Schapire, ``A decision-theoretic generalization of on-line
  learning and an application to boosting,'' \emph{Journal of Computer and
  System Sciences}, vol.~55, no.~1, pp. 119 -- 139, 1997.

\bibitem{Huang2017}
K.~H. Huang and H.~T. Lin, ``Cost-sensitive label embedding for multi-label
  classification,'' \emph{Machine Learning}, vol. 106, no.~9, pp. 1725--1746,
  2017.

\bibitem{Kruskal1964}
J.~B. Kruskal, ``Multidimensional scaling by optimizing goodness of fit to a
  nonmetric hypothesis,'' \emph{Psychometrika}, vol.~29, pp. 1--27, 1964.

\bibitem{Tsoumakas2010b}
G.~Nasierding, A.~Kouzani, and G.~Tsoumakas, ``A triple-random ensemble
  classification method for mining multi-label data,'' in \emph{{{IEEE}
  International Conference on Data Mining Workshops}}.\hskip 1em plus 0.5em
  minus 0.4em\relax Washington, DC, USA: IEEE Computer Society, 2010, pp.
  49--56.

\bibitem{Blockeel1998}
H.~Blockeel, L.~D. Raedt, and J.~Ramon, ``Top-down induction of clustering
  trees,'' in \emph{{Proceedings of the 15th International Conference on
  Machine Learning}}.\hskip 1em plus 0.5em minus 0.4em\relax San Francisco, CA,
  USA: Morgan Kaufmann Publishers, 1998, pp. 55--63.

\bibitem{Quinlan1986}
J.~R. Quinlan, ``Induction of decision trees,'' \emph{Machine Learning},
  vol.~1, pp. 81--106, 1986.

\bibitem{Zhang2006}
M.-L. Zhang and Z.-H. Zhou, ``Multilabel neural networks with applications to
  functional genomics and text categorization,'' \emph{IEEE Transactions on
  Knowledge and Data Engineering}, vol.~18, pp. 1338--1351, 2006.

\bibitem{Read2014}
J.~Read and F.~Perez-Cruz, ``Deep learning for multi-label classification,''
  2014.

\bibitem{Salakutinov2006}
G.~Hinton and R.~Salakhutdinov, ``Reducing the dimensionality of data with
  neural networks,'' \emph{Science}, vol. 313, no. 5786, pp. 504--507, 2006.

\bibitem{Hinton2002}
G.~Hinton, ``Training products of experts by minimizing contrastive
  divergence,'' \emph{Neural Computing}, vol.~14, pp. 1771--1800, 2002.

\bibitem{Sapozhnikova2009}
E.~Sapozhnikova, ``{ART}-based neural networks for multi-label
  classification,'' in \emph{{Advances in Intelligent Data Analysis
  VIII}}.\hskip 1em plus 0.5em minus 0.4em\relax Berlin, Heidelberg: Springer,
  2009, pp. 167--177.

\bibitem{Tan1995}
A.~H. Tan, ``Adaptive resonance associative map,'' \emph{Neural Networks},
  vol.~8, pp. 437 -- 446, 1995.

\bibitem{Chen2016}
W.~J. Chen, Y.~H. Shao, C.~N. Li, and N.~Y. Deng, ``{MLTSVM}: {A} novel twin
  support vector machine to multi-label learning,'' \emph{Pattern Recognition},
  vol.~52, pp. 61--74, 2016.

\bibitem{Chandra2007}
Jayadeva, R.~Khemchandani, and S.~Chandra, ``Twin support vector machines for
  pattern classification,'' \emph{IEEE Transactions on Pattern Analysis and
  Machine Intelligence}, vol.~29, pp. 905--910, 2007.

\bibitem{Zhang2005}
M.-L. Zhang and Z.-H. Zhou, ``A k-nearest neighbor based algorithm for
  multi-label classification,'' in \emph{{{IEEE} International Conference on
  Granular Computing}}.\hskip 1em plus 0.5em minus 0.4em\relax Washington, DC,
  USA: IEEE, 2005, pp. 718 -- 721.

\bibitem{Ruiz1986}
E.~V. Ruiz, ``An algorithm for finding nearest neighbours in (approximately)
  constant average time,'' \emph{Pattern Recognition Letters}, vol.~4, pp.
  145--157, 1986.

\bibitem{Breiman2001}
L.~Breiman, ``Random forests,'' \emph{Machine Learning}, vol.~45, no.~1, pp.
  5--32, 2001.

\bibitem{Bellman1954}
R.~Bellman, ``The theory of dynamic programming,'' \emph{Bulletin of the
  American Mathematical Society}, vol.~60, no.~6, pp. 503--515, 1954.

\bibitem{Guyon2003}
I.~Guyon and A.~Elisseeff, ``An introduction to variable and feature
  selection,'' \emph{Journal of Machine Learning Research}, vol.~3, no.~7, pp.
  1157--1182, 2003.

\bibitem{Kira1992}
K.~Kira and L.~Rendell, ``The feature selection problem: Traditional methods
  and a new algorithm,'' in \emph{{Proceedings of the 10th National Conference
  on Artificial Intelligence}}.\hskip 1em plus 0.5em minus 0.4em\relax San
  Jose, California: AAAI Press, 1992, pp. 129--134.

\bibitem{Jain16:PFAST}
H.~Jain, Y.~Prabhu, and M.~Varma, ``Extreme {{Multi}}-{{Label Loss Functions}}
  for {{Recommendation}}, {{Tagging}}, {{Ranking}} \& {{Other Missing Label
  Applications}},'' in \emph{Proceedings of the 22nd {{ACM SIGKDD International
  Conference}} on {{Knowledge Discovery}} and {{Data Mining}}}, ser. {{KDD}}
  '16.\hskip 1em plus 0.5em minus 0.4em\relax {Association for Computing
  Machinery}, 2016, pp. 935--944.

\bibitem{ismis:hmlc}
T.~Stepi{\v{s}}nik and D.~Kocev, ``Hyperbolic embeddings for hierarchical
  multi-label classification,'' in \emph{International Symposium on
  Methodologies for Intelligent Systems}.\hskip 1em plus 0.5em minus
  0.4em\relax Springer, 2020, pp. 66--76.

\bibitem{Caruana2006}
R.~Caruana and A.~Niculescu-Mizil, ``An empirical comparison of supervised
  learning algorithms,'' in \emph{{Proceedings of the 23rd International
  Conference on Machine Learning}}.\hskip 1em plus 0.5em minus 0.4em\relax New
  York, USA: ACM, 2006, pp. 161--168.

\bibitem{Sechidis2011}
K.~Sechidis, G.~Tsoumakas, and I.~Vlahavas, ``On the stratification of
  multi-label data,'' in \emph{{Machine Learning and Knowledge Discovery in
  Databases}}.\hskip 1em plus 0.5em minus 0.4em\relax Berlin, Heidelberg:
  Springer, 2011, pp. 145--158.

\bibitem{automl}
F.~Hutter, L.~Kotthoff, and J.~Vanschoren, \emph{Automatic machine learning:
  methods, systems, challenges}.\hskip 1em plus 0.5em minus 0.4em\relax Berlin,
  Heilderberg: Springer, 2019.

\bibitem{DeSa2018}
A.~G.~C. de~S\'{a}, G.~L. Pappa, and A.~Freitas, ``Multi-label classification
  search space in the {MEKA} software,'' 2018.

\bibitem{Piotor2017}
P.~Szymański and T.~Kajdanowicz, ``A scikit-based python environment for
  performing multi-label classification,'' 2017.

\bibitem{Pedregosa2013}
L.~Buitinck, G.~Louppe, M.~Blondel, F.~Pedregosa, A.~Mueller, O.~Grisel,
  V.~Niculae, P.~Prettenhofer, A.~Gramfort, J.~Grobler, R.~Layton,
  J.~Vanderplas, A.~Joly, B.~Holt, and G.~Varoquaux, ``{API} design for machine
  learning software: experiences from the scikit-learn project,'' \emph{arxiv},
  2013.

\bibitem{Tsoumakas2011b}
G.~Tsoumakas, E.~Spyromitros-Xioufis, J.~Vilcek, and I.~Vlahavas, ``Mulan: {A}
  java library for multi-label learning,'' \emph{Journal of Machine Learning
  Research}, vol.~12, pp. 2411--2414, 2011.

\bibitem{Read2016}
J.~Read, P.~Reutemann, B.~Pfahringer, and G.~Holmes, ``{MEKA}: A
  multi-label/multi-target extension to {WEKA},'' \emph{Journal of Machine
  Learning Research}, vol.~17, pp. 1--5, 2016.

\bibitem{Singulariy}
G.~M. Kurtzer, S.~Vanessa, and B.~M. W., ``Singularity: Scientific containers
  for mobility of compute.'' \emph{PLoS ONE}, vol.~12, no.~5, 2017.

\bibitem{Reem2014}
A.-O. Reem, P.~Flach, and K.~Meelis, ``Multi-label classification: A
  comparative study on threshold selection methods,'' in \emph{{1st
  International Workshop on Learning over Multiple Contexts {(LMCE)} at
  {ECML-PKDD 2014}}}, 2014.

\bibitem{Iman1980}
R.~Iman and J.~Davenport, ``Approximations of the critical region of the
  {F}riedman statistic,'' \emph{Communications in Statistics-theory and
  Methods}, vol.~9, pp. 571--595, 1980.

\bibitem{Nemenyi1963}
P.~Nemenyi, ``Distribution-free multiple comparisons,'' Ph.D. dissertation,
  Princeton University, Princeton, USA, 1963.

\bibitem{Friedman1940}
M.~Friedman, ``A comparison of alternative tests of significance for the
  problem of {\textit{m}} rankings,'' \emph{The Annals of Mathematical
  Statistics}, vol.~11, no.~1, pp. 86--92, 1940.

\bibitem{NAS}
\BIBentryALTinterwordspacing
T.~Elsken, J.~H. Metzen, and F.~Hutter, ``Neural architecture search: A
  survey,'' \emph{Journal of Machine Learning Research}, vol.~20, no.~55, pp.
  1--21, 2019. [Online]. Available:
  \url{http://jmlr.org/papers/v20/18-598.html}
\BIBentrySTDinterwordspacing

\bibitem{shap}
M.~Lundberg, S., G.~Erion, and H.~e.~a. Chen, ``{From local explanations to
  global understanding with explainable AI for trees.}'' \emph{{Nature Machine
  Intelligence}}, vol.~2, pp. 56--67, 2020.

\end{thebibliography}



%


%
\vfil
\begin{IEEEbiographynophoto}{Jasmin Bogatinovski} is a research associate in the group of Distributed Operating Systems at TU Berlin. He received his MSc. title in Computer Science from the IPS Jo\v{z}ef Stefan in 2019 while working at the Department of Knowledge Technologies as collaborator. His research interest are in the areas of: artificial intelligence, machine learning and distributed systems. More specifically, he is interested in the area of meta learning and its impact across different learning tasks including single-target, multi-target classification/regression and the anomaly detection task. From the practical aspects he is interested in developing and applying novel machine learning methods in the domain of IT operation (AIOps). 
\end{IEEEbiographynophoto}
\begin{IEEEbiographynophoto}{Ljup\v{c}o Todorovski} is a full professor of informatics at the Faculty of Public Administration, University of Ljubljana and a senior researcher at the Department of Knowledge Technologies, Jo\v{z}ef Stefan Institute, Ljubljana, Slovenia. His research interests are in the field of machine learning; he develops algorithms for learning models of dynamical systems from observations of their behavior and domain-specific knowledge. In collaboration with researchers from various scientific domains he applies computational, machine learning and data mining methods to practical problems in biomedicine, earth and administrative sciences.
\end{IEEEbiographynophoto}
\begin{IEEEbiographynophoto}{Sa\v{s}o D\v{z}eroski} received his Ph.D. degree in computer science from the University of Ljubljana in 1995. He is Head of the Department of Knowledge Technologies at the Jo\v{z}ef Stefan Institute and full professor at the JS International Postgraduate School (both in Ljubljana, Slovenia). He is also a visiting professor at the Phi-Lab of ESRIN, European Space Agency (Frascati, Italy). His research interests fall in the field of artificial intelligence (AI), focusing on the development of data mining and machine learning methods for a variety of tasks - including the prediction of structured outputs and the automated modeling of dynamical systems - and their applications to practical problems in a variety of domains - ranging from agriculture and ecology, through medicine and pharmacology, to earth observation and space operations. 
In 2008, he was elected fellow of the European AI Society for his "Pioneering Work in the field of AI and Outstanding Service for the European AI community". In 2015, he became a foreign member of the Macedonian Academy of Sciences and Arts. In 2016, he was elected a member of Academia Europaea. He is currently serving as Chair of the Slovenian AI Society.
\end{IEEEbiographynophoto}
\begin{IEEEbiographynophoto}{Dragi Kocev}
is a researcher at the Department of Knowledge Technologies, JSI. He completed his PhD in 2011 at the JSI Postgraduate School in Ljubljana on the topic of learning ensemble models for predicting structured outputs. He was a visiting research fellow at the University of Bari, Italy in 2014/2015. He has participated in several national Slovenian projects, the EU funded projects IQ, PHAGOSYS and HBP. He was co-coordinator of the FP7 FET Open project MAESTRA. He is currently the Principal Investigator of two ESA funded projects: GALAXAI – Machine learning for spacecraft operation and AiTLAS – AI prototyping environment for EO. 
He has been member of the PC of many conferences (e.g., DS, ECML PKDD, AAAI, IJCAI, KDD) and member of the editorial board of Data Mining and Knowledge Discovery, Machine Learning Journal, Expert Systems with Applications (as Action Editor) and Ecological Informatics. He served as PC co-chair for DS 2014 and Journal track co-chair for ECML PKDD 2017.
\end{IEEEbiographynophoto}





\appendix
\section*{Complete results for the best performing methods}

\begin{figure*}[!h]
\centering
\subfloat[Accuracy]{\includegraphics[width=0.5\textwidth]{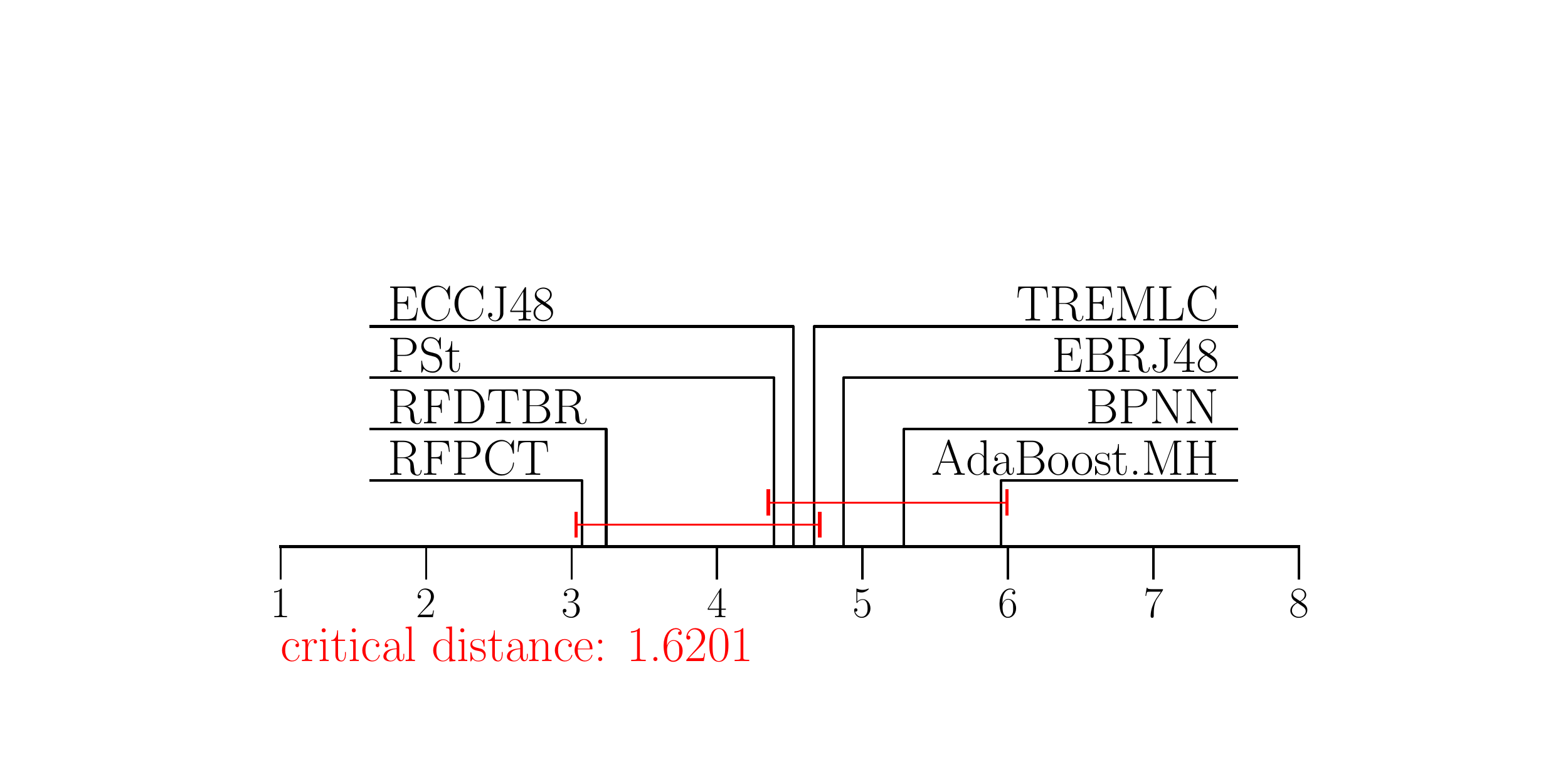}%
\label{fig:BST:HammingLoss}}
\subfloat[Subset accuracy]{\includegraphics[width=0.5\textwidth]{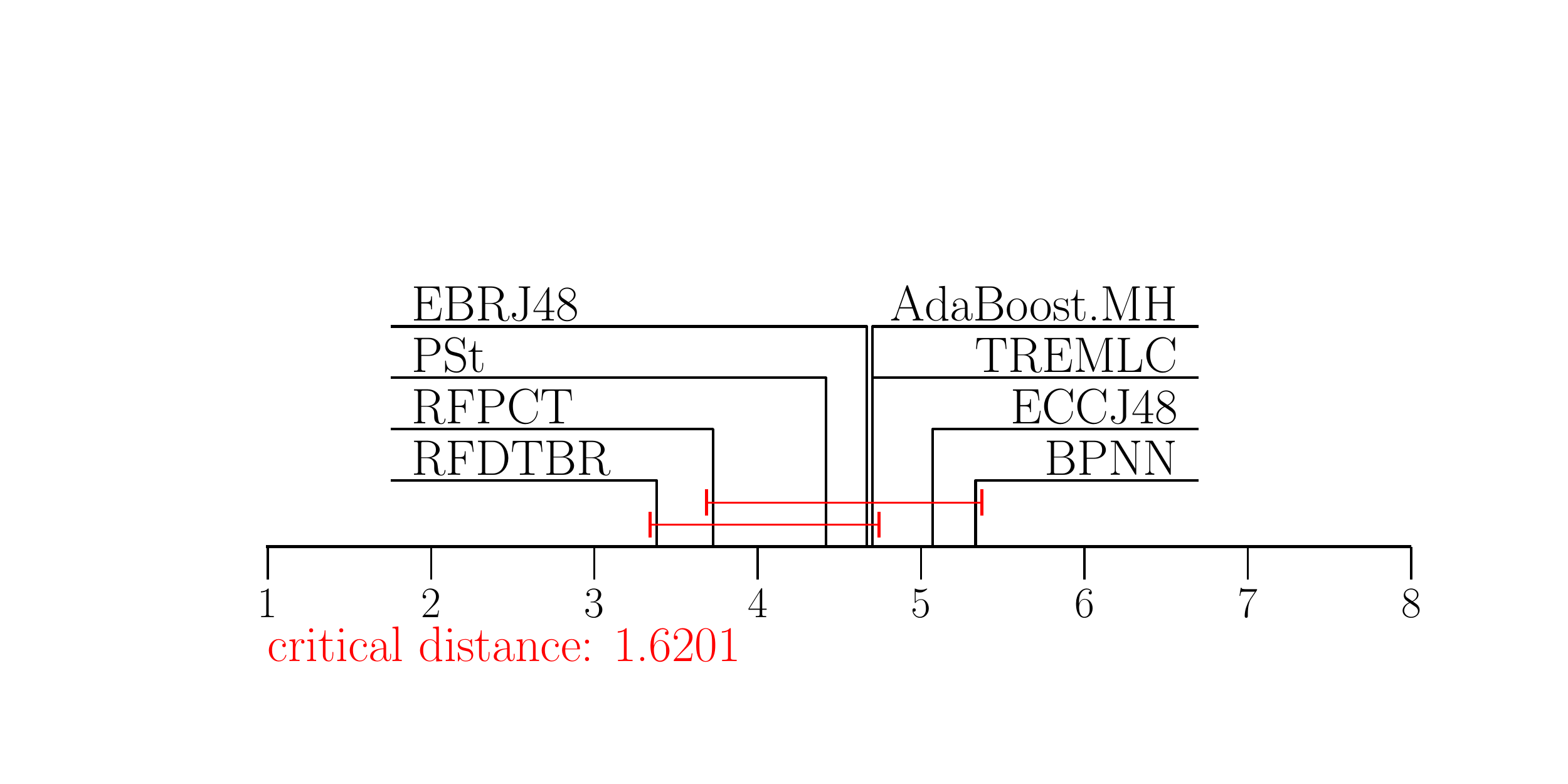}%
\label{fig:BST:Accuracy}}

\subfloat[Precision example-based]{\includegraphics[width=0.5\textwidth]{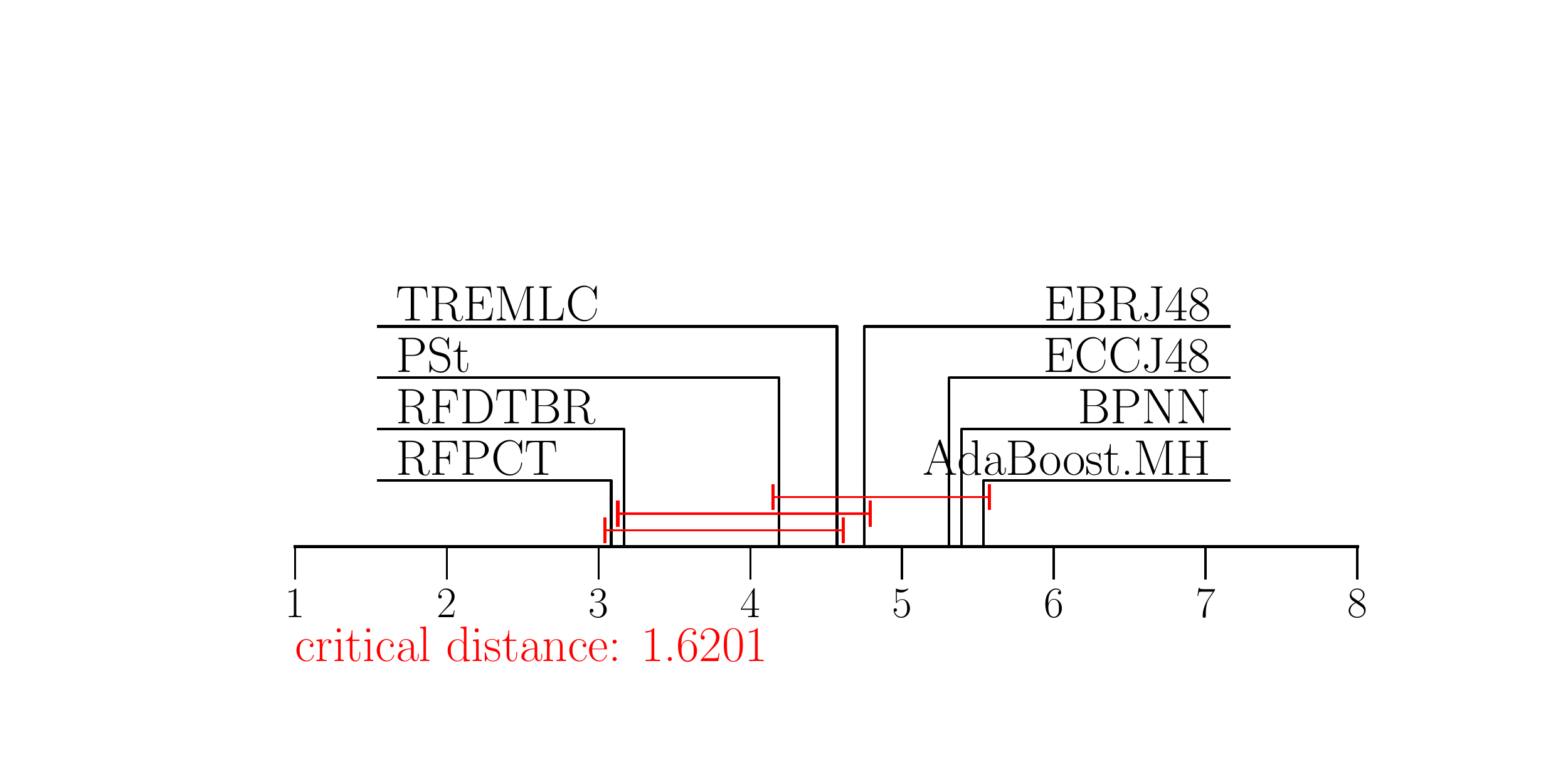}%
\label{fig:BST:Precision}}
\subfloat[Recall example-based]{\includegraphics[width=0.5\textwidth]{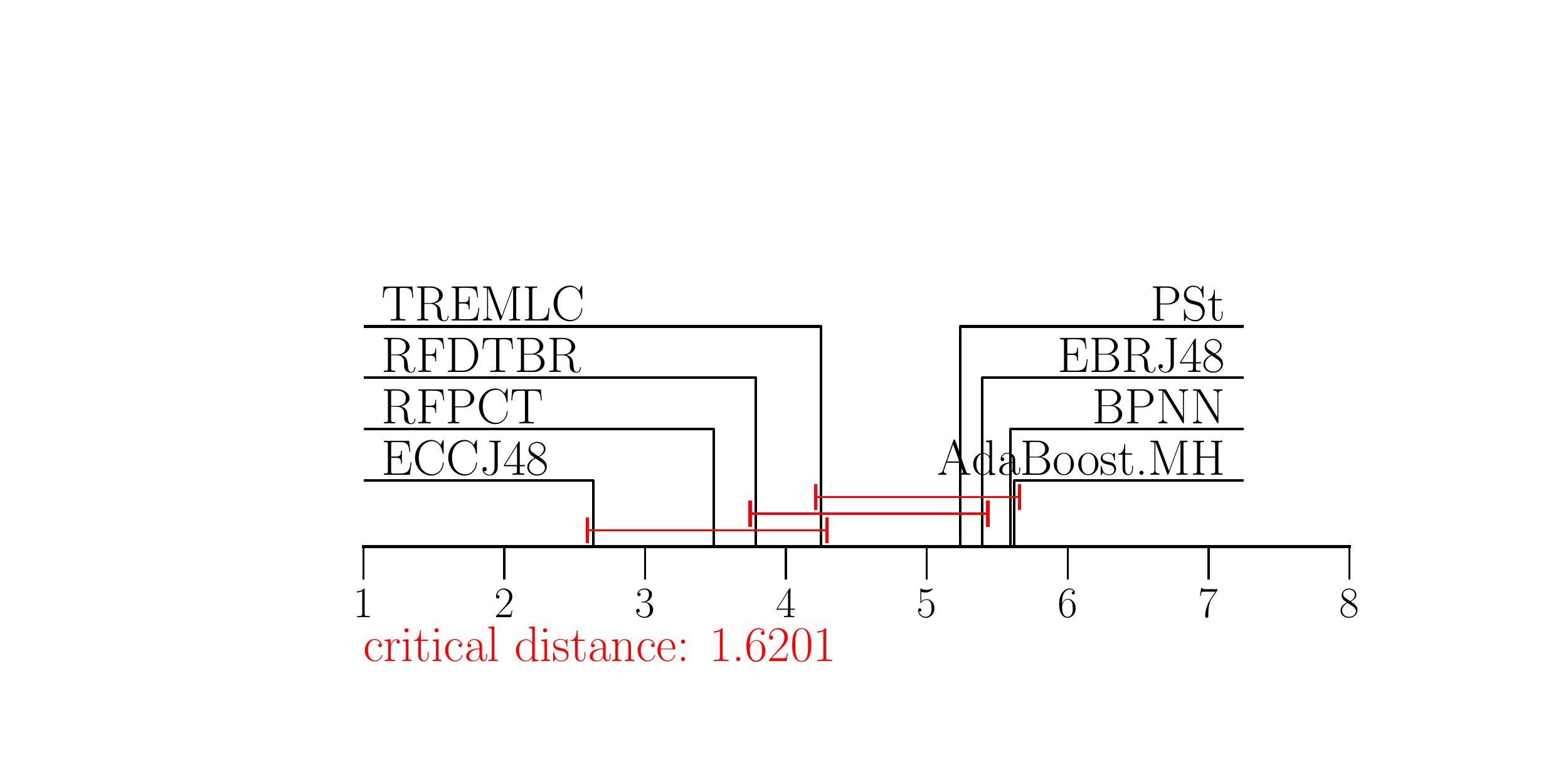}%
\label{fig:BST:Recall}}
\caption{Average rank diagrams comparing the best MLC methods using example-based measures.}
\label{fig:main:example_based}
\end{figure*}

\begin{figure*}[!h]
\centering
\subfloat[Ranking loss]{\includegraphics[width=0.5\textwidth]{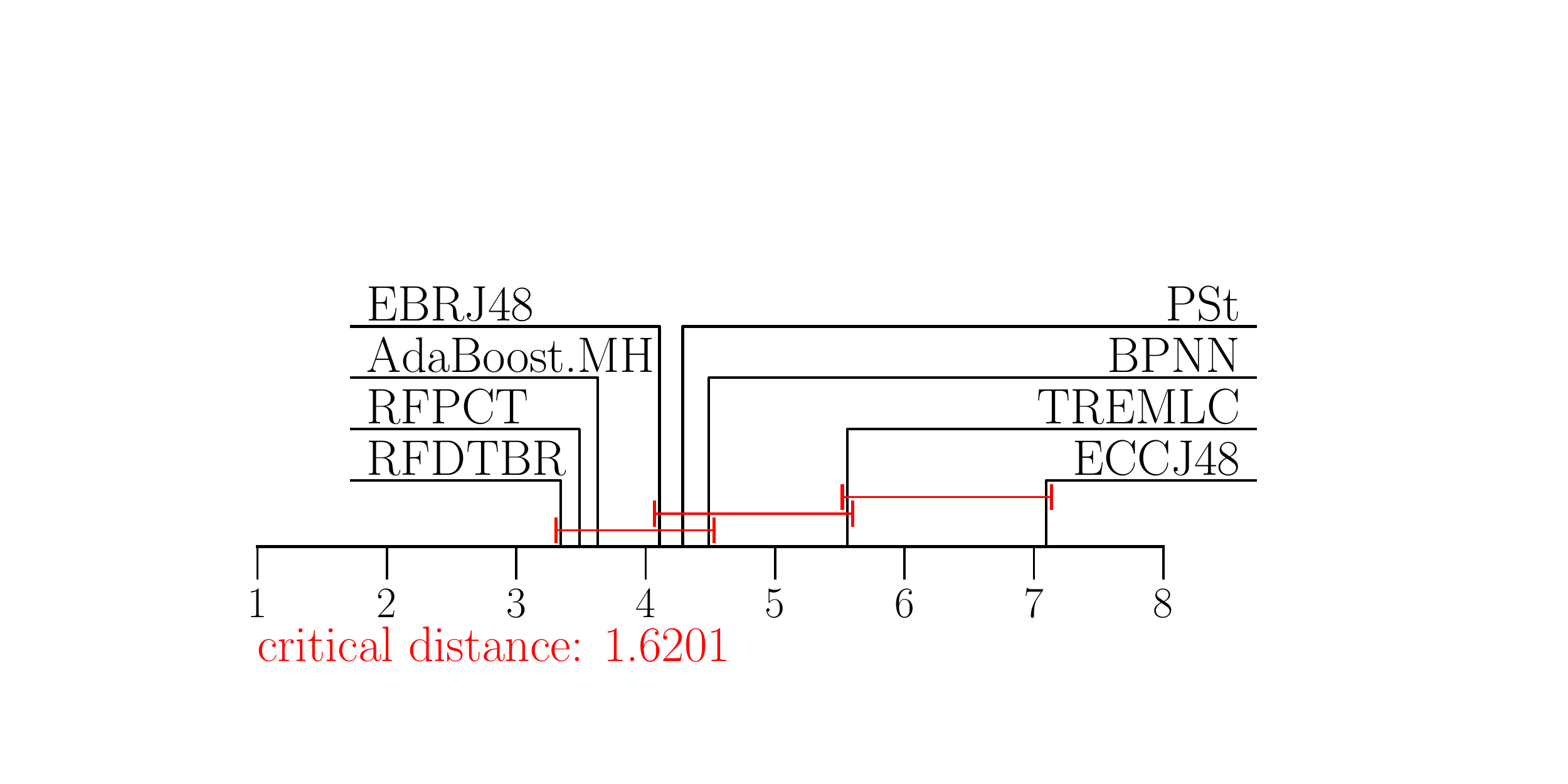}%
\label{fig:DBN:Rankingloss}}
\subfloat[Average precision]{\includegraphics[width=0.5\textwidth]{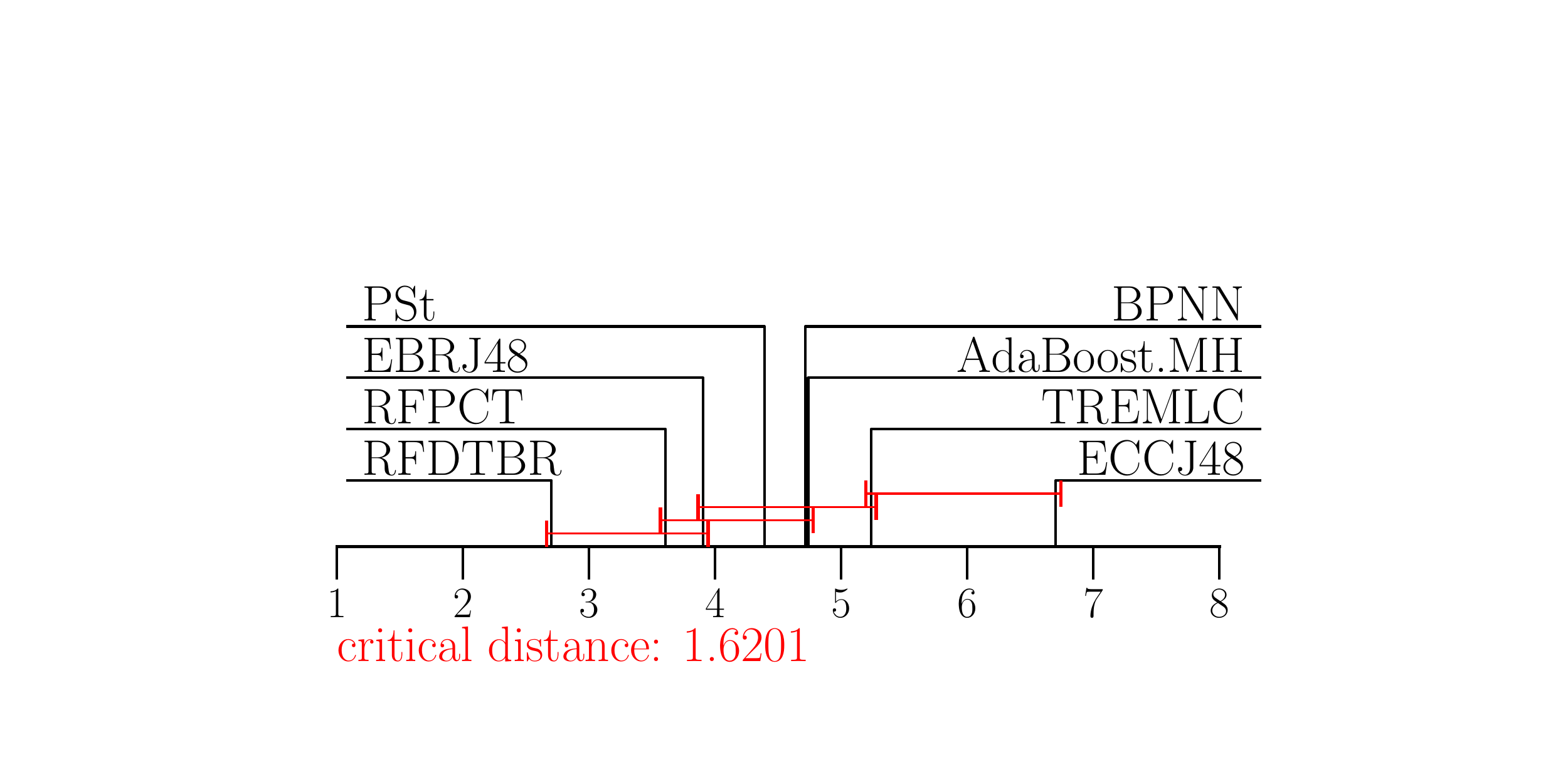}%
\label{fig:DBN:Averageprecision}}

\subfloat[Coverage]{\includegraphics[width=0.5\textwidth]{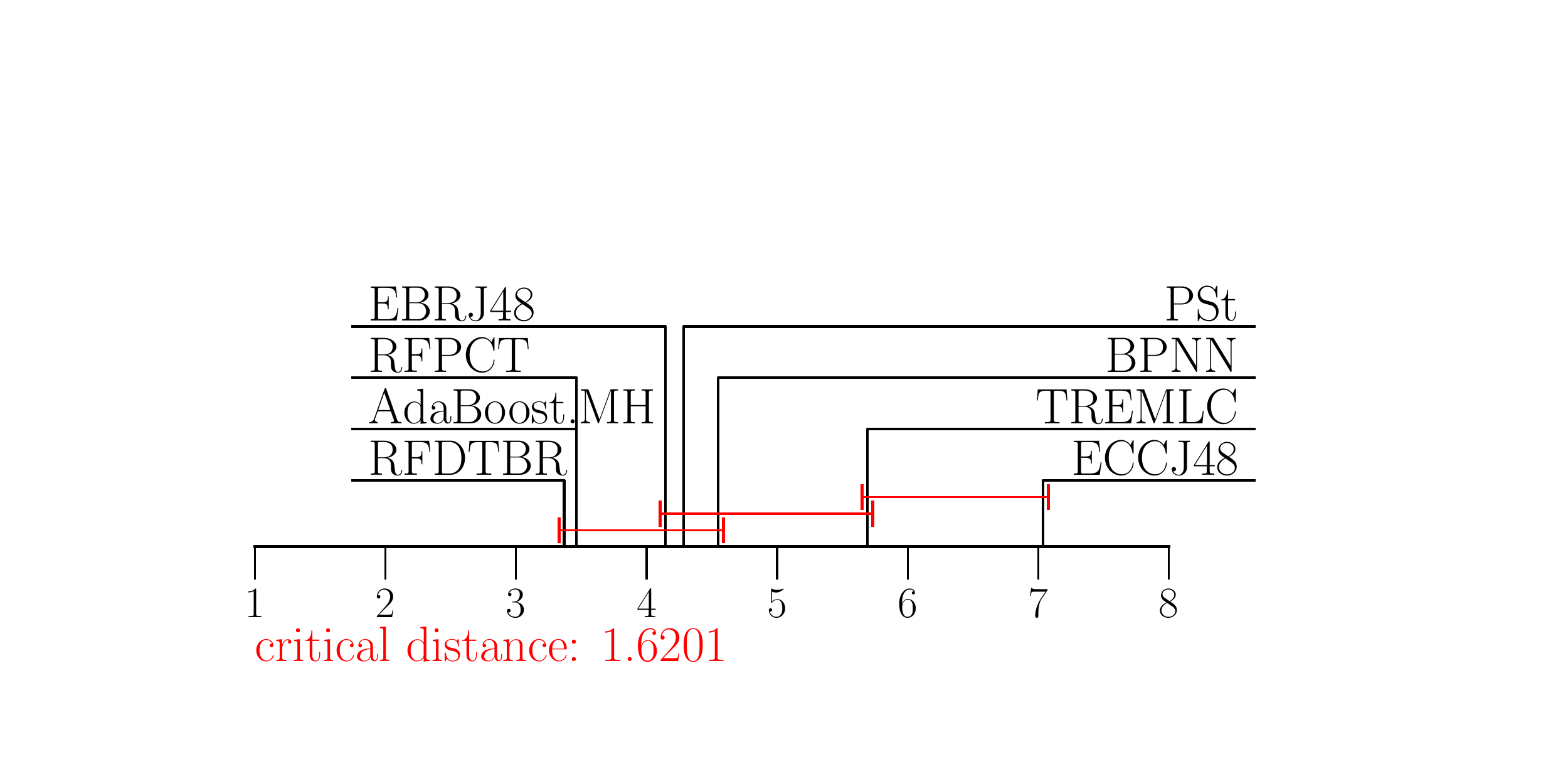}%
\label{fig:DBN:Coverage}}
\subfloat[One error]{\includegraphics[width=0.5\textwidth]{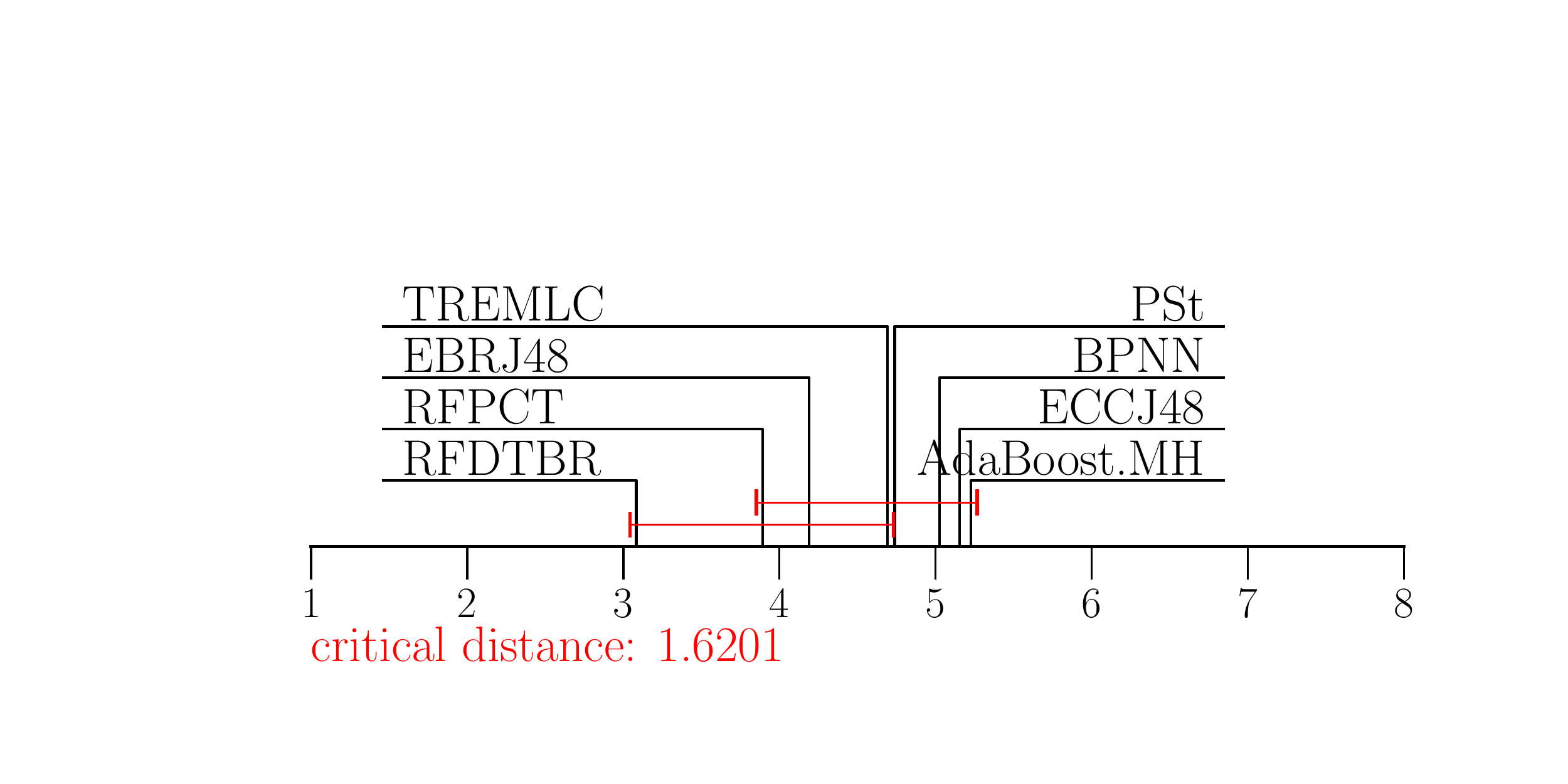}%
\label{fig:DBN:OneError}}

\caption{Average rank diagrams comparing the best MLC methods using ranking-based measures.}
\label{fig:main:ranking}
\end{figure*}

\begin{figure*}[!h]
\centering
\subfloat[AUROC]{\includegraphics[width=0.5\textwidth]{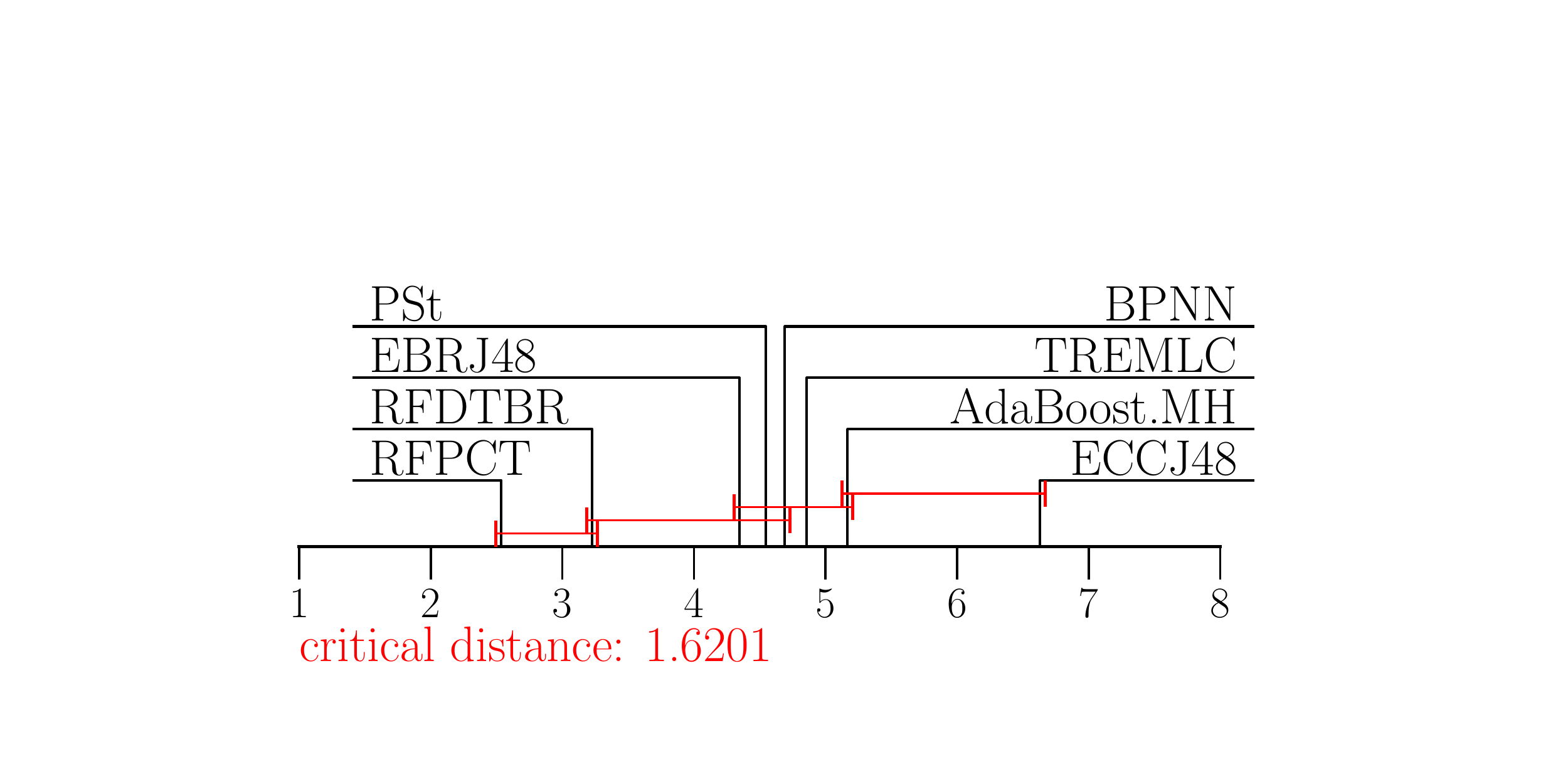}%
\label{fig:DBN:AUCROC}}
\subfloat[AUPRC]{\includegraphics[width=0.5\textwidth]{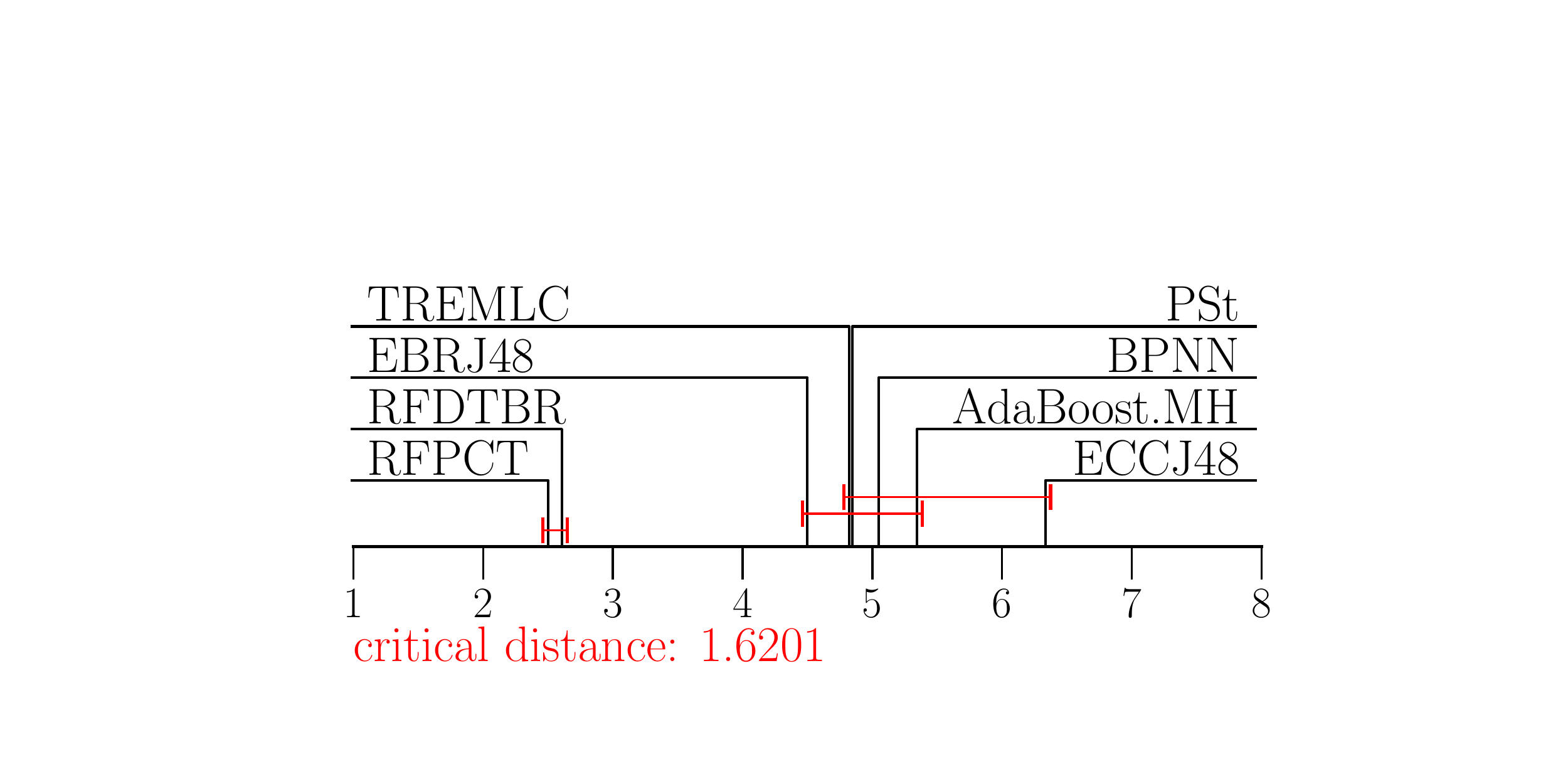}%
\label{fig:DBN:AUPRC}}
\caption{Average rank diagrams comparing the best MLC methods using threshold-independent measures.}
\label{fig:main:th_independent}
\end{figure*}

\begin{figure*}[!b]
\centering
\subfloat[Macro Precision]{\includegraphics[width=0.5\textwidth]{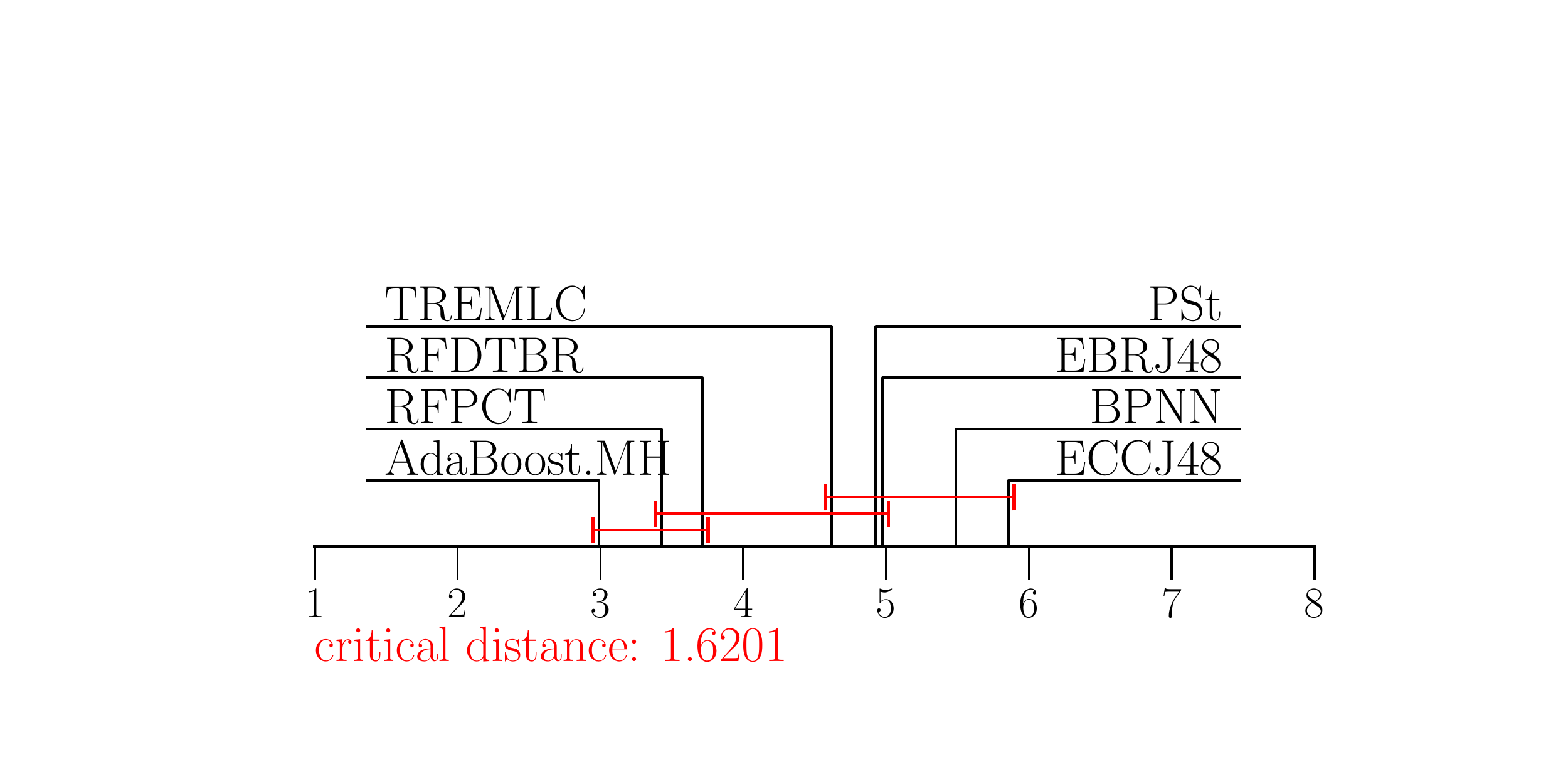}%
\label{fig:main:MacroPrecision}}
\subfloat[Micro Precision]{\includegraphics[width=0.5\textwidth]{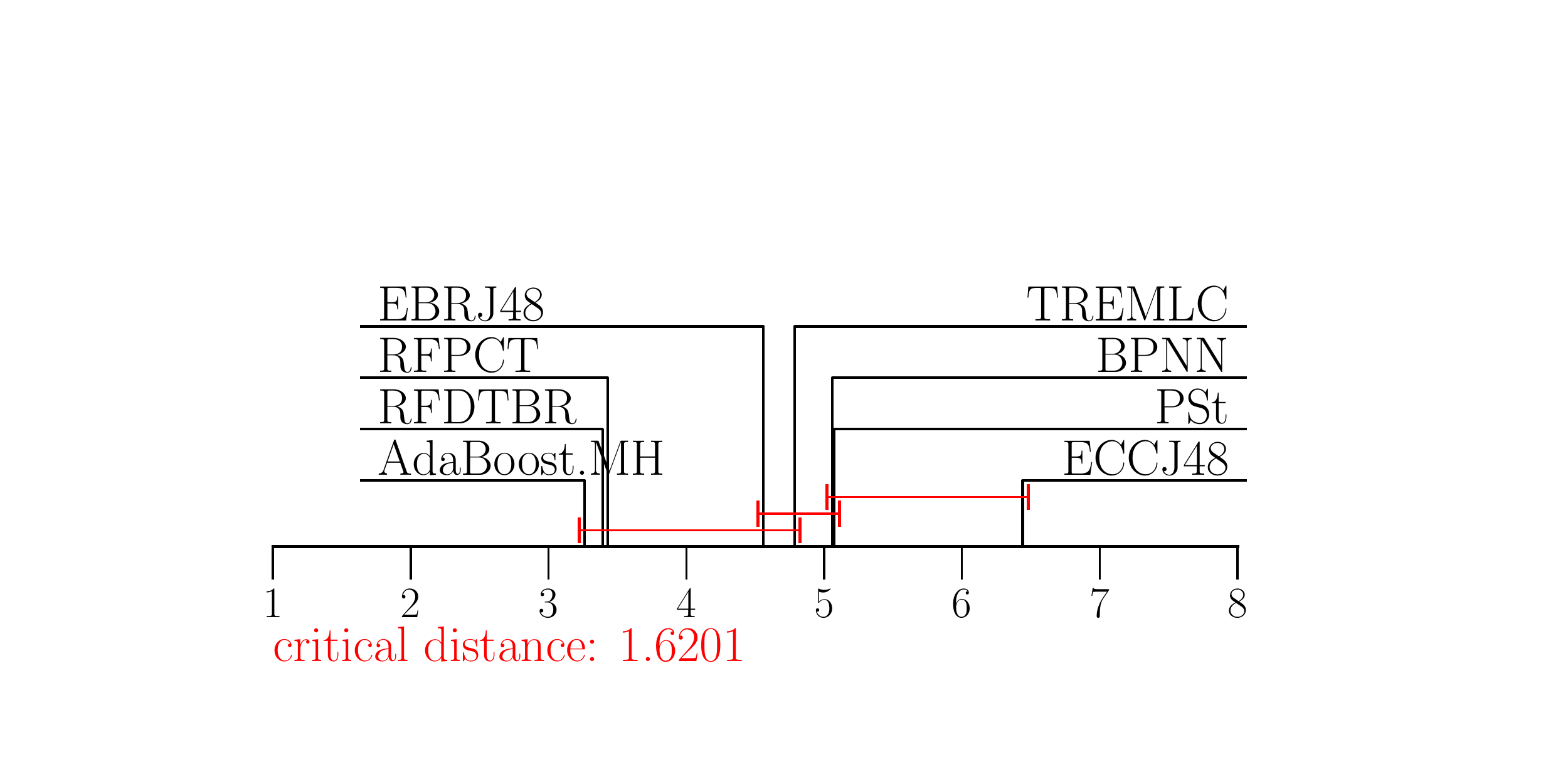}%
\label{fig:main:MicroPrecision}}

\subfloat[Macro Recall]{\includegraphics[width=0.5\textwidth]{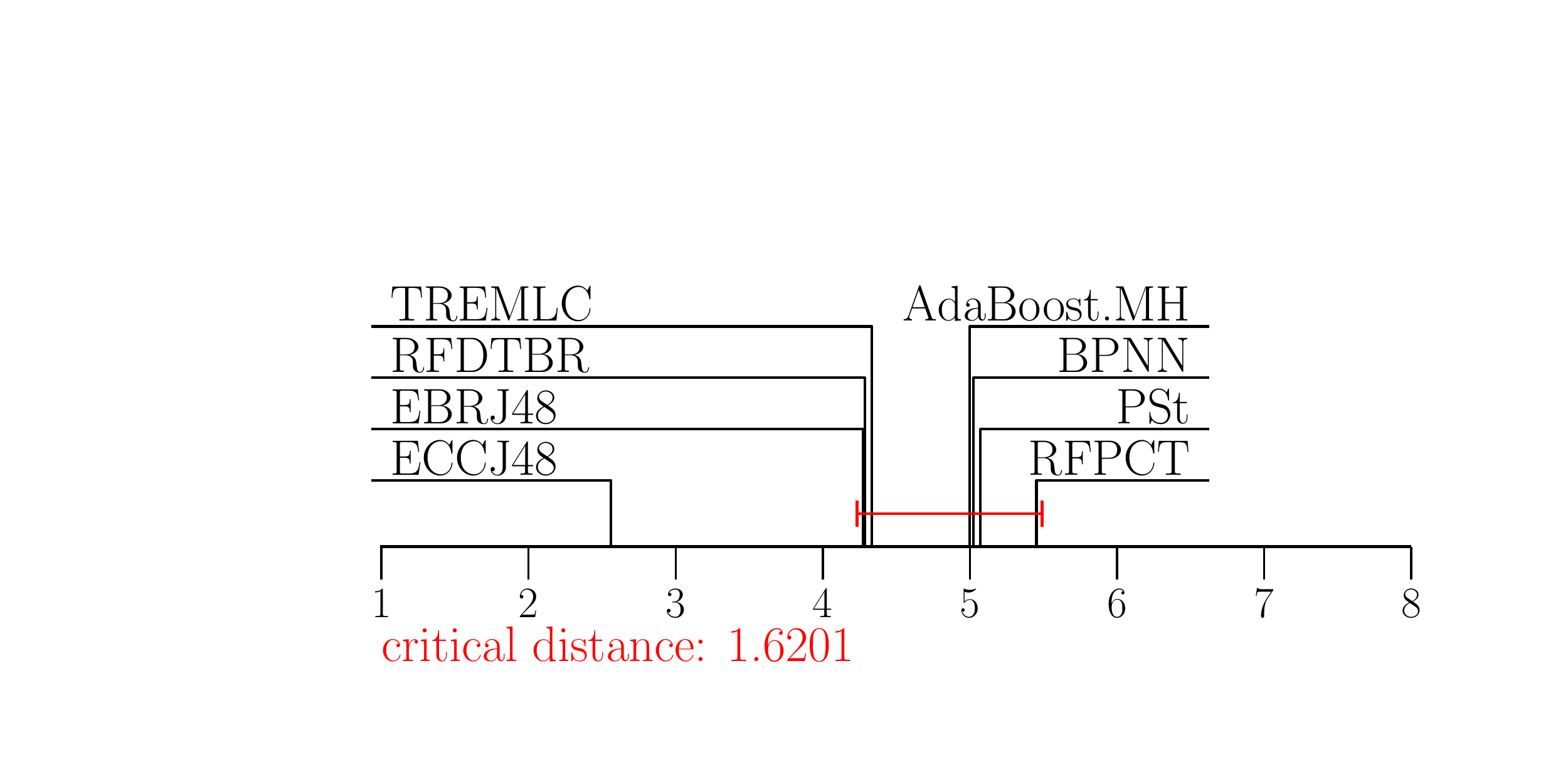}%
\label{fig:main:MacroRecall}}
\subfloat[Micro recall]{\includegraphics[width=0.5\textwidth]{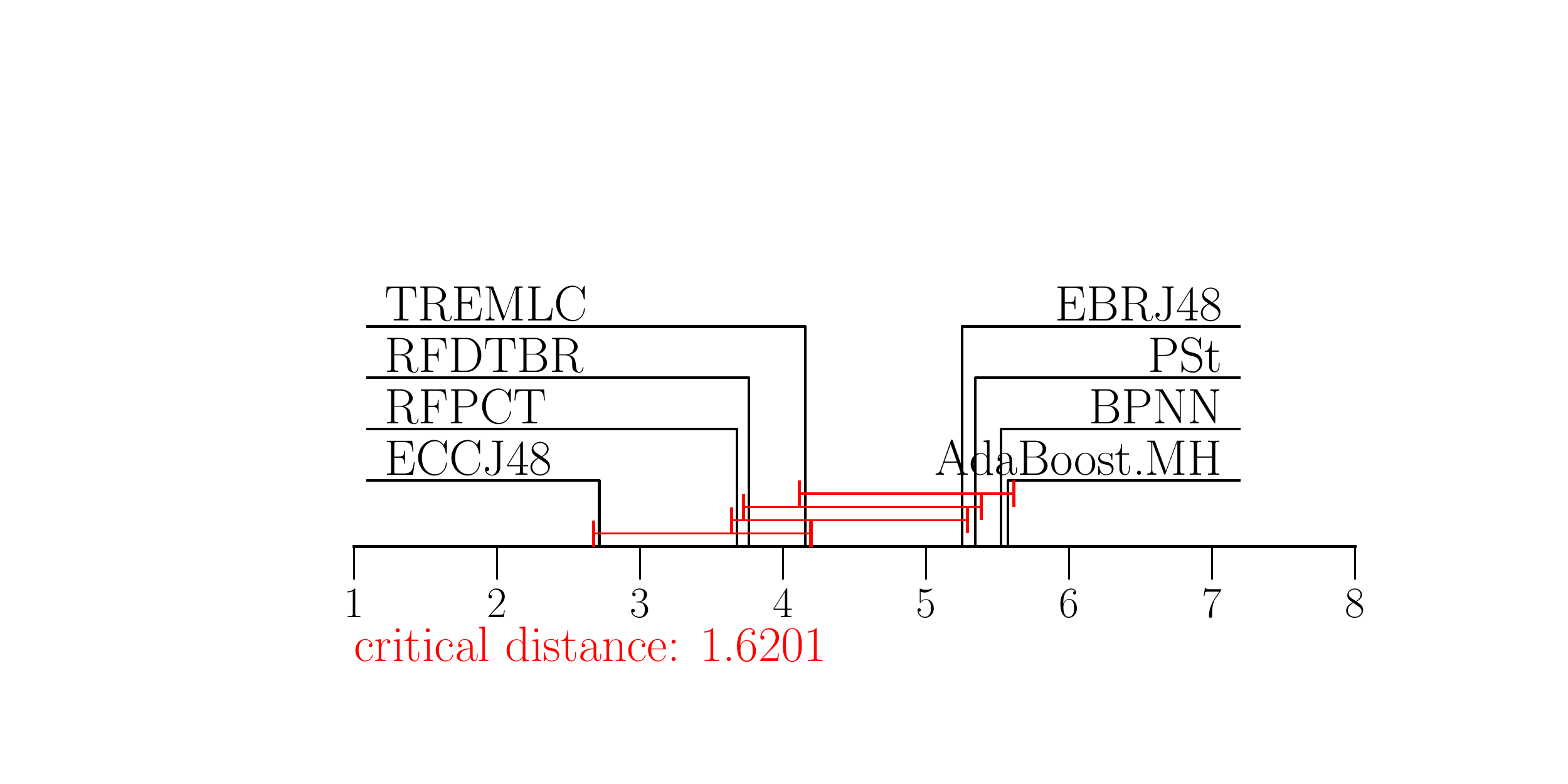}%
\label{fig:main:MicroRecall}}

\subfloat[Macro F1]{\includegraphics[width=0.5\textwidth]{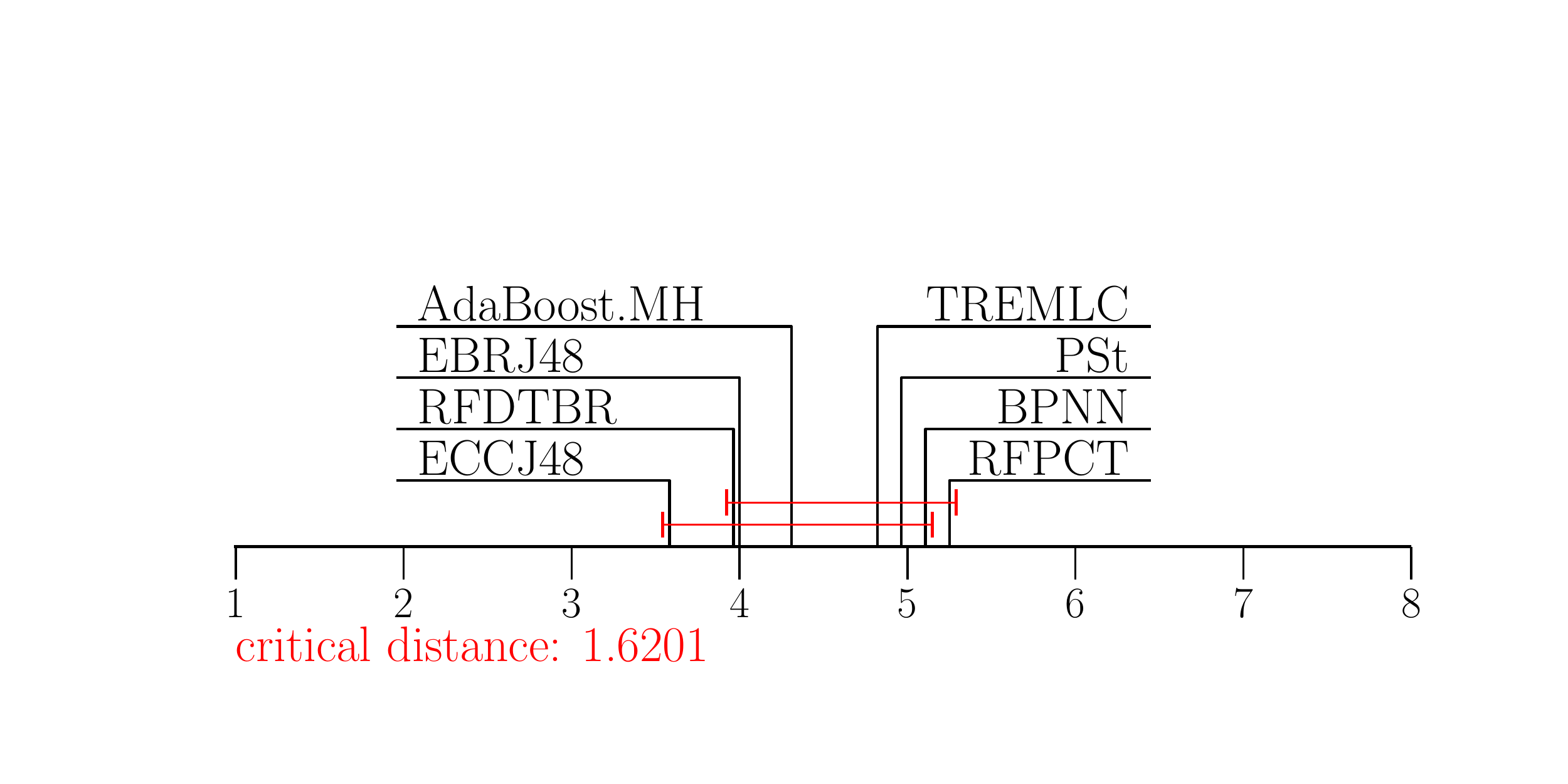}%
\label{fig:main:MacroF1}}
\subfloat[Micro F1]{\includegraphics[width=0.5\textwidth]{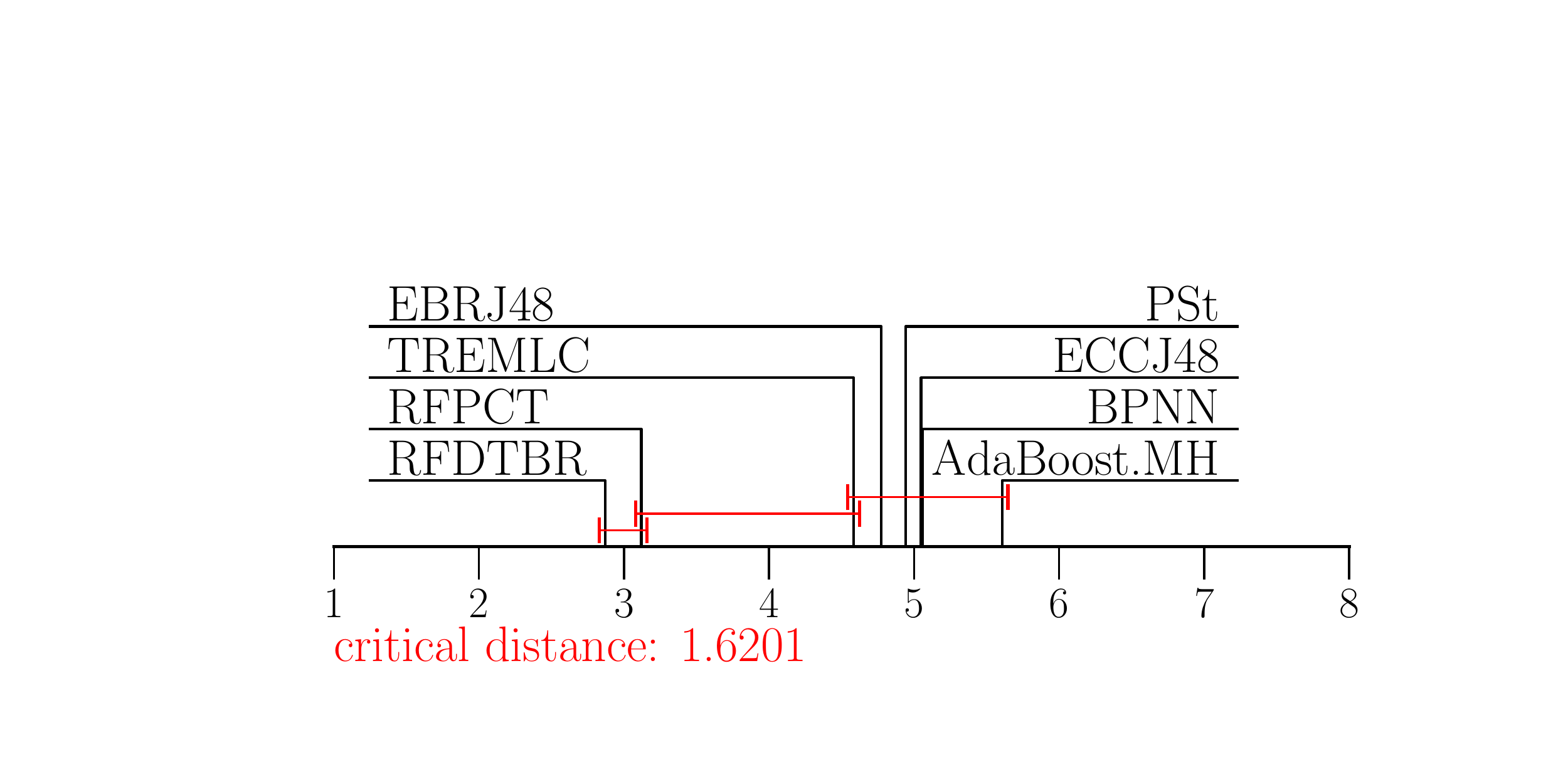}%
\label{fig:main:MicroF1}}
\caption{Average rank diagrams comparing the best MLC methods using label-based measures.}
\label{fig:main:label}
\end{figure*}

\renewcommand\theadalign{cl}

\end{document}